\def\year{2021}\relax
\documentclass[letterpaper]{article} 
\usepackage{aaai21}  
\usepackage{times}  
\usepackage{helvet} 
\usepackage{courier}  
\usepackage[hyphens]{url}  
\usepackage{graphicx} 
\usepackage{natbib}  
\usepackage{caption} 
\usepackage{xspace}

\newcommand{\ie}{}
\def\ie/{i.\,e.}
\newcommand{\eg}{}
\def\eg/{e.\,g.}
\newcommand{\naive}{}
\def\naive/{na\"{\i}ve}

\def\sys/{DocParser}
\def\dataset/{arXivdocs}
\def\datasetmanual/{arXivdocs-target}
\def\datasetauto/{arXivdocs-weak}

\def\Entity/{Entity}
\def\Entities/{Entities}
\def\entity/{entity}
\def\entities/{entities}
\def\relation/{relation}
\def\relations/{relations}
\def\Relation/{Relation}
\def\Relations/{Relations}

\title{\sys/: Hierarchical Document Structure Parsing from Renderings}

\author {
    Johannes Rausch,\textsuperscript{\rm 1}  Octavio Martinez,\textsuperscript{\rm 1} Fabian Bissig,\textsuperscript{\rm 1}\\
    Ce Zhang,\textsuperscript{\rm 1}
    Stefan Feuerriegel\textsuperscript{\rm 2}\\
}
\affiliations {
    \textsuperscript{\rm 1} Department of Computer Science, ETH Zurich \\
    \textsuperscript{\rm 2} Department of Management, Technology, and Economics, ETH Zurich\\
    \small{johannes.rausch@inf.ethz.ch, octaviom@student.ethz.ch, fbissig@student.ethz.ch\\ce.zhang@inf.ethz.ch, sfeuerriegel@ethz.ch}
}

\usepackage{latexsym}

\usepackage{multibib}
\newcites{appendix}{Appendix-References}
\usepackage{placeins}
\usepackage{textcomp}
\usepackage{pgfplots}
\usepackage{pgfplotstable}
\pgfplotsset{compat=newest}
\usepackage{adjustbox}
\usepackage{array}
\usepackage{tikz}
\usepackage{makecell}

\usepackage{amsmath}
\usepackage[title,toc]{appendix}
\usepackage{csvsimple}
\usepackage[eulergreek]{sansmath}
\usepackage{bm}
\usepackage{algorithm2e} 
\usepackage[group-separator={,}]{siunitx}

\newcommand\clipright[1][white]{
  \fill[#1](current axis.south east)rectangle(current axis.north-|current axis.outer east);
  \pgfresetboundingbox
  \useasboundingbox(current axis.outer south west)rectangle([xshift=.5ex]current axis.outer north-|current axis.east);
}

\usepackage{csquotes}
\usepackage{booktabs}
\usepackage{subcaption}
\usepackage{enumitem}

\makeatletter
\renewcommand{\fps@figure}{htbp}         
\renewcommand{\fps@table}{htbp}         
\makeatother 

\definecolor{ETH3}{RGB}{0, 105, 180} 
\definecolor{ETH6}{RGB}{111,111,110} 
\definecolor{ETH7}{RGB}{168,50,45}

\begin{document}

\maketitle
\begin{abstract}
Translating renderings (\eg/ PDFs, scans) into hierarchical document structures is extensively demanded in the daily routines of many real-world applications. 
However, a holistic, principled approach to inferring the complete hierarchical structure of documents is missing. As a remedy, we developed ``\sys/'': an end-to-end system for parsing the complete document structure -- including all text elements, nested figures, tables, and table cell structures. 
Our second contribution is to provide a dataset for evaluating hierarchical document structure parsing. Our third contribution is to propose a scalable learning framework for settings where domain-specific data are scarce, which we address by a novel approach to weak supervision that significantly improves the document structure parsing performance. Our experiments confirm the effectiveness of our proposed weak supervision: Compared to the baseline without weak supervision, it improves the mean average precision for detecting document \entities/ by $39.1\,\%$ and improves the F1 score of classifying hierarchical \relations/ by $35.8\,\%$.
\end{abstract}

\section{Introduction}

The structural and layout information in a document can be a rich source of information that facilitates Natural Language Processing (NLP) tasks (\eg/ information extraction). Over the years, the NLP community has developed a range of techniques to \textit{detect}, \textit{understand}, and \textit{take advantage of} document structures 
\cite{hurst-nasukawa-2000-layout,10.3115/990820.990845,tengli-etal-2004-learning,Luong2011,govindaraju-etal-2013-understanding,Katti2018,Schafer2011, Schafer2012, garncarek2020lambert}.

However, structural information in documents is becoming increasingly challenging to obtain --- many file formats that are prevalent today are being rendered without structural information. Prominent examples are PDF documents: this file format benefits from portability and immutability, yet it is flat in the sense that it stores all content as isolated \entities/ (\eg/, combinations of characters and positions) and, thus, hierarchical information is lacking. As such, the structure behind figures and especially tables is discarded and thus no longer available to computerized analyses in NLP. In contrast, file formats such as XML or JSON naturally encode hierarchical document structures among textual \entities/. Hence, techniques are required in order to convert renderings into structured, textual document representations to enable \textit{joint inference} between text, layout, and other document structures.

Earlier attempts for structure parsing on documents focused on a subset of simpler tasks such as segmentation of text regions \citep{Antonacopoulos2009}, locating tables \citep{Zanibbi2004,Embley2006}, or parsing them \citep{Schreiber2018}, but not parsing complete document structures. However, document structures are required as a representation of many downstream tasks in NLP. For instance, recent efforts in the NLP community \citep{Katti2018, apostolova2014combining, liu2019graph} have shown that utilizing 2D document information, \eg/ character and word positions, can be an effective way to improve upon standard NLP tasks such as information extraction.

A holistic, principled approach for inferring the complete hierarchical structure from documents is missing. On the one hand, such a task is nontrivial due to the complexity of documents, particularly their deeply-nested structures. For instance, nested tables are fairly easy to recognize for human readers, yet detecting them is known to impose computational hurdles \citep[cf.][]{Schreiber2018}. On the other hand, efficient learning is prevented as large-scale training sets are lacking \citep[cf.][]{Arif2019,Schreiber2018}. Notably, prior datasets are limited to table structures \citep{Gobel2013,Rice1995} and not the complete document structures. Needless to say, complex structures also make the labeling process significantly more costly \citep{Wang2004}. Therefore, an effective implementation that makes only a scarce use of labeled data is demanded.

\begin{figure}[t!]
\centering
\includegraphics[width=0.8\linewidth]{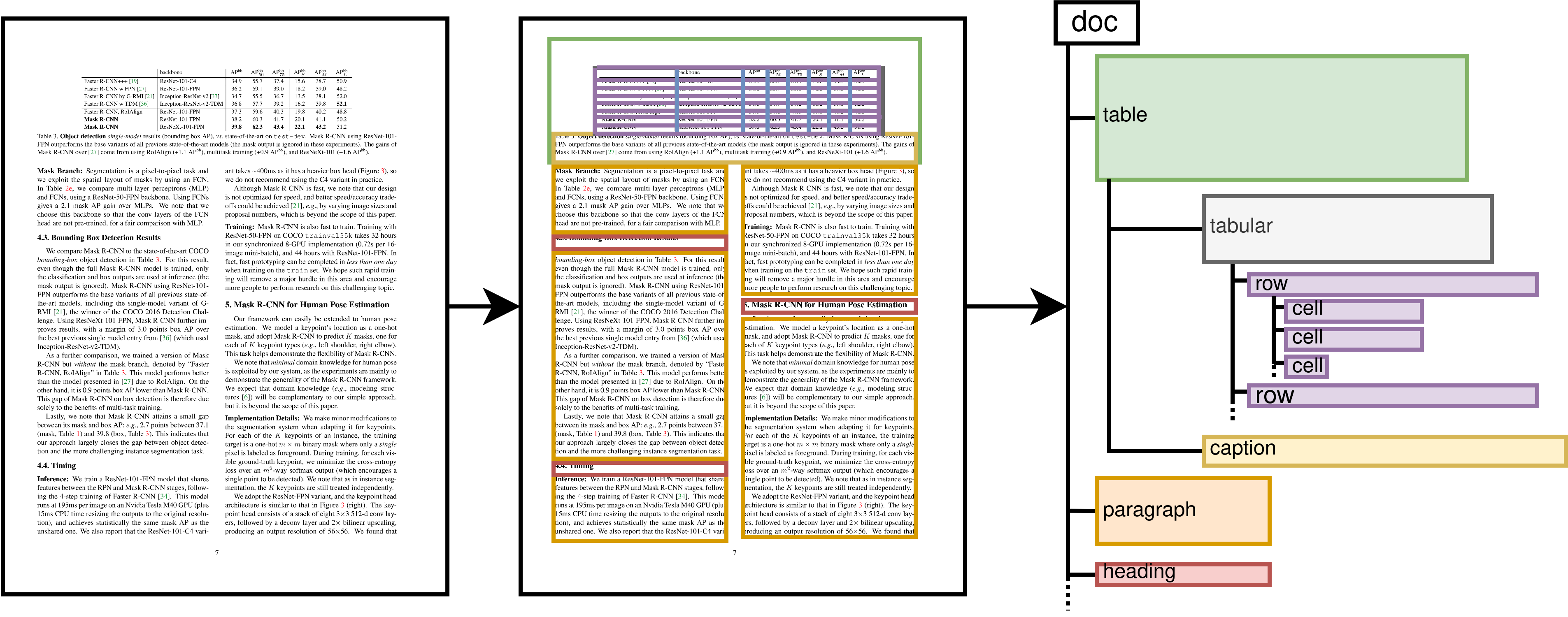}
\caption{\sys/ takes rendered document images (left) as input, performs segmentation into bounding boxes (center), and then outputs the hierarchical structure of the full document (right). Shown is an illustrative sketch; examples are provided in the supplements.} 
\label{fig:inputoutput}
\end{figure}

This work focuses on parsing the hierarchical document structure from renderings. We develop an end-to-end system for inferring the complete document structure (see Figure~\ref{fig:inputoutput}). This includes all \entities/ (\eg/, text, bibliography regions, figures, equations, headings, tables, and table cells), as well as the hierarchical \relations/ among them. We specifically adapt to settings in practice that suffer from data scarcity. For this purpose, we propose a novel learning framework for scalable weak supervision. It is intentionally tailored to the specific needs of parsing document renderings; that is, we create weakly-supervised labels by utilizing the reverse rendering process of \LaTeX. The reverse rendering returns the bounding boxes of all \entities/ in documents together with their category (\eg/, whether the \entity/ is a table or a figure, etc.). Yet the outcomes are noisy (\ie/, imprecise bounding boxes, missing \entities/, incorrect labels) and without deep structure information (\eg/ information such as table row numbers is missing). Nevertheless, as we shall see later, the generated data greatly facilitates learning by being treated as weak labels.

\textbf{Contributions}:\footnote{Source codes and the \dataset/ dataset are available from \url{https://github.com/DS3Lab/DocParser}.} We extend prior literature on document parsing in the following directions:
\begin{enumerate}[topsep=0pt,itemsep=-1ex,partopsep=1ex,parsep=1ex,leftmargin=2.5ex]
\item We contribute ``\sys/''. This presents the first end-to-end system for parsing renderings into \emph{hierarchical} document structures. Prior literature has merely focused on simpler tasks such as table detection or table parsing but not on the parsing of complete documents. As a remedy, we present a system for inferring document structures in a holistic, principled manner. 
\item We contribute the first dataset (called ``{\dataset/}'') for evaluating document parsing. It extends existing datasets for parsing in two directions: (i)~it includes all \entities/ that can appear in documents (\ie/ not just tables) and (ii)~it includes the hierarchical \relations/ among them. The dataset is based on \num{127472} scientific articles from the arXiv repository.
\item We propose a novel weakly-supervised learning framework
to foster efficient learning in practice where annotated documents are scarce. It is based on an automated and thus scalable labeling process, where annotations are retrieved by reverse rendering the source code of documents. Specifically, in our work, we utilize \TeX\ source files from arXiv together with \texttt{synctex} for this objective. This then yields weakly-supervised labels by reverse rendering of the \TeX\ code. 
\item We conduct extensive evaluation of our proposed techniques, outperforming the state-of-the-art on the related task of table parsing.
\end{enumerate}

\section{\sys/ System} 

\subsection{Problem Description}

Given a set of document renderings $D_1, \ldots, D_n$, the objective is to generate hierarchical structures $T_1, \ldots, T_n$. A hierarchical structure $T_i$, $i=1,\ldots,n$, consists of both \entities/ and \relations/ as follows:\footnote{For consistency, we use the term ``\entity/'' throughout the article when referring to all elements in the document structure (\eg/ a figure, table, or text) that need to be detected. While the term ``object'' is common in computer vision, we chose the term ``\entity/'' to highlight its semantic nature for NLP.}

\noindent
\textbf{Entities} $E_j$, $j = 1, \ldots, m$, refer to the various elements within a document, such as a figure, table, row, cell, etc. Each \entity/ is described by three attributes: (1)~its semantic category $c_j \in \mathcal{C} = \{C_1, \ldots, C_l\}$ (\ie/, which defines the underlying type) and (2)~the coordinates given by rectangular bounding box $B_j$ in the document rendering. There further is (3)~a confidence score $P_j$. This is not part of the ground truth labels; however, it comes from the predictions inside the \sys/ system.

\noindent
\textbf{Relations} $R_j$, $j = 1, \ldots, k$ of type $\Psi$ are given by triples $(E_{\text{subj}}, E_{\text{obj}}, \Psi)$ consisting of a subject $E_{\text{subj}}$, an object $E_{\text{obj}}$, and a \relation/ type $\Psi \in \{ \mathit{parent\_of},  \mathit{followed\_by}, \text{null} \}$. The latter, null, is reserved for \entities/ with meta-information that do not have designated order (\ie/, header, footer, keywords, date, page number). All other \entities/ must have $\Psi \neq \text{null}$.

The combination of \entities/ and \relations/ is sufficient to reconstruct the hierarchical structure $T_i$ for a document. However, generating such a hierarchical structure from a document rendering $D_i$ is subject to inherent challenges: the similar appearance of \entities/ impedes detection and, further, the hierarchy can be nested arbitrarily, with substantial variation across different documents.

\begin{figure}[t!]
\centering
\includegraphics[width=0.8\linewidth]{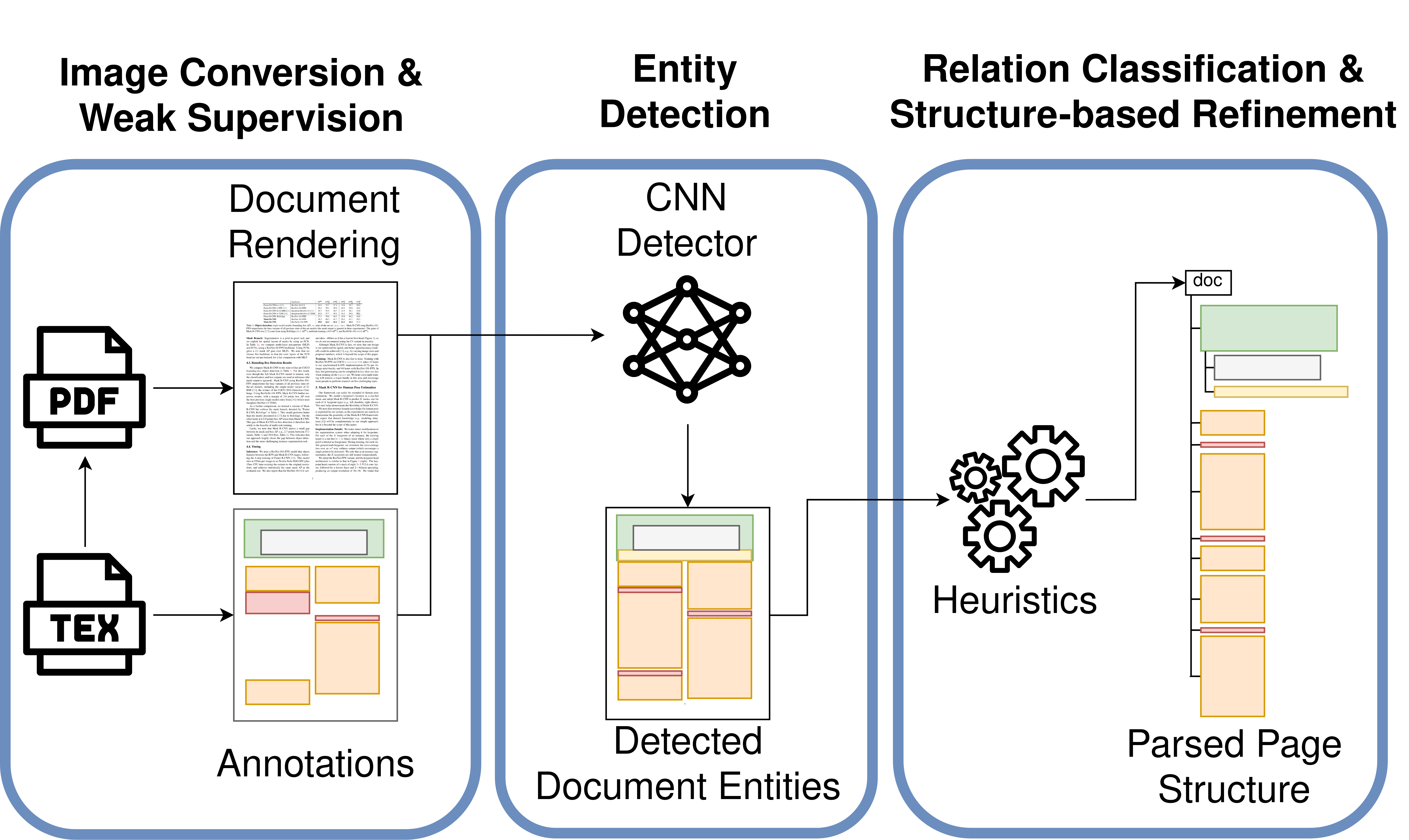}
\caption{System overview.}
\label{fig:sysoverview}
\end{figure}

\noindent
\subsection{System Components}
\sys/ performs document structure parsing via 5 components (see overview in  Figure \ref{fig:sysoverview}): (1) image conversion, (2) entity detection, (3) relation classification, (4) structure-based refinement, and (5) scalable weak supervision.
To store document structures, we developed a customized, JSON-based file format.

\noindent
\textbf{Component 1: Image Conversion}

Document renderings are converted into images with a predefined resolution $\rho$. Furthermore, all images are resized to a fixed rectangular size $\phi$ (if necessary, with zero padding). 

The document images are further pre-processed: the RGB channels of all document images are normalized analogous to the MS COCO dataset (\ie/, by subtracting the mean RGB channel values from the inputs). The reason is that all neural models are later initialized with pre-trained weights from the MS COCO dataset \citep{Lin2014}.

\noindent
\textbf{Component 2: Entity Detection}

To detect all document \entities/ within a document image, we build upon a neural model for image segmentation, namely Mask~R-CNN \citep{He2017}. Specifically, it takes the images from the previous component as input and then returns a flat list of \entities/ $E_1 ,\ldots, E_m$ as output. For each \entity/ Mask~R-CNN determines (i)~its rectangular bounding box, (ii)~confidence score, (iii)~a binary segmentation mask that distinguishes between the detected \entity/ and background pixels within the bounding box, and (iv)~a category label for the \entity/. Our implementation makes use of \num{23} categories $\mathcal{C}$: \textsc{content block}, \textsc{table}, \textsc{table row}, \textsc{table column}, \textsc{table cell}, \textsc{tabular}, \textsc{figure}, \textsc{heading}, \textsc{abstract}, \textsc{equation}, \textsc{itemize}, \textsc{item}, \textsc{bibliography block}, \textsc{table caption}, \textsc{figure graphic}, \textsc{figure caption}, \textsc{header}, \textsc{footer}, \textsc{page number}, \textsc{date}, \textsc{keywords}, \textsc{author}, \textsc{affiliation}.\footnote{For consistency, this formatting is utilized for all \entities/.}

\noindent
\textbf{Component 3: \Relation/ Classification}

A set of heuristics is applied to translate the flat list of \entities/ into hierarchical \relations/ $R_1, \ldots, R_k$. Here, we distinguish the heuristics according to whether they generate (1)~the \textbf{nesting} among \entities/ or (2)~the \textbf{ordering} for \entities/ of the same nesting level. The former case corresponds to $\Psi = \mathit{parent\_of}$, while the latter determines all \relations/ with $\Psi = \mathit{followed\_by}$. In this component, we ignore all \entities/ with meta-information, \eg/ footers, as these have no designated hierarchy (cf. document grammar in the supplements).

\textbf{\Relations/ with Nesting} ($\mathit{parent\_of}$): Four heuristics $h_1, \ldots, h_4$ determine parent-child \relation/ as follows:

\noindent
\textbf{($\bm{h_1}$: Overlaps)} A list of candidate parent-child \relations/ is compiled based on the overlap of bounding boxes. That is, \sys/ loops over all bounding boxes and, for each bounding box $B_{\text{subj}}$, it determines all other bounding boxes that are contained within  $B_{\text{subj}}$. 

Formally, this is given by all tuples of bounding boxes $(B_\text{subj}, B_\text{obj})$ with $\text{subj} \in m$, $\text{obj} \in m$, and $\text{subj}\neq \text{obj}$ where $h_1(B_\text{subj}, B_\text{obj})$ is satisfied: Tuples for which the bounding box of $B_\text{obj}$ is fully or partially enclosed by the bounding box of $B_\text{subj}$ are added to the candidate list. 
Furthermore, we add tuples to the candidate list that satisfy $\frac{\mathit{area}(B_\text{subj} \cap B_\text{obj})}{\mathit{area}(B_\text{obj})} \geq \theta_1$ and $\frac{\mathit{area}(B_\text{subj})}{\mathit{area}(B_\text{obj})} > \theta_2$, \ie/ they must have a certain overlap fraction $\theta_1$ and size ratio $\theta_2$. In \sys/, thresholds of $\theta_1 = 0.45$ and $\theta_2=1.2$ are used.

\noindent
\textbf{($\bm{h_2}$: Grammar Check)} This heuristic validates the candidate list against a predefined document grammar (see document grammar in the supplements). Concretely, all illegal candidates, e.g., a \textsc{tabular} nested inside a \textsc{figure}, are removed.

\noindent

\noindent
\textbf{($\bm{h_3}$: Direct children)} The candidate list is further pruned so that it contains only direct children of the parent and not sub-children. For this purpose, all sub-children are removed. As an example, this should remove $(E_{\text{subj}}^1, E_{\text{obj}}^3)$ from a candidate list $\{(E_{\text{subj}}^1, E_{\text{obj}}^2), (E_{\text{subj}}^1, E_{\text{obj}}^3), (E_{\text{subj}}^2, E_{\text{obj}}^3) \}$, since it represents a sub-child and not a direct child of $E_{\text{subj}}$.

\noindent
\textbf{($\bm{h_4}$: Unique Parents)} The candidate list is altered so that each \entity/ has only a single parent. Formally, if an \entity/ $E_{\text{obj}}$ has multiple candidate parents, we first compare the Intersection over Union (IoU) of the bounding boxes of all candidate parents with $E_{\text{obj}}$: $\text{IoU} = \frac{\mathit{area}(B_\text{subj} \cap \hat{B}_\text{obj})}{\mathit{area}(B_\text{subj} \cup \hat{B}_\text{obj})}$. We then keep the parent with the maximal IoU, while all others are removed. If two parents have the same IoU, we select the element with the highest confidence score $P_j$ as parent. If that value is also equal, we choose the \entity/ with the largest bounding box.

\textbf{\Relations/ with Ordering} (\textit{followed\_by}): The \entities/ are ordered according to the general reading flow (\ie/, from left to right). Here care is needed so that multi-column pages are processed correctly. For this, two heuristics $o_1$ and $o_2$ are used. By default, all \entities/ are processed by both heuristics. Children of floating \entities/ are only processed by heuristic $o_2$, however.

\noindent
\textbf{($\bm{o_1}$: Page Layout Entities)} First, all \entities/ are grouped according to their coordinates on the document page, namely, into groups belonging to the (a)~left side $G_l$, (b)~center $G_c$, or (c)~right side $G_r$. Formally, this is achieved by computing the overlap for each \entity/ $E_j$, $j = 1, \ldots, m$ with the left (and right) side of a document page, \ie/, $\tau_\text{ovlp} = \text{overlap} / \textit{width}(B)$. 
If the overlap with either the left (or the right) side is above a threshold (\ie/, $\tau_\text{ovlp}>0.7$), the \entity/ $E_j$ is assigned to the left (or right) side.

Otherwise, if such assignment is not possible with high confidence, the \entity/ $E_j$ is assigned to center group $G_c$. In essence, the center group is an indicator whether the document is in single- or multi-column.

If no \entities/ have been assigned to the center group (\ie/, $G_c = \emptyset$), then the \entities/ are ordered first according to $G_l$ followed by $G_r$. Within each group, the \entities/ are ordered top-to-bottom and then left-to-right by applying heuristic $o_2$. In sum, this approach should find an appropriate ordering for multi-column pages.
If \entities/ have been assigned to the center group  (\ie/, $G_c \neq \emptyset$), then grouping is further decomposed into additional subgroups: the \entities/ $E \in G_c$ from the center group are used to split $G_l$, $G_c$, and $G_r$ into vertical subgroups $G_l^\iota$, $G_c^\iota$, and $G_r^\iota$, respectively. Afterward, we loop over all vertical subgroups $\iota$. For each, we order the \entities/ according to the group (first $G_l^\iota$, followed by $G_c^\iota$ and then $G_r^\iota$). Within each subgroup, we perform the ordering via heuristic $o_2$. This approach should correctly arrange \entities/ in two cases: (1)~in single-column pages and (2)~when multi-column pages are split into different chunks by full-width figures or tables. 

For each subgroup, we perform the ordering via heuristic $o_2$.

\noindent
\textbf{($\bm{o_2}$: Reading Flow)} The \entities/ $E_j$, $j = 1, \ldots, m$, are ordered top-to-bottom and, within lines, left-to-right, so that it matches the usual reading flow in documents. Formally, let the top-left corner of a document image refer to the coordinate $(0,0)$. Furthermore, let us consider the top-left location of all bounding boxes $B_j$. The top-left location is then used to sort the \entities/ first by their $y$-coordinate of $B_j$ and, if equal, by their $x$-coordinate (both ascending).

\noindent
\textbf{Component 4: Structure-Based Refinement}
We utilize the classified relations to iteratively refine \entities/ and \relations/ in four steps when parsing full document pages:

\textbf{(1)} For each \entity/ $E_\text{parent}$ with $l$ child entities $E_\text{child}^{1}, \ldots ,E_\text{child}^{l}$, we update its bounding box such that $B_\text{parent} = \text{union}(B_\text{parent},  B_\text{child}^{1},  \ldots, B_\text{child}^{l})$.
\textbf{(2)} For parent \entities/ $E_\text{parent}$ with exactly one child \entity/ of the same category, we remove the child entity and update $B_\text{parent}$ such that it is the union of parent and child bounding boxes. We also consider \entity/ pairs of categories that do not conform to the document grammar. This allows us to dismiss duplicate entities of any category.
\textbf{(3)} If an \entity/ $E_\text{child}$ is sibling to other \entities/ in a way that conflicts the document grammar, we generate a new \entity/ that encloses $E_\text{child}$ to achieve conformity with the document grammar. Concretely, nested \textsc{figure} structures are defined such that one \textsc{figure} should at most contain one \textsc{figure graphic} \entity/ child. If multiple \textsc{figure graphic} are classified as children, we wrap each of them individually into new \textsc{figure} \entities/.
\textbf{(4)} If no parent is found for an \entity/ $E_\text{child}$ that should only occur as a child \entity/, we identify a suitable parent \entity/ by analyzing its neighboring siblings as follows: we consider all \entities/ that jointly appear in an ordering relation with $E_\text{child}$ as a candidates $E_\text{cand}$. We dismiss candidates of category $\mathcal{C}$ that would not conform to the hierarchies defined in the document grammar. Finally, we dismiss any candidate for which $B_\text{cand} \cap B_\text{child} = \emptyset$. If exactly one candidate remains, we update its bounding box $B_\text{cand} = \text{union}(B_\text{cand}, B_\text{child})$. 

The updates to the set of \entities/ can lead to further changes to the classified relations. For this reason, whenever changes are made to \entities/ in one of the four refinement steps, we update the relations via Component 3 and move back to refinement step (1). The refinement is completed once no change is applied in any of the steps or a limit of $r$ loop iterations has been reached.\footnote{Details on our parameter choice and pseudocode are included in the supplements.}

\noindent
\textbf{Component 5: Scalable Weak Supervision}

The system is further extended by scalable weak supervision. This aims at improving the performance of \entity/ detection and, as a consequence, of end-to-end parsing. 

Our weak supervision builds upon an additional dataset that consists of source codes (rather than document renderings). The source codes allow us to create a mapping between \entities/ in the source code and their renderings. This process has three particular characteristics: first, the mapping is noisy and thus creates only weak labels. Despite that, the weak labels can aid efficient learning. Second, annotations are obtained only for some \entities/ and \relations/. Third, if automated, this process circumvents human annotations and is thus highly scalable.

Let the unlabeled \entities/ found in the source code be given by $S_1, \ldots, S_k$. For them, we generate weak labels $W_1, \ldots, W_k$ consisting of a semantic category and coordinates of the bounding box. However, both the semantic category and the bounding box can be subject to noise. Furthermore, weak labels are generated merely for a subset $\mathcal{C}' \subseteq \mathcal{C}$ of the semantic categories.

In \sys/, the weak supervision is based on \TeX\ source files that are used to generate document renderings in the form of PDF files. The mapping between both formats is then obtained via \texttt{synctex} \cite{laurens2008direct}. \texttt{synctex} is a synchronization tool that performs a reverse rendering, so that PDF locations are mapped to \TeX\ code. For given coordinates in the document rendering, \texttt{synctex} returns a list of rectangular bounding boxes and the corresponding source code. Notably, the inference bounding boxes represent noisy labels, since the resulting \entity/ annotations could be wrongly labeled, shifted, or entirely missing.

We proceed as follows. We iterate through the source code and retrieve bounding boxes for all \TeX\ commands.
We then map the source code to our \entities/ $E$. For instance, the bounding box for \TeX\ code \texttt{\textbackslash includegraphics} inside a \texttt{\textbackslash begin\{figure\} ,\ldots, \textbackslash end\{figure\}} environment is mapped onto a \textsc{figure\_graphic} \entity/ that is nested inside a \textsc{figure} \entity/. 
Bounding boxes for all \entities/ that act as inner children are created dynamically by computing the union bounding of all child bounding boxes.

We perform following processing steps to generate noisy labels for weak supervision:

\begin{enumerate}
    \item Bounding boxes that are retrieved for simple text tokens inside the source code are mapped to \textsc{content line} \entities/.
    \item  If we encounter \texttt{environments} or \texttt{commands} (\eg/, \texttt{\textbackslash begin\{itemize\}} or \texttt{\textbackslash item}), we create corresponding candidate \entities/. All \entities/ retrieved for tokens inside the scope of these environments are created as nested child \entities/. This approach is used to create the following \entity/ types, namely \textsc{figure}, \textsc{figure graphic}, \textsc{figure caption}, \textsc{table}, \textsc{tabular}, \textsc{table caption}, \textsc{itemize}, \textsc{item}, \textsc{abstract}, and \textsc{bibliography}. 
Any other \entities/ are mapped onto the \textsc{content line} category.
    \item We utilize a special characteristic of \texttt{synctex} to identify \textsc{equation}, \textsc{equation formula} and \textsc{equation label} \entities/: bounding boxes returned by \texttt{synctex} are highly uniform and typically consist of per-line bounding boxes of consistent width and $x$-coordinates. Equations and labels are an exception to this rule and typically only consist of vertically aligned bounding boxes of smaller width.
    \item The sectioning structure of documents is considered: any type of \texttt{section} command is mapped to a \textsc{section} \entity/. The argument of the sectioning command, \eg/ \texttt{\textbackslash subsection\{titlearg\}} is mapped via \texttt{synctex} to a \textsc{header} \entity/. Entities generated from code in the scope of a section are created as children to the section \entity/ that corresponds to the current section scope.
    \item Within sections, we sort \entities/ based on a top-to-bottom, left-to-right reading order. Using these sorted lists of sibling \entities/, we form \textsc{content block} \entities/ from subsequent groups of \textsc{content line} \entities/ within page columns. If such block occurs within a \textsc{bibliography} environment, we instead map it to a \textsc{bibliography block} \entity/.
    \item In \textsc{table} environments, we consider all child \entities/ (except captions) that do not span across a whole table width as \textsc{cell} and the remainder as \textsc{table row}. As we shall see later, this is effective at retrieving complex table structures.
    \item We use the detected table cells to generate rows and columns as follows: We compute the centroids of all cells. To identify rows, we consider the sorted $y$-coordinates of the centroids and group them such that the pixel-wise distance between two consecutive y-coordinates in a group is smaller or equal to $5$. If any identified group contains two or more centroid y-coordinates, we create a \textsc{table row} \entity/ from the union of the corresponding \textsc{table cell} \entities/. Analogously, using the $x$-coordinates of the cell centroids, we identify \textsc{table column} \entities/. 
    \item Additional cleaning steps are performed for tables and figures: Child \entities/ with width or height of 2 or fewer pixels are discarded. Caption bounding boxes that enclose other non-caption child \entities/ are also discarded. 
    \item We make sure that \entities/ contain at most one leaf node by moving excess leaves into newly generated \textsc{content line} \entities/.
    \item We remove duplicate bounding boxes and \entities/ without any leaf nodes in their respective sub-tree. Candidates are filtered such that only a group of \entities/ and their respective sub-tree are preserved: \textsc{itemize}, \textsc{figure}, \textsc{table}, \textsc{equation}, \textsc{heading}, \textsc{content block}, \textsc{bibliography}, \textsc{abstract}.
\end{enumerate} 

During training, \entities/ with obvious errors are dismissed, \ie/ leaf nodes or \entities/ with bounding boxes that extend beyond page limits or with area of $0$.

\section{Datasets with Document Structures}

We contribute the dataset ``\dataset/'' that is tailored to the task of hierarchical structure parsing. It comes in two variants: \textbf{\datasetmanual/} and \textbf{\datasetauto/}. (1)~\datasetmanual/ contains documents that have been manually checked and annotated. (2)~\datasetauto/ contains a large-scale set of documents that have no manual annotations but that can be used for weak supervision.

\subsection{\datasetmanual/}

\datasetmanual/ provides a set of documents with manual annotations of the complete document structure. These documents were randomly selected from arXiv as an open repository of scientific articles, but in a way such that each has at most \num{30} pages and contains at least one \textsc{table} within the source code. Altogether, it counts \num{362} documents. \datasetmanual/ comes with predefined splits for \emph{training}, \emph{validation}, and \emph{eval} that consist of $160$, $79$, $123$ documents, respectively. The dataset comprises of $30$ different \entity/ categories.\footnote{Some \entity/ categories are extremely rare and, hence, only a subset is later used as part of our experiments.} We ensure a fairly uniform distribution of \entity/ categories across different splits by sampling one random page rendering for each of the \num{362} documents that contain an \textsc{abstract}, \textsc{figure}, or \textsc{table}. 
On average, each document contains $86.32$ \entities/. The number of leaf nodes in the document graph as well as the frequency and average depth of the different \entities/ are reported in the supplements. 

Evidently, the most common category in the dataset is \textsc{content line} (34.33\,\%). This is because they typically represent leaf nodes in the graph and are children of larger \entities/ such as \textsc{abstract}, \textsc{caption}, or \textsc{content block}. 

Annotators were instructed to follow the document grammar during labeling. Annotation of disallowed hierarchies is, however, possible to provide them the freedom to deal with the range of different document representations. Document annotations are automatically initialized by our scalable weak supervision mechanism to speed up the annotation process. The labelers were instructed to annotate \entities/ only up to the coarseness that is used by \sys/, \eg/ labeling content blocks, rather than individual lines. 

\subsection{\datasetauto/}

\datasetauto/ contains \num{127472} documents with an average length of \num{12.84} pages that were retrieved from arXiv. We selected only documents that have a length of at most $30$ pages and contain at least one \textsc{table} within their source code. For reproducibility, we make our weak labels available.\footnote{For this purpose, the dataset was labeled via our proposed weak supervision mechanism and thus contains both \entities/ $E_j$ and hierarchical \relations/ $R_j$. For reasons of space of the physical files, bounding boxes are only stored for \entities/ in leaf nodes. For all other \entities/, the bounding boxes can be calculated by taking the union bounding box of their children.}

\section{Computational Setup}

\subsection{Mask R-CNN} 
Mask R-CNN extends the architecture of a convolution neural network with skip connections \cite{He2016} so that it is highly effective for image segmentation and \entity/ detection.\footnote{A model illustration is included in the supplements.} Formally, it comprises of multiple stages with decreasing spatial resolution. The output of these stages is then fed into a so-called feature pyramid network (FPN) \citep{Lin2017}. The FPN then interconnects these inputs in multiple stages of increasing spatial resolution to produce multi-scale feature maps. Specifically, we use a ResNet-110 architecture \cite{He2016} to extract features in 5 stages at different resolutions. The outputs of stages 2 to 5, denoted as $C_2, \ldots, C_5$, are passed to the FPN.
The FPN outputs a total of 5 feature maps ${P_2, \ldots, P_6}$ at different resolutions. We refer the reader to \citep{Lin2017} for a detailed description of the five feature maps. 
The multi-scale feature maps are then input to different prediction networks: first, a region proposal network (RPN) generates a list of candidate bounding boxes that should contain an \entity/. Second, a Region of Interest (RoI) alignment layer filters out the multi-scale feature maps that correspond to the candidate regions. We note that all 5 feature maps are used by the RPN, but $P_6$ is not included in the inputs to the RoI alignment layer. Third, for each region proposal, a mask sub-network predicts the segmentation masks, based on the RoI aligned features. These segmentation masks are not used in subsequent steps of \sys/ at prediction time; however, they are utilized in our loss function during the training process. Fourth, these bounding boxes are subsequently refined in a detection sub-network, thereby yielding the final bounding boxes $B$. It also provides the label for classifying the \entity/ category.

All of the above sub-networks were carefully adapted to the specific characteristics of our task: (1)~We modified the region proposal network so that it uses a maximum base aspect ratio of 1:8 per \entity/. The reason for this modification is that document \entities/ (as opposed to classical image segmentation) contain \entities/ that have highly rectangular shapes. This is the case for most \entities/, \eg/, single \textsc{content line} or \textsc{table row} \entities/. (2)~The output size of the classifier sub-network is modified so that it can produce predictions for \entities/ across all semantic categories $\mathcal{C}$. (3)~During training of the mask sub-network, we treat all pixels in ground truth bounding boxes as foreground. We do this to incorporate our understanding of the exact shape of many \entities/ that span very wide rectangular regions. (4)~We use a mask sub-network loss with a weighting factor of $0.5$. This is to prioritize that features relevant for the correct prediction of bounding boxes and \entity/ categories are learned.
The Mask R-CNN stage of \sys/ comprises \num{63891032} parameters and is built upon the implementation of Mask R-CNN provided by \citet{matterport_maskrcnn_2017}, yet which we carefully adapted as described above.

\noindent
\textbf{Training Procedure:} All neural models are initialized with pre-trained weights based on the MS COCO dataset \cite{Lin2014}. We then train each model across three phases for a total of \num{80000} iterations. This is split into three phases of \num{20000}, \num{40000}, and \num{20000} iterations, respectively. During the first phase, we freeze all layers of the CNN that is used as the initial block in Mask~R-CNN. In the second phase, stages four and five of the CNN are unfrozen. In the last phase, all network layers are trainable. Early stopping is applied based on the performance on the validation set for unrefined predictions. The performance is measured every \num{2000} iterations via the so-called intersection over union  with a threshold of $0.8$.

We train all models in a multi-GPU setting, using 8 GPUs with a vRAM of 12\,GB. Each GPU was fed with one image per training iteration. Accordingly, the batch size per training iteration is set to $8$. Furthermore, we use stochastic gradient descent with a learning rate of $0.001$ and learning momentum of $0.9$.

\noindent
\textbf{Parameter Settings:} During training, we sampled randomly 100 \entities/ from the ground truth per document image (\ie/, up to 100 \entities/ as some document images might have fewer). In Mask~R-CNN, the maximum number of \entity/ predictions per image is set to $200$. During prediction, we only keep \entities/ with a confidence score $P_j$ of $0.7$ or higher.

\noindent
\textbf{Weak Supervision}: Training with weak supervision is as follows: all models are initialized with the weights of our pre-trained \sys/~WS instead of default weights. We perform the training with learnable parameters analogous to phase 1 above but for \num{2000} steps with early stopping. In our experiments, we use only a subset of 80\,\% of the annotated documents from \datasetauto/, while the other 20\,\% remain unused. The intention is that we want to allow for additional annotations in the future while ensuring comparability to our results. We further ensure a fairly uniform distribution of \entities/ by utilizing only document pages that contain at least an \textsc{abstract}, a \textsc{figure}, or \textsc{table}, while all others are discarded. This amounts to \num{593583} pages.

\subsection{System Variants}

We compare the following variants of \sys/: \textbf{\sys/~Baseline} is trained solely on the noise-free labels provided for the training dataset (here: \datasetmanual/);
\textbf{\sys/~WS} benefits from weak supervision~(WS). It is trained based on a second dataset (here: \datasetauto/) with noisy labels for weak supervision. This is to test whether training systems on noisy labels can lead to higher performance, compared to training on small but noise-free training datasets;
\textbf{\sys/~WS+FT} is initialized with the weights from \sys/~WS, but then fine-tuned~(FT) on the target dataset.

\subsection{Performance Metrics}

We separately evaluate the performance of our system for (i)~detection of \entities/ $E_j$ and (ii)~classification of hierarchical \relations/ $R_j$. The former aims at a high detection rate (\ie/ recognizing true positives out of all positives). Hence, we use the average precision as evaluation metric. The latter is based on the F1 score as it represents a typical classification task (\ie/ recognizing one of the relations from $\Psi$). 

\noindent
\textbf{Entity Detection:}  
\entity/ detection is commonly measured by the mean average precision~(mAP) of a model (0: worst, 100: best).
The inferred \entities/ $E_j = (c_j, B_j, P_j)$ are compared against the ground truth label consisting of the true category $\hat{c}_j$ with a bounding box $\hat{B}_j$. Here we follow common practice in computer vision \citep{Everingham2012} and measure the overlap between bounding boxes from the same category. Specifically, we calculate the so-called intersection over union (IoU): $\text{IoU} = \frac{\mathit{area}(B_j \cap \hat{B}_j)}{\mathit{area}(B_j \cup \hat{B}_j)}$.
If the IoU is higher than a user-defined threshold, a predicted \entity/ is considered a true positive. If multiple \entities/ are matched with the same ground truth \entity/, we only consider the \entity/ with the highest IoU as a true positive. Unmatched predictions and ground truth \entities/ are considered false positives and false negatives, respectively. This is then used to calculate the average precision~(AP) per semantic category $C_k \in \mathcal{C}$. The overall performance across all categories is given by the mean average precision. We compare IoU thresholds of $0.5$ and $0.65$.\footnote{Additional results for IoU=0.8 are in the supplements.} 

\noindent
\textbf{Prediction of Hierarchical \Relations/:} 
Here we measure the classification performance for predicting the correct \relations/. A \relation/ $R = (E_{\text{subj}}, E_{\text{obj}}, \Psi)$ is counted as correct only if the complete tuple is identical. However, the performance depends on the correct \entity/ detection as input. Hence, we later vary the IoU thresholds for \entity/ detection analogous to above and then report the corresponding F1 score for correctly predicting hierarchical \relations/. The F1 score is the harmonic average of precision and recall for predicting these triples (0: worst, 1: best).

Note that our performance measure is relatively strict. We show that, even if some F1 scores are in a lower range, we can recover the overall document structure successfully. In particular, we outperform state-of-the-art OCR results, as illustrated in the qualitative samples in our supplements.

\subsection{Robustness Check: Table Structure Parsing} We additionally train our model for structure parsing so that it identifies table structures to demonstrate the robustness of our system and weak supervision.

We confirm the effectiveness of our weak supervision as follows: we draw upon the ICDAR 2013 dataset \citep{Gobel2013} for table structure parsing and compare it with the state-of-the-art. The ICDAR 2013 dataset consists of a variety of real-world documents and is not limited to scientific articles. We proceed analogously to full document structure parsing and train the three system variants for the task of table structure recognition.

\textbf{\sys/~Baseline} is trained solely on the samples provided in the ICDAR 2013 training dataset;
\textbf{\sys/~WS} is trained on table structures generated from \datasetauto/. \textbf{\sys/~WS+FT} is generated by subsequent fine-tuning on the ICDAR training split.\footnote{Details about the setting and additional experiments are provided in the supplements.} 
  
Both training and fine-tuning of all variants follow the 3 phase training scheme for a total of \num{80000} iterations.\footnote{Due to the different domain of the target dataset, we experimented with other weak supervision strategies, \eg/ randomly sampling images from \datasetauto/ and ICDAR 2013 during the same training procedure. However, the performance of models trained by sequential fine-tuning could not be surpassed.}

\section{Results}
\begin{table}[t!]
\small 
    \centering
\sisetup
  { 
    round-precision    = 1           ,
    round-integer-to-decimal,
    round-mode = places,
    table-align-text-post=false,
    table-number-alignment=center,
    table-format=2.1,
  }
\renewrobustcmd{\bfseries}{\fontseries{b}\selectfont}
\begingroup
\setlength{\tabcolsep}{2.25pt} 
    \begin{tabular}{@{}l  SSS  SSS@{}}
    \toprule
\multicolumn{1}{c}{} &    \multicolumn{3}{c}{IoU=0.5} & \multicolumn{3}{c}{IoU=0.65}\\
\cmidrule(lr){2-4} \cmidrule(lr){5-7}
\multicolumn{1}{@{}l}{AP}   & \multicolumn{1}{c}{Baseline} & \multicolumn{1}{c}{WS} & \multicolumn{1}{c}{WS+FT}& \multicolumn{1}{c}{Baseline} & \multicolumn{1}{c}{WS} & \multicolumn{1}{c}{WS+FT}\\
\midrule
          mean AP &           49.93 &           34.59 &  \bfseries 69.35 &           38.53 &           32.41 &  \bfseries 56.54 \\
\midrule
\textsc{     abstract }&  \bfseries 95.24 &           90.48 &           95.16 &           90.48 &           81.01 &  \bfseries 95.24 \\
\textsc{  affiliation }&  \bfseries 51.62 &               0 &           46.02 &            5.92 &               0 &   \bfseries 16.2 \\
\textsc{       author }&           17.95 &               0 &  \bfseries 23.61 &  \bfseries 20.37 &               0 &            16.7 \\
\textsc{   bib. block }&           42.42 &           79.09 &   \bfseries 94.7 &           43.18 &  \bfseries 93.94 &           80.26 \\
\textsc{  cont. block }&  \bfseries 89.31 &           69.75 &           88.41 &           83.17 &           67.03 &  \bfseries 84.38 \\
\textsc{         date }&               0 &               0 &  \bfseries 24.07 &               0 &               0 &   \bfseries 9.26 \\
\textsc{     equation }&           65.84 &           54.53 &  \bfseries 82.05 &            40.6 &           52.14 &   \bfseries 72.8 \\
\textsc{ fig. caption }&           47.77 &            30.5 &  \bfseries 69.23 &           43.95 &           17.73 &  \bfseries 59.54 \\
\textsc{ fig. graphic }&           22.28 &            5.19 &  \bfseries 60.21 &           15.93 &            4.36 &   \bfseries 54.5 \\
\textsc{       figure }&           47.82 &           35.28 &  \bfseries 63.52 &           43.96 &           33.94 &  \bfseries 59.39 \\
\textsc{       footer }&            55.7 &               0 &  \bfseries 69.26 &           48.86 &               0 &  \bfseries 59.68 \\
\textsc{       header }&           79.69 &               0 &  \bfseries 88.28 &  \bfseries 64.84 &               0 &           56.56 \\
\textsc{      heading }&           53.74 &            52.1 &  \bfseries 66.35 &           33.11 &  \bfseries 46.04 &           45.39 \\
\textsc{         item }&               0 &           33.57 &  \bfseries 50.49 &               0 &  \bfseries 35.26 &           33.47 \\
\textsc{      itemize }&               0 &              25 &  \bfseries 58.33 &               0 &              25 &   \bfseries 50.0 \\
\textsc{     keywords }&           36.36 &               0 &  \bfseries 58.98 &           36.36 &               0 &  \bfseries 42.95 \\
\textsc{     page nr. }&           74.72 &               0 &  \bfseries 77.31 &           28.53 &               0 &  \bfseries 42.04 \\
\textsc{ tab. caption }&           55.18 &           69.11 &  \bfseries 76.64 &           40.16 &           61.56 &  \bfseries 63.42 \\
\textsc{        table }&           84.47 &  \bfseries 96.33 &           94.31 &           62.67 &           87.85 &  \bfseries 89.62 \\
\textsc{      tabular }&           78.41 &           50.82 &  \bfseries 99.98 &           68.44 &            42.4 &  \bfseries 99.45 \\
\bottomrule
    \end{tabular}
    \endgroup
\caption{Average precision (AP) of \entity/ detection.}
    \label{tab:entity_results}
\end{table}

The key focus of our experiments is to confirm the effectiveness of \sys/ for parsing the complete document structures. However, we emphasize again that both suitable baselines and datasets for this task are hitherto lacking. Hence, we proceed two-fold. On the one hand, we evaluate the performance based on \dataset/ as the first dataset for document structure parsing. On the other hand, we draw upon the table structure ICDAR 2013 dataset: it is limited to table structures and not complete holistic parsing of document structures. However, it allows to test the effectiveness of our weak supervision against state-of-the-art.

\subsection{Document Structure Parsing}

We compare the performance of document structure parsing based on our \datasetmanual/ dataset across both performance metrics.

\subsubsection{Entity Detection}

The overall performance for \entity/ detection is detailed in Table~\ref{tab:entity_results} (first row). We discuss the performance for $\text{IoU}=0.5$ in the following. \sys/~Baseline achieves an mAP of 49.9. This is higher than \sys/~WS with an mAP of 34.6. We attribute this to the fact that several \entity/ categories from \datasetmanual/ are not part of \datasetauto/. Notably, the fine-tuned system \sys/~WS+FT results in significant performance improvements: it obtains a mAP of 69.4, which, in comparison to the baseline \sys/, is an improvement by $39.1\,\%$. 

\sys/~WS+FT consistently outperforms the baseline system, even for categories that are not annotated during weak supervision (\eg/ \textsc{author}, \textsc{footer}, \textsc{header}, \textsc{page number}). We attribute this to the better model initialization due to the prior weakly supervised pre-training. There is a small number of entity categories for which the Baseline achieves higher AP values. We attribute this to our experimental protocol which yields the best model via early stopping, based on mAP and not on individual entity AP values. 
For a few entities a decrease can be observed after fine-tuning (\eg/ \textsc{table} at IoU=0.5). We attribute this to the high quality of weak annotations for this category and, consequently, a slight decrease of generalization due to fine-tuning. Some AP values (for both \sys/~Baseline and \sys/~WS) amount to 0.0, \eg/ for \textsc{date}. This is caused by the absence of some categories in \datasetauto/ in the case of \sys/~WS. For \sys/~Baseline, we attribute this to the limited amount of samples in \datasetmanual/ for the affected categories, coupled with an inferior model initialization, compared to \sys/~WS+FT.

\sys/~WS+FT outperforms the \sys/~Baseline system across all measured IoU thresholds by a considerable margin. Using IoU thresholds above 0.5 leads to a performance decrease. Even though higher IoUs should generally correspond to better matches with the ground truth, they can penalize ambiguous cases and thus a correct detection. In sum, this confirms the effectiveness of our weak supervision in bolstering the overall performance.

Table~\ref{tab:entity_results} breaks down the performance by \entity/ category. For \sys/~WS+FT, we observe an especially good performance for detecting tabulars and figures. This is owed to the strong initialization of our system due to the high quality and large number of samples in our scalable weak supervision.\footnote{For a few \entities/, the best performance is achieved a combination of the WS system together with a high IoU (\eg/, \textsc{bibliography block}). A likely reason for this is the composition of \datasetmanual/. As \textsc{bibliography} \entities/ were not specifically used as a criterion for the per-page sampling, fewer documents in the target dataset contained relevant \entities/, leading to decreased performance of the baseline and WS+FT systems.}

\pgfplotstableread[col sep=semicolon]{
model;0;10;20;30;60;100;160
test;34.59;48.22;52.87;58.21;60.24;63.8;69.35
}\highleveldetectionFT

\pgfplotstabletranspose[string type,
    colnames from=model,
    input colnames to=model
]\highleveldetectionFTtransp{\highleveldetectionFT}
\begin{figure}[t!]
    \centering
\begin{adjustbox}{width=\linewidth}
\begin{tikzpicture}
  \begin{axis}[
    width=\textwidth,
    height=.35\textwidth,
      xlabel={Number of Fine-tuning Images},
      ylabel={mAP [\%]},
        legend style={draw=none},
        ymajorgrids = true,
        xmajorgrids = true,
        axis line style = {ultra thin, ETH6},
        enlarge x limits={0.05},
        legend style={at={(1,0.35)},
        legend columns=-1},
      tick label style = {font=\large\sansmath\sffamily},
      every axis label = {font=\large\sansmath\sffamily},
      legend style = {font=\large\sansmath\sffamily},
      label style = {font=\large\sansmath\sffamily},
      every axis plot/.append style={ultra thick}
  ]
     \addplot[mark=diamond*, ETH7, mark options={solid}]  table[x=model,y=test]{\highleveldetectionFTtransp};
     \addlegendentry{\sys/~WS+FT};
     \addplot[mark=*, mark options={solid}, dashed, ETH3, domain=0:160, samples=5] {34.59};
    \addlegendentry{\sys/~WS};
     \addplot[mark=square*, mark options={solid},black, dashed, domain=0:160, samples=5] {49.93};
    \addlegendentry{Baseline \sys/};
  \end{axis}
\end{tikzpicture}
\end{adjustbox}
\caption{Performance of \entity/ detection (mAP for $\text{IoU}=0.5$) during fine-tuning.}
\label{fig:subsampling_finetuning_highlevel}
\end{figure}
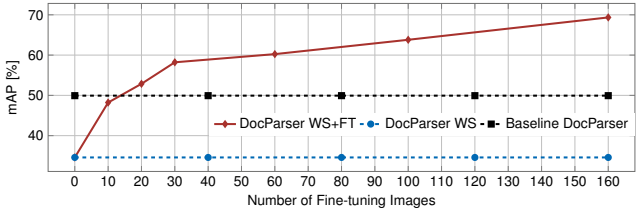

Figure~\ref{fig:subsampling_finetuning_highlevel} shows the fine-tuning. Only $20$ fine-tuning samples are sufficient for \sys/~WS+FT to surpass the baseline system \sys/ (which is trained on 160 samples from the target dataset). It thus helps in reducing the labeling effort by a factor of around 8. Furthermore, we observe a steady increase in the performance of the fine-tuned networks with more samples. Notably, the highest performance increase is already achieved by the first 10 document images for fine-tuning.

\subsubsection{Prediction of Hierarchical \Relations/}

\begin{table}[t!]
\small
\sisetup
  { 
    round-precision    = 3           ,
    round-integer-to-decimal,
    round-mode = places,
    table-number-alignment=center,
    table-format=1.3,
  }
\renewrobustcmd{\bfseries}{\fontseries{b}\selectfont}
    \centering
\begingroup
\setlength{\tabcolsep}{1.25pt} 
\begin{tabular}{@{}l  SSS  SSS@{}}
    \toprule
 &    \multicolumn{3}{c}{IoU=0.5} & \multicolumn{3}{c}{IoU=0.65} \\
\cmidrule(lr){2-4} \cmidrule(lr){5-7} 
  &{Baseline}&{WS} & {WS+FT}&{Baseline}&{WS}&{WS+FT}\\
\midrule
All &    0.4155 &  0.3425 &  \bfseries 0.5037 &    0.3215 &  0.3181 &  \bfseries 0.4454 \\
\textit{followed\_by} &  0.4130 &  0.3865 &  \bfseries 0.5056 &    0.3135 &  0.3659 &  \bfseries 0.4466 \\
\textit{parent\_of} &   0.4208 &  0.2348 &  \bfseries 0.4999 &    0.3388 &  0.1981 &  \bfseries 0.4428 \\
\midrule
\textbf{Refined}: & & & & & &\\
All  &    0.4533 &  0.3815 &  \bfseries 0.6153 &    0.3633 &  0.3544 &  \bfseries 0.5577 \\
   \textit{followed\_by}  &   0.4545 &  0.4101 &  \bfseries 0.5807 &    0.3510 &  0.3936 &  \bfseries 0.5235 \\
\textit{parent\_of}  &  0.4508 &  0.3169 &  \bfseries 0.6785 &    0.3887 &  0.2626 &  \bfseries 0.6203 \\
\bottomrule
    \end{tabular}
    \endgroup
\caption{Performance in predicting hierarchical \relations/ (as measured by F1).}
    \label{tab:parsing_results}
\end{table}

Table~\ref{tab:parsing_results} compares the classification of \relations/ with and without post-processing. The best performance (across all $\Psi$) is achieved by \sys/~WS+FT with an IoU of 0.5: it registers an F1 score of 0.615. Here, the use of weak supervision with fine-tuning yields consistent improvements. This is also due to the significant improvements of the prior entity detection for this system variant. In particular, for an IoU of 0.5, it outperforms the F1 score of the baseline system (F1 of 0.453) by 0.162. This amounts to a relative improvement of 35.8\,\%. Evidently, a smaller IoU threshold of 0.5 is beneficial. Higher IoU thresholds reduce the overall parsing performance as structure parsing builds on the prior detection of document \entities/. 

The performance on hierarchical relations (F1 score of 0.615) is largely explained by our choice of a strict evaluation (\ie/ the complete tuple including both entities must be correct). Overall, this performance is already highly effective in recovering the overall document structure. This is later confirmed as part of a qualitative assessment.

\subsection{Robustness Check: Table Structure Parsing}

\begin{table}[t!]
\small
\centering
\sisetup{detect-weight,mode=text, round-integer-to-decimal}
\renewrobustcmd{\bfseries}{\fontseries{b}\selectfont}
\begin{tabular}{l cccc}
\toprule
System & \multicolumn{1}{c}{\makecell[c]{Schreiber et\\ al. (2018)}} & \multicolumn{1}{c}{Baseline} & \multicolumn{1}{c}{WS} & \multicolumn{1}{c}{WS+FT} \\
\midrule
F1* & 0.9144 & 0.8443 & 0.8117 &  \bfseries 0.9292 \\
F1 &  --- & 0.8209 & 0.8056 & \bfseries 0.9292 \\
\bottomrule

\end{tabular}
\caption{ICDAR 2013 result on table structure parsing.}
\begin{minipage}{\linewidth}
\footnotesize{\emph{Notes:} Evaluation of image-based systems on ``ICDAR~50\,\%'', which uses a random subset containing 50\,\% of the competition set for testing. \citet{Schreiber2018} use a different, non-public 50\,\% random subset. Furthermore, \citet{Schreiber2018} choose the best system based on the test set as indicated by F1*. In contrast, F1 refers to the performance when the selection is based on the validation set.}
\end{minipage}

\label{tab:icdar_results_overview}
\end{table}
\noindent
\textbf{Results:} Table~\ref{tab:icdar_results_overview} compares the state-of-the-art for table structure parsing with our weak supervision strategy. Altogether, our weak supervision outperforms the state-of-the-art \cite{Schreiber2018} by a considerable margin.

\noindent
\textbf{Discussion:} Our system shows significant improvement over the image-based state of the art. We also compare our approach to the state-of-the-art heuristic-based system that operates on raw PDF files, instead of images, as input \cite{Nurminen2013}. Even though our system does not utilize the additional information provided by raw PDF files, \sys/ achieves an F1 score of 0.9292, compared to 0.9221 for the PDF-based system. We refrain from directly comparing the aforementioned F1 score with that from earlier experiments as the underlying target domains differ.

\section{Related Work}
\noindent
\textbf{OCR:} Extracting text from document images has been extensively studied as part of optical character recognition~(OCR) within the NLP community \cite[\eg/,][]{Schafer2011, Schafer2012}. To this end, the work by \citet{Katti2018} argued that OCR should be seen as a preprocessing step for downstream NLP tasks. As such, the authors extract text-based information but not the hierarchical document structure as in our research.

\noindent
\textbf{Table Detection:} Document renderings are commonly used for the task of table detection (rather than table structure parsing). Here, the objective is to predict the bounding boxes of tables, \ie/, whether a pixel refers to a table or not \citep[\eg/,][]{Yildiz2005,Wang2004}. Prior research on table detection has utilized data augmentation \citep{Gilani2017}, weak supervision \citep{li2019tablebank}, and transfer learning \citep[\eg/,][]{Siddiqui2018} to address the lack of large-scale domain-specific datasets. Similar to our research, efficient learning presents an issue for table detection. However, parsing of full pages requires effective identification of a much larger number of \entities/ of multiple categories and high variety in shape per input.

\noindent
\textbf{Table Structure Parsing:} There are works that recognize table structures from text or other syntactic tokens \citep{Kieninger1998,Pivk2007} rather than directly from document renderings. As such, these works are tailored to tokens as input, and it is thus unclear how such an approach could theoretically be adapted to document renderings since our task inherently relies upon images as input. Because of the different input and thus the different datasets for benchmarking, the performance of the aforementioned works is not comparable to our approach. The works by \citet{Schreiber2018, qasim2019rethinking} draw upon deep neural networks to identify table structures for rendered inputs. However, they aim at a different purpose: parsing table structures, but not complete document hierarchies. As such, the authors do not attempt to identify text elements, nested figures, etc.

\noindent
\textbf{Weak Supervision for Document Layout:} \citep{zhong2019publaynet} use weak supervision for detection of page layout entities. The WS mechanism relies on matching external XML annotations with text extractions by a heuristic-based third-party tool. In contrast, our weak supervision directly builds on the \LaTeX compilation and can be readily extended to any new dataset of \LaTeX source files. Furthermore, the dataset features only 5 coarse categories and the system does not feature a relation classification component, thus being insufficient to acquire full document structures.\footnote{Additional comparison is included in the supplements.} 

\noindent
\textbf{Weak Supervision in NLP:} Annotations in NLP are oftentimes costly and, as a result, there has been a recent surge in weak supervision. Weak supervision has now been applied to various tasks, such as text classification \citep[\eg/,][]{Hingmire2014,Lin2011}, information extraction \citep[\eg/,][]{Hoffmann2011}, and semantic parsing \citep[\eg/,][]{Goldman2018}. The methodological levers for obtaining weak labels are versatile and include, \eg/, manual rules \citep[\eg/,][]{Rabinovich2018}, estimated models \citep[\eg/,][]{Hoffmann2011}, or reinforcement learning \citep{Prollochs2019}; however, not for document structure parsing.

\section{Discussion and Conclusion}

\noindent
\textbf{Efficiency:} Our system requires only $\sim$340\,ms/document during entity detection (averaged over our validation set of 79 documents for \sys/~WS+FT) on a single Titan Xp GPU with 12 GB VRAM and a batch size of 1. The relation detection in stage 2 only adds a minimal overhead of an average of 5.67\,ms/document (10.81\,ms/document with refinement) on a single CPU @ 2.1 GHz.

\noindent
\textbf{Qualitative Assessment:} We performed a qualitative analysis on a subset of documents. We observe that, even for F1 scores below 0.5, the final document structure is often still very accurate. In fact, state-of-the-art OCR systems as natural baselines are outperformed significantly. This can be explained by our experiment design: we used very strict evaluation metrics. Hence, even small mismatches or ambiguities between the ground truth and predicted entities result in fairly large F1 penalties, despite high overall similarity. Details are in the supplements (including qualitative examples).  

\noindent
\textbf{Detection Model Choice:} Deep CNN models, including recent work \cite{tan2019efficientnet, duan2019centernet}, are heavily reliant on large training datasets. As such, we expect the impact of our technical contribution, as shown in our comparison of baseline and WS+FT models, to be the same across different modern CNN backbones. Our choice of Mask R-CNN as a tool for instance segmentation was also done in consideration of possible future extensions of \sys/ to non-rectified documents. Here, the additional instance masks could guide the OCR or rectification process. 

\noindent
\textbf{Future Work:}
In future work, we plan to explore approaches that can jointly learn entity and relation detection.
Furthermore, we aim to further improve our system by enriching 2D inputs with textual features, \eg/ high-dimensional word embeddings. 
The robustness of WS pretraining w.r.t. smaller subsets of \datasetauto/ is another area of future investigation.

\noindent
\textbf{Conclusion:} Despite the extensive interest of the NLP community in leveraging document structures \cite[\eg/,][]{apostolova2014combining,Schafer2011, Schafer2012, Schreiber2018, Katti2018}, the task of parsing complete document structures from renderings has been overlooked. To the best of our knowledge, we present the first system for this task. In particular, \sys/ provides an effective alternative to state-of-the-art OCR which is still widespread in practice. In addition, \sys/ allows to provide additional semantic input to downstream NLP tasks (\eg/ information extraction).

\section{ Acknowledgments}
Ce Zhang and the DS3Lab gratefully acknowledge the support from the Swiss National Science Foundation (Project Number 200021\_184628), Innosuisse/SNF BRIDGE Discovery (Project Number 40B2-0\_187132), European Union Horizon 2020 Research and Innovation Programme (DAPHNE, 957407), Botnar Research Centre for Child Health, Swiss Data Science Center, Alibaba, Cisco, eBay, Google Focused Research Awards, Oracle Labs, Swisscom, Zurich Insurance, Chinese Scholarship Council, and the Department of Computer Science at ETH Zurich.

\bibliography{references}

\begin{thebibliography}{43}
\providecommand{\natexlab}[1]{#1}
\providecommand{\url}[1]{\texttt{#1}}
\providecommand{\urlprefix}{URL }
\expandafter\ifx\csname urlstyle\endcsname\relax
  \providecommand{\doi}[1]{doi:\discretionary{}{}{}#1}\else
  \providecommand{\doi}{doi:\discretionary{}{}{}\begingroup
  \urlstyle{rm}\Url}\fi

\bibitem[{Abdulla(2017)}]{matterport_maskrcnn_2017}
Abdulla, W. 2017.
\newblock {Mask R-CNN for Object Detection and Instance Segmentation on Keras
  and TensorFlow}.

\bibitem[{Antonacopoulos et~al.(2009)Antonacopoulos, Bridson, Papadopoulos, and
  Pletschacher}]{Antonacopoulos2009}
Antonacopoulos, A.; Bridson, D.; Papadopoulos, C.; and Pletschacher, S. 2009.
\newblock {A Realistic Dataset for Performance Evaluation of Document Layout
  Analysis}.
\newblock In \emph{International Conference on Document Analysis and
  Recognition (ICDAR)}.
\newblock ISBN 9780769537252.
\newblock ISSN 15205363.

\bibitem[{Apostolova and Tomuro(2014)}]{apostolova2014combining}
Apostolova, E.; and Tomuro, N. 2014.
\newblock Combining visual and textual features for information extraction from
  online flyers.
\newblock In \emph{Proceedings of the 2014 Conference on Empirical Methods in
  Natural Language Processing (EMNLP)}, 1924--1929.

\bibitem[{Arif and Shafait(2018)}]{Arif2019}
Arif, S.; and Shafait, F. 2018.
\newblock {Table Detection in Document Images using Foreground and Background
  Features}.
\newblock In \emph{2018 Digital Image Computing: Techniques and Applications
  (DICTA)}.
\newblock ISBN 978-1-5386-6602-9.

\bibitem[{Chen, Tsai, and Tsai(2000)}]{10.3115/990820.990845}
Chen, H.-H.; Tsai, S.-C.; and Tsai, J.-H. 2000.
\newblock Mining Tables from Large Scale HTML Texts.
\newblock In \emph{Proceedings of the 18th Conference on Computational
  Linguistics - Volume 1}, COLING ’00, 166–172. USA: Association for
  Computational Linguistics.
\newblock ISBN 155860717X.

\bibitem[{Duan et~al.(2019)Duan, Bai, Xie, Qi, Huang, and
  Tian}]{duan2019centernet}
Duan, K.; Bai, S.; Xie, L.; Qi, H.; Huang, Q.; and Tian, Q. 2019.
\newblock Centernet: Keypoint triplets for object detection.
\newblock In \emph{Proceedings of the IEEE International Conference on Computer
  Vision}, 6569--6578.

\bibitem[{Embley et~al.(2006)Embley, Hurst, Lopresti, and Nagy}]{Embley2006}
Embley, D.~W.; Hurst, M.; Lopresti, D.; and Nagy, G. 2006.
\newblock {Table-processing Paradigms: A Research Survey}.

\bibitem[{Everingham et~al.(2010)Everingham, Van~Gool, Williams, Winn, and
  Zisserman}]{Everingham2012}
Everingham, M.; Van~Gool, L.; Williams, C.~K.; Winn, J.; and Zisserman, A.
  2010.
\newblock The pascal visual object classes (voc) challenge.
\newblock \emph{International Journal of Computer Vision} 88(2): 303--338.

\bibitem[{Garncarek et~al.(2020)Garncarek, Powalski, Stanis{\l}awek, Topolski,
  Halama, and Grali{\'n}ski}]{garncarek2020lambert}
Garncarek, {\L}.; Powalski, R.; Stanis{\l}awek, T.; Topolski, B.; Halama, P.;
  and Grali{\'n}ski, F. 2020.
\newblock LAMBERT: Layout-Aware language Modeling using BERT for information
  extraction.
\newblock \emph{arXiv preprint arXiv:2002.08087} .

\bibitem[{Gilani et~al.(2017)Gilani, Qasim, Malik, and Shafait}]{Gilani2017}
Gilani, A.; Qasim, S.~R.; Malik, I.; and Shafait, F. 2017.
\newblock Table Detection using Deep Learning.
\newblock In \emph{14th IAPR International Conference on Document Analysis and
  Recognition (ICDAR)}.

\bibitem[{Gobel et~al.(2013)Gobel, Hassan, Oro, and Orsi}]{Gobel2013}
Gobel, M.; Hassan, T.; Oro, E.; and Orsi, G. 2013.
\newblock {ICDAR 2013 Table Competition}.
\newblock In \emph{International Conference on Document Analysis and
  Recognition (ICDAR)}.
\newblock ISBN 978-0-7695-4999-6.
\newblock ISSN 15205363.

\bibitem[{Goldman et~al.(2018)Goldman, Latcinnik, Nave, Globerson, and
  Berant}]{Goldman2018}
Goldman, O.; Latcinnik, V.; Nave, E.; Globerson, A.; and Berant, J. 2018.
\newblock {Weakly Supervised Semantic Parsing with Abstract Examples}.
\newblock In \emph{Annual Meeting of the Association for Computational
  Linguistics (ACL)}.

\bibitem[{Govindaraju, Zhang, and
  R{\'e}(2013)}]{govindaraju-etal-2013-understanding}
Govindaraju, V.; Zhang, C.; and R{\'e}, C. 2013.
\newblock Understanding Tables in Context Using Standard {NLP} Toolkits.
\newblock In \emph{Proceedings of the 51st Annual Meeting of the Association
  for Computational Linguistics (Volume 2: Short Papers)}, 658--664. Sofia,
  Bulgaria: Association for Computational Linguistics.

\bibitem[{He et~al.(2017)He, Gkioxari, Doll{\'{a}}r, and Girshick}]{He2017}
He, K.; Gkioxari, G.; Doll{\'{a}}r, P.; and Girshick, R. 2017.
\newblock {Mask R-CNN}.
\newblock In \emph{IEEE International Conference on Computer Vision (ICCV)}.
\newblock ISBN 978-1-5386-0457-1.
\newblock ISSN 0006-291X.

\bibitem[{He et~al.(2016)He, Zhang, Ren, and Sun}]{He2016}
He, K.; Zhang, X.; Ren, S.; and Sun, J. 2016.
\newblock {Deep Residual Learning for Image Recognition}.
\newblock In \emph{IEEE Conference on Computer Vision and Pattern Recognition
  (CVPR)}.

\bibitem[{Hingmire and Chakraborti(2014)}]{Hingmire2014}
Hingmire, S.; and Chakraborti, S. 2014.
\newblock {Sprinkling Topics For Weakly Supervised Text Classification}.
\newblock In \emph{Annual Meeting of the ACL}.

\bibitem[{Hoffmann et~al.(2011)Hoffmann, Zhang, Ling, Zettlemoyer, and
  Weld}]{Hoffmann2011}
Hoffmann, R.; Zhang, C.; Ling, X.; Zettlemoyer, L.; and Weld, D.~S. 2011.
\newblock {Knowledge-based Weak Supervision for Information Extraction of
  Overlapping Relations}.
\newblock In \emph{Annual Meeting of the ACL}.

\bibitem[{Hurst and Nasukawa(2000)}]{hurst-nasukawa-2000-layout}
Hurst, M.; and Nasukawa, T. 2000.
\newblock Layout and Language: Integrating Spatial and Linguistic Knowledge for
  Layout Understanding Tasks.
\newblock In \emph{{COLING} 2000 Volume 1: The 18th International Conference on
  Computational Linguistics}.

\bibitem[{Katti et~al.(2018)Katti, Reisswig, Guder, Brarda, Bickel, H{\"o}hne,
  and Faddoul}]{Katti2018}
Katti, A.~R.; Reisswig, C.; Guder, C.; Brarda, S.; Bickel, S.; H{\"o}hne, J.;
  and Faddoul, J.~B. 2018.
\newblock Chargrid: Towards Understanding 2D Documents.
\newblock In \emph{Conference on Empirical Methods in Natural Language
  Processing (EMNLP)}.

\bibitem[{Kieninger and Dengel(1998)}]{Kieninger1998}
Kieninger, T.; and Dengel, A. 1998.
\newblock The T-Recs Table Recognition and Analysis System.
\newblock In \emph{International Workshop on Document Analysis Systems (DAS)}.

\bibitem[{Laurens(2008)}]{laurens2008direct}
Laurens, J. 2008.
\newblock {Direct and reverse synchronization with SyncTEX}.
\newblock \emph{TUGBoat} 29: 365--371.

\bibitem[{Li et~al.(2019)Li, Cui, Huang, Wei, Zhou, and Li}]{li2019tablebank}
Li, M.; Cui, L.; Huang, S.; Wei, F.; Zhou, M.; and Li, Z. 2019.
\newblock TableBank: Table Benchmark for Image-based Table Detection and
  Recognition.
\newblock \emph{arXiv preprint arXiv:1903.01949} .

\bibitem[{Lin, He, and {and Everson Richard}(2011)}]{Lin2011}
Lin, C.; He, Y.; and {and Everson Richard}. 2011.
\newblock {Sentence Subjectivity Detection With Weakly-Supervised Learning}.
\newblock In \emph{International Joint Conference on Natural Language
  Processing (IJCNLP)}.

\bibitem[{Lin et~al.(2017)Lin, Doll{\'{a}}r, Girshick, He, Hariharan, and
  Belongie}]{Lin2017}
Lin, T.~Y.; Doll{\'{a}}r, P.; Girshick, R.; He, K.; Hariharan, B.; and
  Belongie, S. 2017.
\newblock {Feature Pyramid Networks for Object Detection}.
\newblock In \emph{IEEE Conference on Computer Vision and Pattern Recognition
  (CVPR)}.
\newblock ISBN 9781538604571.

\bibitem[{Lin et~al.(2014)Lin, Maire, Belongie, Hays, Perona, Ramanan,
  Doll{\'{a}}r, and Zitnick}]{Lin2014}
Lin, T.~Y.; Maire, M.; Belongie, S.; Hays, J.; Perona, P.; Ramanan, D.;
  Doll{\'{a}}r, P.; and Zitnick, C.~L. 2014.
\newblock {Microsoft COCO: Common Objects in Context}.
\newblock In \emph{European Conference on Computer Vision (ECCV)}.
\newblock ISBN 978-3-319-10601-4.
\newblock ISSN 16113349.

\bibitem[{Liu et~al.(2019)Liu, Gao, Zhang, and Zhao}]{liu2019graph}
Liu, X.; Gao, F.; Zhang, Q.; and Zhao, H. 2019.
\newblock Graph Convolution for Multimodal Information Extraction from Visually
  Rich Documents.
\newblock In \emph{Proceedings of the 2019 Conference of the North American
  Chapter of the Association for Computational Linguistics: Human Language
  Technologies, Volume 2 (Industry Papers)}, 32--39.

\bibitem[{Luong, Nguyen, and Kan(2012)}]{Luong2011}
Luong, M.-T.; Nguyen, T.~D.; and Kan, M.-Y. 2012.
\newblock Logical Structure Recovery in Scholarly Articles with Rich Document
  Features.
\newblock In \emph{Multimedia Storage and Retrieval Innovations for Digital
  Library Systems}, 270--292. IGI Global.

\bibitem[{Nurminen(2013)}]{Nurminen2013}
Nurminen, A. 2013.
\newblock \emph{Algorithmic Extraction of Data in Tables in PDF Documents}.
\newblock Master's thesis, Tampere University of Technology.

\bibitem[{Pivk et~al.(2007)Pivk, Cimiano, Sure, Gams, Rajkovi{\v{c}}, and
  Studer}]{Pivk2007}
Pivk, A.; Cimiano, P.; Sure, Y.; Gams, M.; Rajkovi{\v{c}}, V.; and Studer, R.
  2007.
\newblock {Transforming Arbitrary Tables into Logical Form with TARTAR}.
\newblock \emph{Data and Knowledge Engineering} 567--595.
\newblock ISSN 0169023X.

\bibitem[{Pr{\"{o}}llochs, Feuerriegel, and Neumann(2019)}]{Prollochs2019}
Pr{\"{o}}llochs, N.; Feuerriegel, S.; and Neumann, D. 2019.
\newblock {Learning Interpretable Negation Rules via Weak Supervision at
  Document Level: A Reinforcement Learning Approach}.
\newblock In \emph{NAACL-HLT}.

\bibitem[{Qasim, Mahmood, and Shafait(2019)}]{qasim2019rethinking}
Qasim, S.~R.; Mahmood, H.; and Shafait, F. 2019.
\newblock Rethinking table recognition using graph neural networks.
\newblock In \emph{2019 International Conference on Document Analysis and
  Recognition (ICDAR)}, 142--147. IEEE.

\bibitem[{Rabinovich et~al.(2018)Rabinovich, Sznajder, Spector, Shnayderman,
  Aharonov, Konopnicki, and Slonim}]{Rabinovich2018}
Rabinovich, E.; Sznajder, B.; Spector, A.; Shnayderman, I.; Aharonov, R.;
  Konopnicki, D.; and Slonim, N. 2018.
\newblock {Learning Concept Abstractness using Weak Supervision}.
\newblock In \emph{EMNLP}.

\bibitem[{Rice, Jenkins, and Nartker(1995)}]{Rice1995}
Rice, S.~V.; Jenkins, F.~R.; and Nartker, T.~A. 1995.
\newblock The Fourth Annual Test of OCR Accuracy.
\newblock Technical report, Technical Report 95.

\bibitem[{Sch{\"a}fer et~al.(2011)Sch{\"a}fer, Kiefer, Spurk, Steffen, and
  Wang}]{Schafer2011}
Sch{\"a}fer, U.; Kiefer, B.; Spurk, C.; Steffen, J.; and Wang, R. 2011.
\newblock The ACL Anthology Searchbench.
\newblock In \emph{49th Annual Meeting of the Association for Computational
  Linguistics: Human Language Technologies: Systems Demonstrations (ACL-HLT)}.
  Association for Computational Linguistics.

\bibitem[{Sch\"{a}fer and Weitz(2012)}]{Schafer2012}
Sch\"{a}fer, U.; and Weitz, B. 2012.
\newblock Combining OCR Outputs for Logical Document Structure Markup:
  Technical Background to the ACL 2012 Contributed Task.
\newblock In \emph{ACL-2012 Special Workshop on Rediscovering 50 Years of
  Discoveries}, ACL '12.

\bibitem[{Schreiber et~al.(2018)Schreiber, Agne, Wolf, Dengel, and
  Ahmed}]{Schreiber2018}
Schreiber, S.; Agne, S.; Wolf, I.; Dengel, A.; and Ahmed, S. 2018.
\newblock {DeepDeSRT: Deep Learning for Detection and Structure Recognition of
  Tables in Document Images}.
\newblock In \emph{International Conference on Document Analysis and
  Recognition (ICDAR)}.
\newblock ISBN 9781538635865.
\newblock ISSN 15205363.

\bibitem[{Siddiqui et~al.(2018)Siddiqui, Malik, Agne, Dengel, and
  Ahmed}]{Siddiqui2018}
Siddiqui, S.~A.; Malik, M.~I.; Agne, S.; Dengel, A.; and Ahmed, S. 2018.
\newblock {DeCNT: Deep Deformable CNN for Table Detection}.
\newblock \emph{IEEE Access} 74151--74161.
\newblock ISSN 21693536.

\bibitem[{Tan and Le(2019)}]{tan2019efficientnet}
Tan, M.; and Le, Q.~V. 2019.
\newblock Efficientnet: Rethinking model scaling for convolutional neural
  networks.
\newblock \emph{arXiv preprint arXiv:1905.11946} .

\bibitem[{Tengli, Yang, and Ma(2004)}]{tengli-etal-2004-learning}
Tengli, A.; Yang, Y.; and Ma, N.~L. 2004.
\newblock Learning Table Extraction from Examples.
\newblock In \emph{{COLING} 2004: Proceedings of the 20th International
  Conference on Computational Linguistics}, 987--993. Geneva, Switzerland:
  COLING.

\bibitem[{Wang, Phillips, and Haralick(2004)}]{Wang2004}
Wang, Y.; Phillips, I.~T.; and Haralick, R.~M. 2004.
\newblock {Table Structure Understanding and its Performance Evaluation}.
\newblock \emph{Pattern Recognition} 1479--1497.
\newblock ISSN 00313203.

\bibitem[{Yildiz, Kaiser, and Miksch(2005)}]{Yildiz2005}
Yildiz, B.; Kaiser, K.; and Miksch, S. 2005.
\newblock {pdf2table: A Method to Extract Table Information from PDF Files}.
\newblock \emph{2nd Indian International Conference on Artificial Intelligence
  (IICAI)} .

\bibitem[{Zanibbi, Blostein, and Cordy(2004)}]{Zanibbi2004}
Zanibbi, R.; Blostein, D.; and Cordy, J. 2004.
\newblock {A Survey of Table Recognition}.
\newblock \emph{Document Analysis and Recognition} 1--33.
\newblock ISSN 1433-2833.

\bibitem[{Zhong, Tang, and Yepes(2019)}]{zhong2019publaynet}
Zhong, X.; Tang, J.; and Yepes, A.~J. 2019.
\newblock Publaynet: largest dataset ever for document layout analysis.
\newblock In \emph{2019 International Conference on Document Analysis and
  Recognition (ICDAR)}, 1015--1022. IEEE.

\end{thebibliography}

\pgfplotstableread[col sep=semicolon]{
model;0;10;20;30;60;87
test;92.04;93.05;94.01;93.74;94.1;94.04
}\lowleveldetectionFT
\pgfplotstabletranspose[string type,
    colnames from=model,
    input colnames to=model
]\lowleveldetectionFTtransp{\lowleveldetectionFT}
\begin{appendices}
\setcounter{page}{1}

\section{Performance of Document Structure Parsing}
\label{app:qualitative_eval}
\subsection{Qualitative Evaluation}
\begin{figure*}[t!]
\centering
\begin{subfigure}[b]{0.24\linewidth}
    \centering
    \adjincludegraphics[width=\linewidth,trim={.08\width} {.05\height} {0.08\width} {.05\height}, clip]{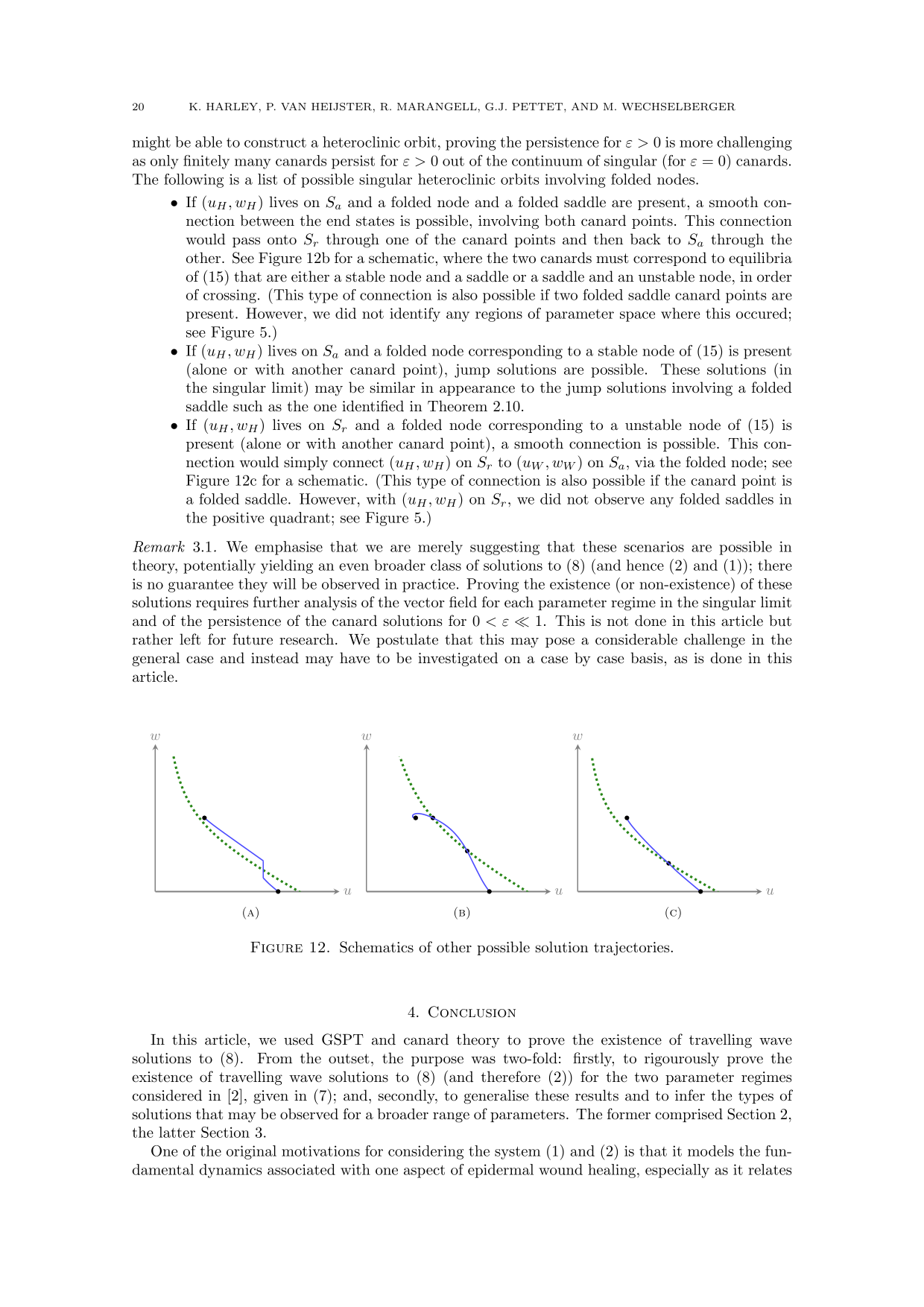}
\end{subfigure}
\begin{subfigure}[b]{0.24\linewidth}
    \centering
    \adjincludegraphics[width=\linewidth,trim={.08\width} {.05\height} {0.08\width} {.05\height}, clip]{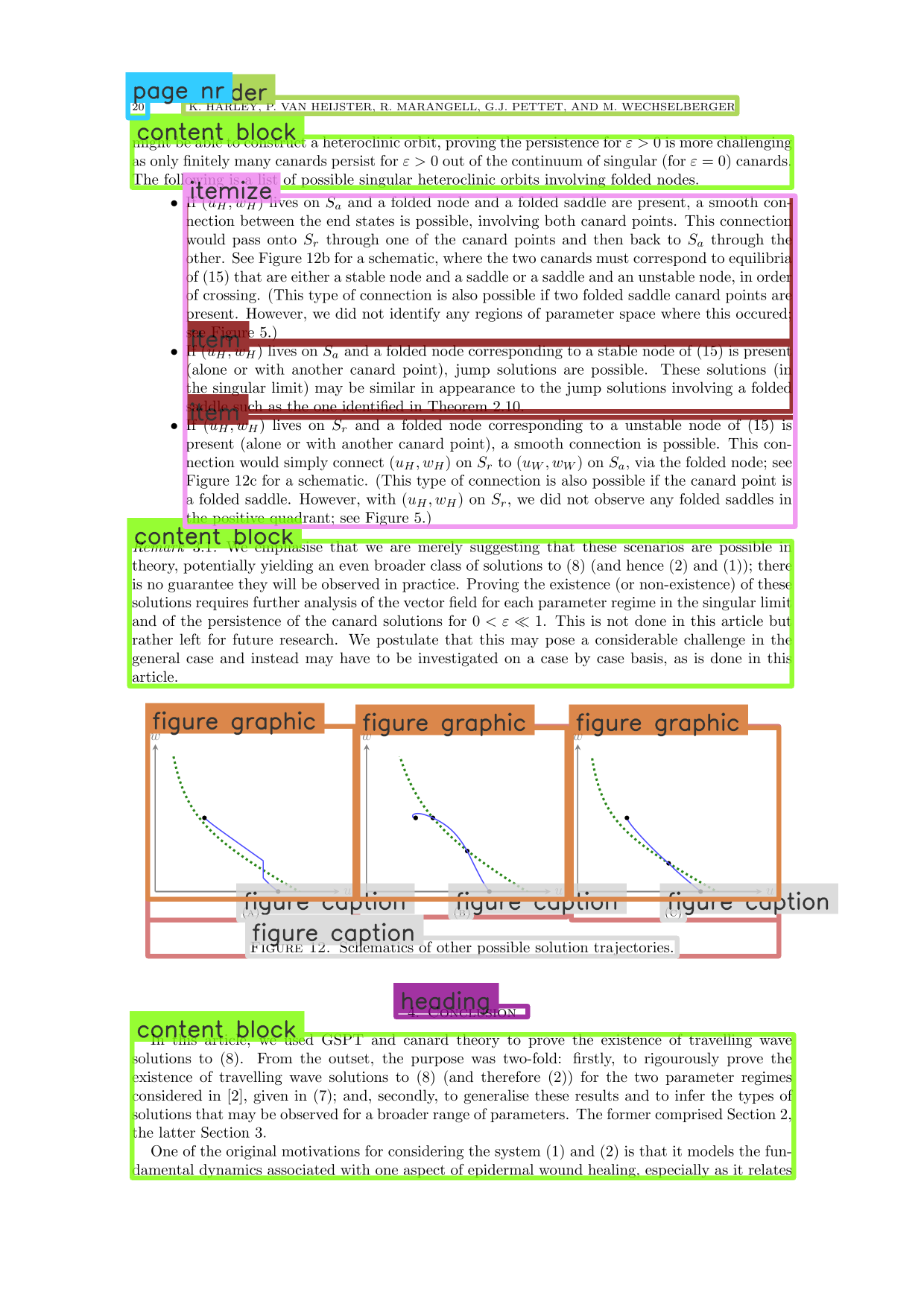}
\end{subfigure}
\begin{subfigure}[b]{0.24\linewidth}
    \centering
    \adjincludegraphics[width=\linewidth,trim={.08\width} {.05\height} {0.08\width} {.05\height},clip]{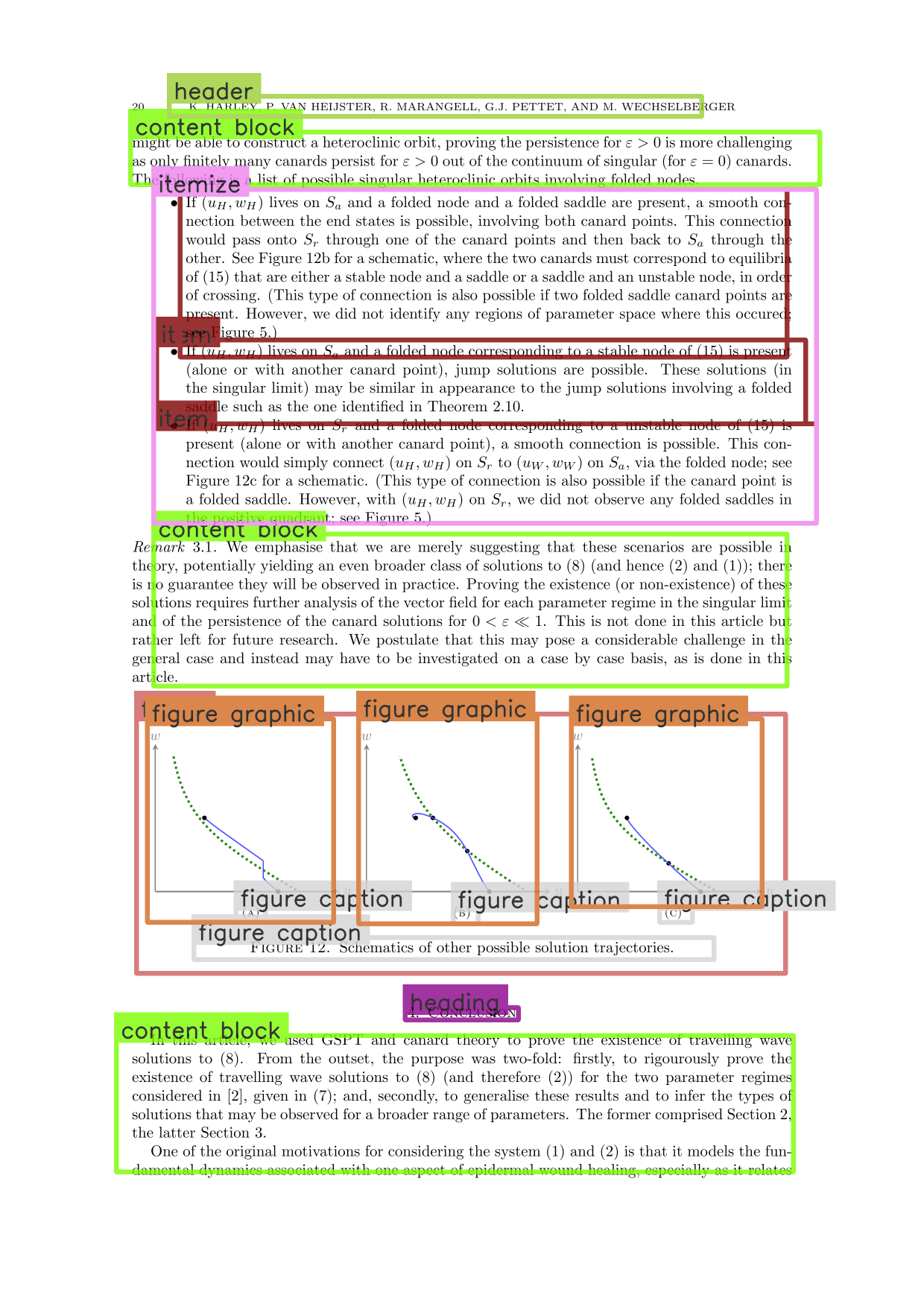}
\end{subfigure}
\begin{subfigure}[b]{0.24\linewidth}
    \centering
    \includegraphics[width=0.85\linewidth]{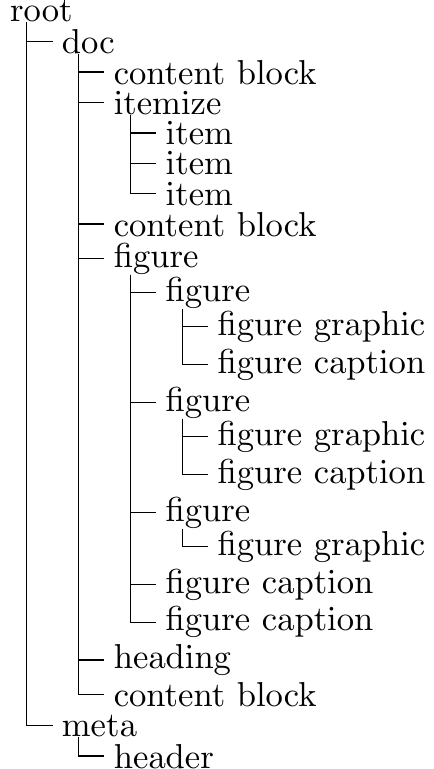}
    \end{subfigure}
\begin{subfigure}[b]{0.24\linewidth}
    \centering
    \adjincludegraphics[width=\linewidth,trim={.08\width} {.12\height} {0.08\width} {.03\height}, clip]{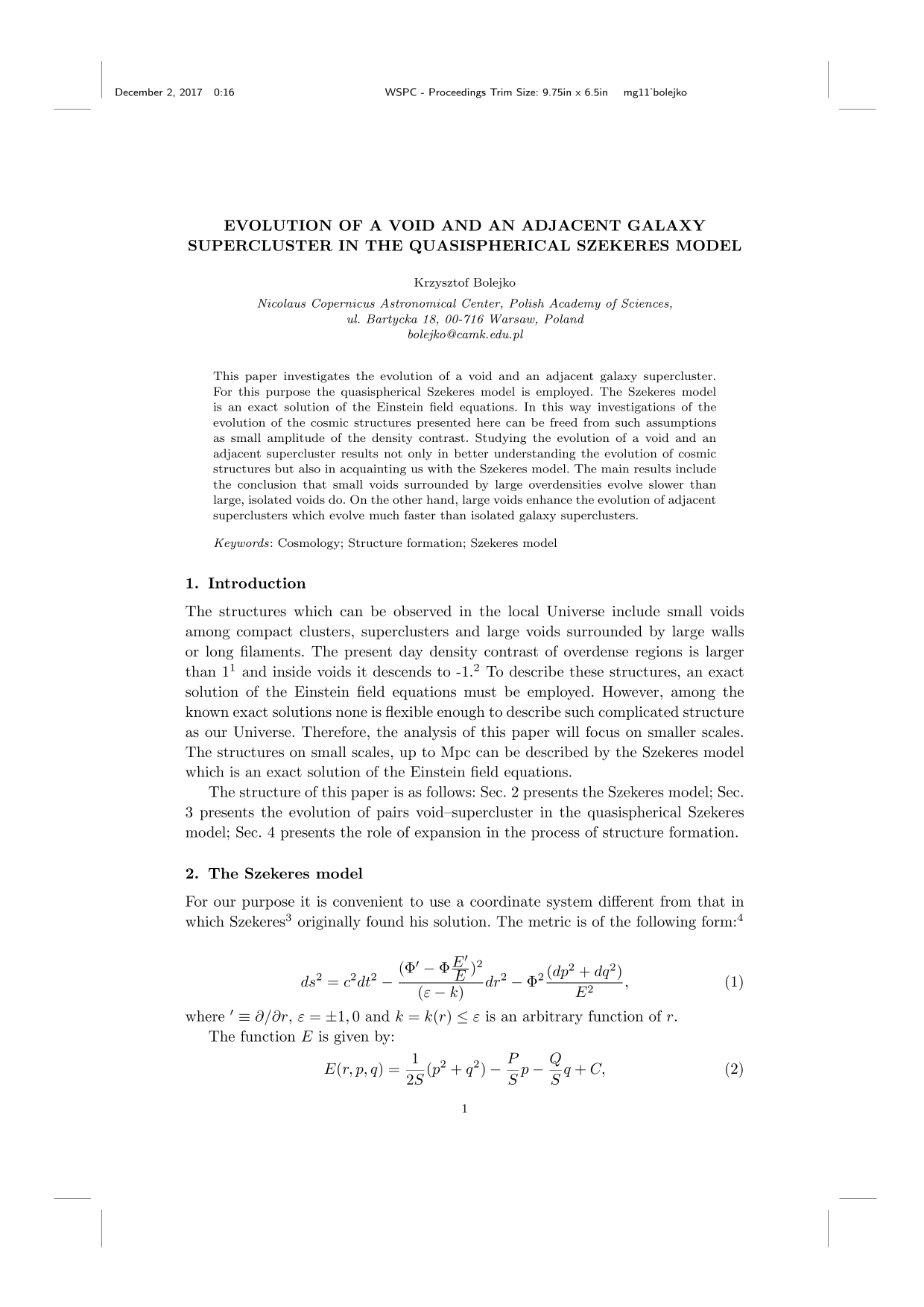}
    \caption{Input}
\end{subfigure}
\begin{subfigure}[b]{0.24\linewidth}
    \centering
    \adjincludegraphics[width=\linewidth,trim={.08\width} {.12\height} {0.08\width} {.03\height},clip]{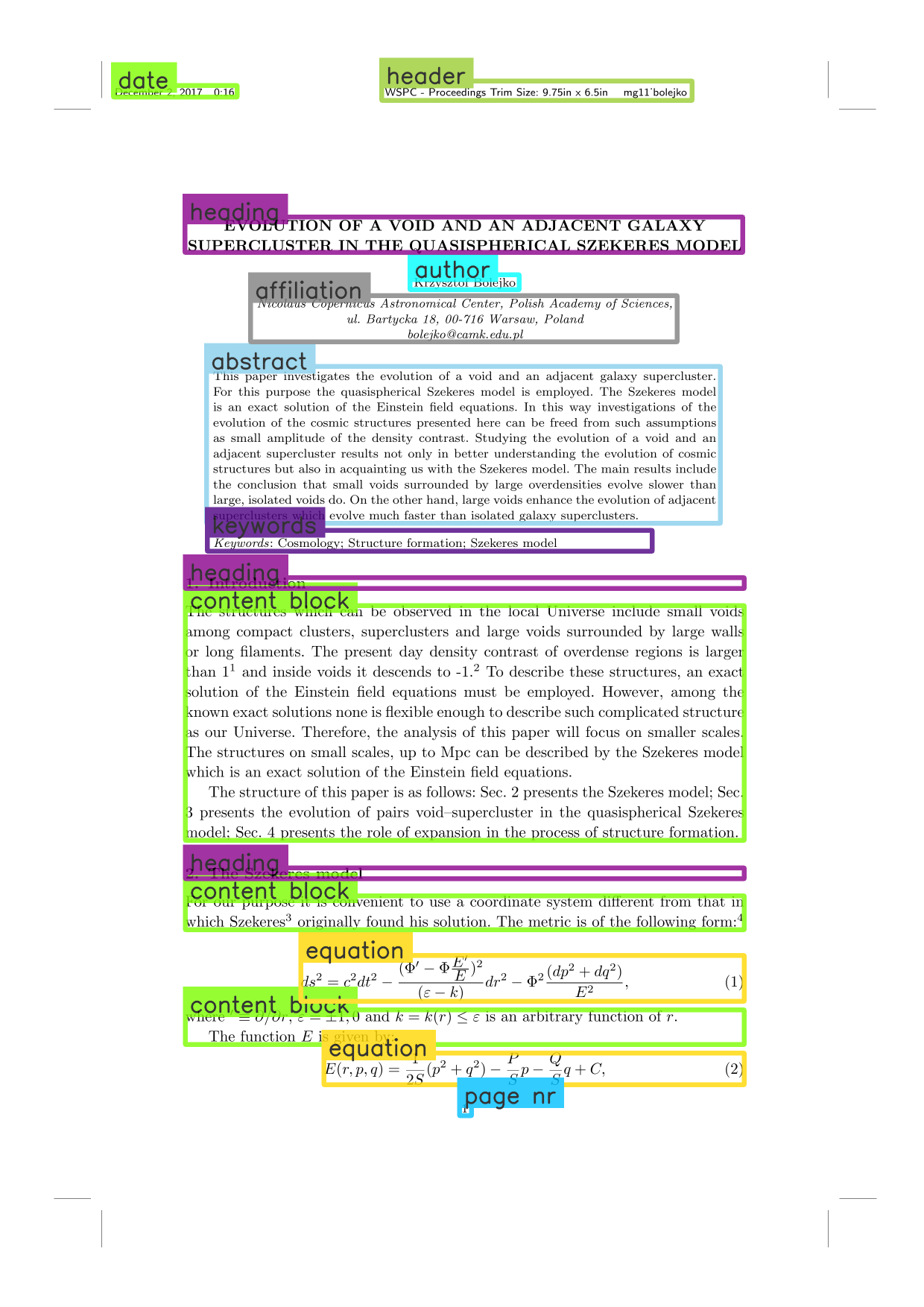}
    \caption{Ground Truth}
\end{subfigure}
\begin{subfigure}[b]{0.24\linewidth}
    \centering
    \adjincludegraphics[width=\linewidth,trim={.08\width} {.12\height} {0.08\width} {.03\height},clip]{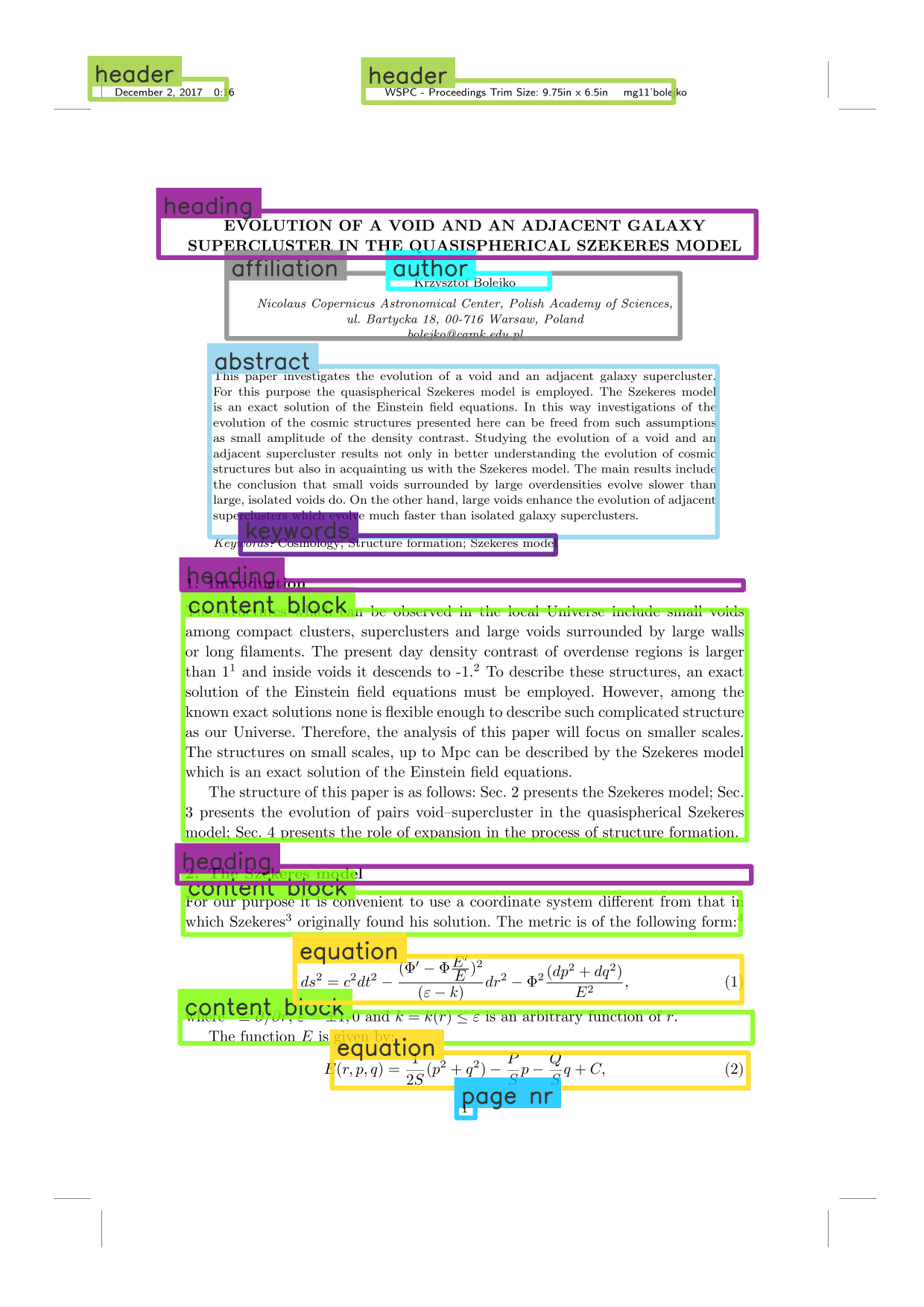}
    \caption{Predicted \entities/}
\end{subfigure}
\begin{subfigure}[b]{0.24\linewidth}
    \centering
    \includegraphics[width=0.65\linewidth]{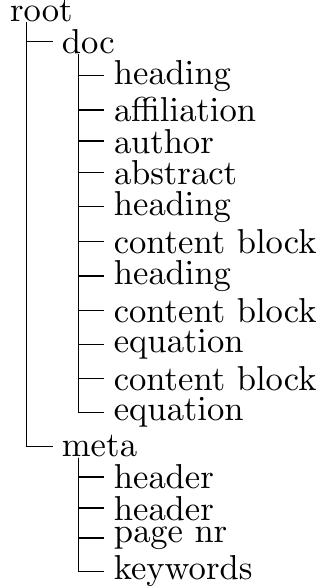}
        \caption{Predicted structure}
    \end{subfigure}
    \caption{Qualitative results of \sys/ for two samples (top and bottom rows).}
    \label{fig:qualitativeeval}
\end{figure*}

Figure \ref{fig:qualitativeeval} shows examples of parsed page structures that are generated by \sys/~WS+FT. 

\begin{figure}[t!]
\centering
\begin{subfigure}[b]{0.3\linewidth}
    \centering
    \adjincludegraphics[width=0.95\linewidth,trim={.0\width} {.0\height} {0.0\width} {.05\height}, clip]{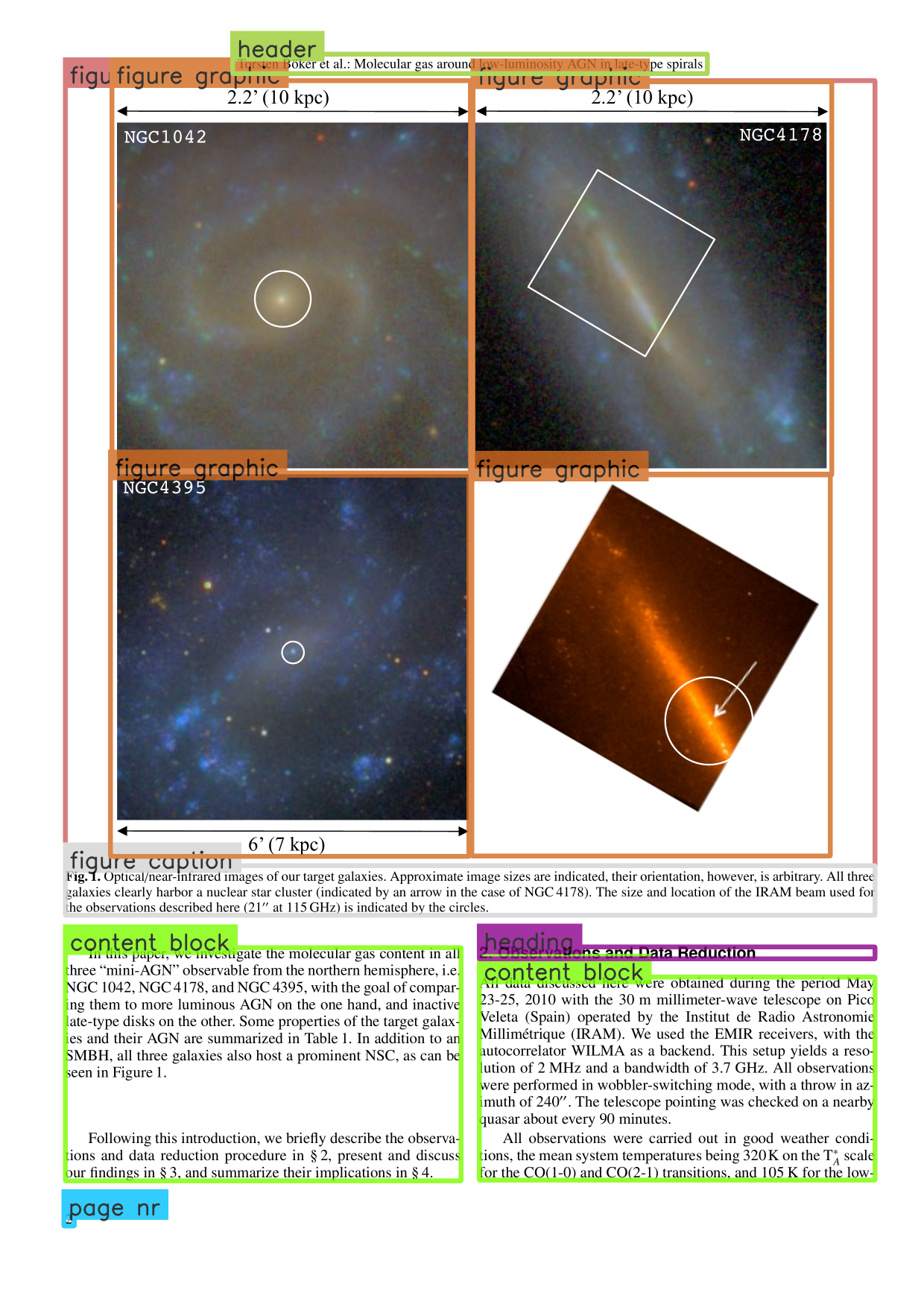}
\end{subfigure}
\begin{subfigure}[b]{0.3\linewidth}
    \centering
    \adjincludegraphics[width=0.95\linewidth,trim={.0\width} {.0\height} {0.0\width} {.05\height},clip]{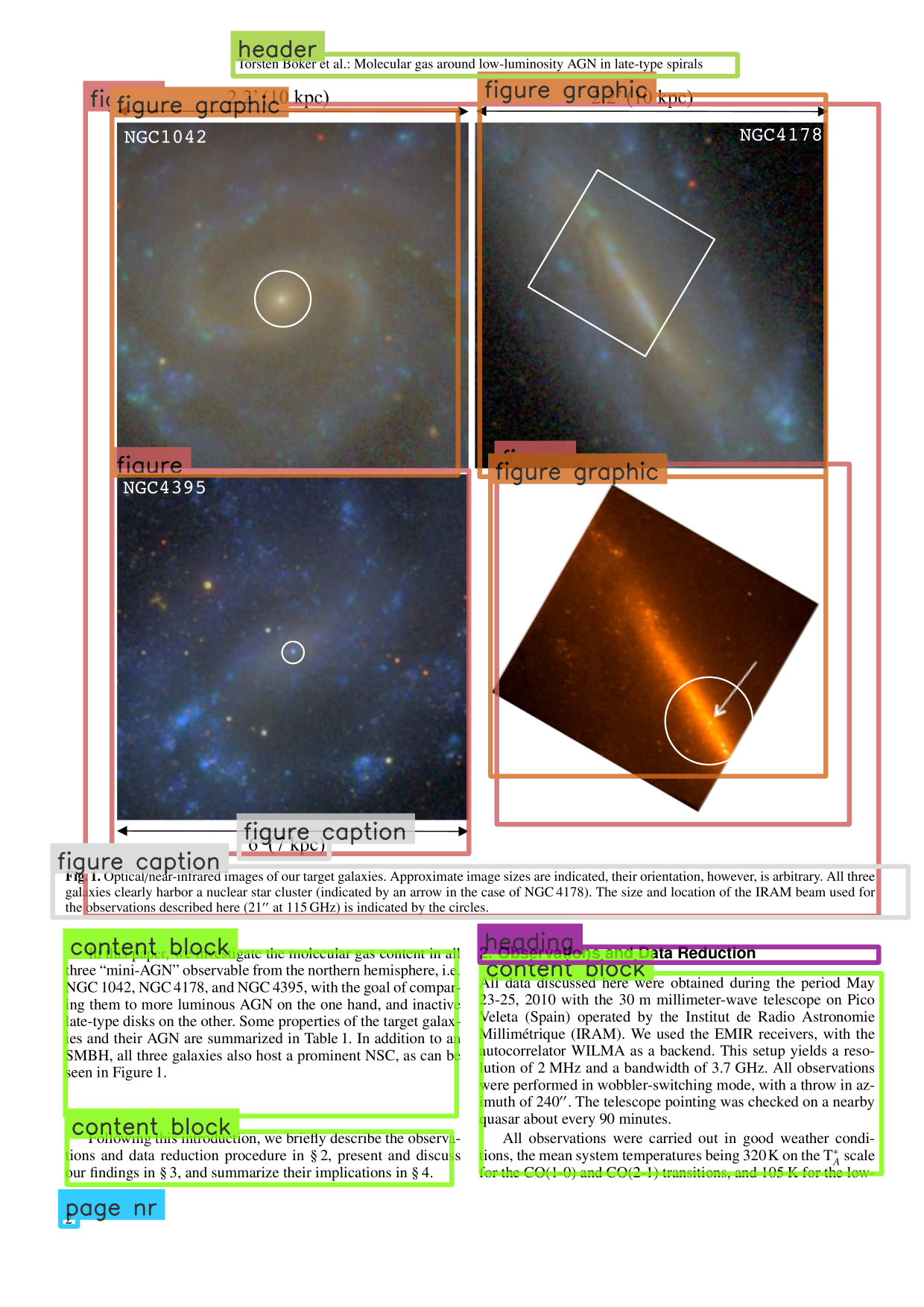}
\end{subfigure}
\begin{subfigure}[b]{0.3\linewidth}
    \centering
    \adjincludegraphics[width=0.95\linewidth,trim={.0\width} {.0\height} {0.0\width} {.05\height},clip]{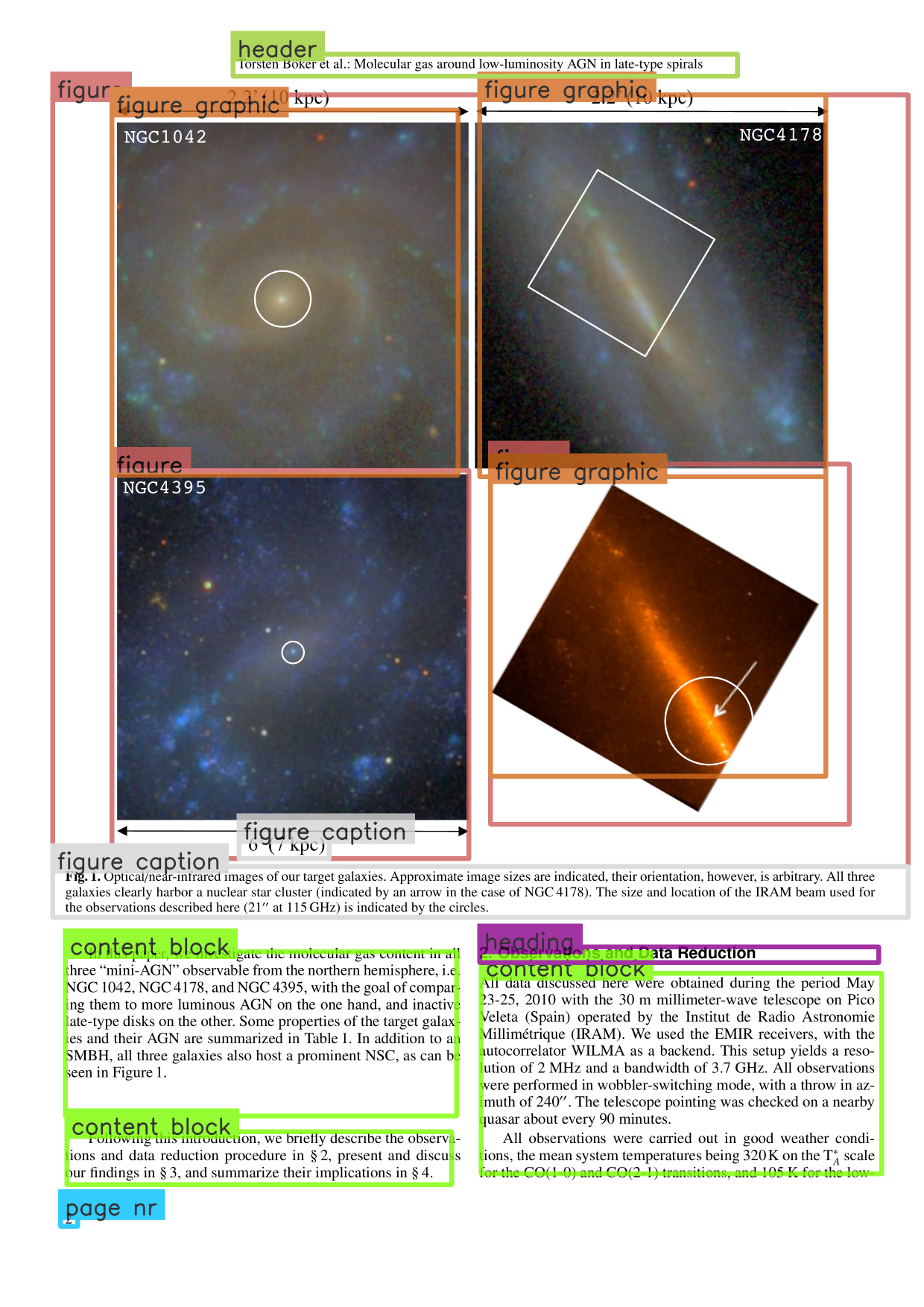}
\end{subfigure}
\begin{subfigure}[b]{0.3\linewidth}
    \centering
    \includegraphics[width=\linewidth]{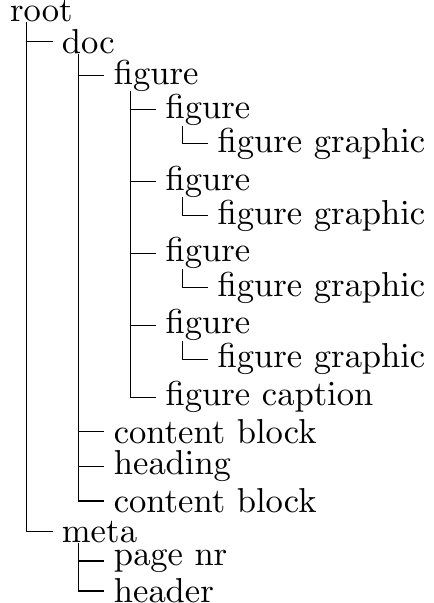}
    \caption{Ground truth}
\end{subfigure}
\begin{subfigure}[b]{0.3\linewidth}
    \centering
    \includegraphics[width=\linewidth]{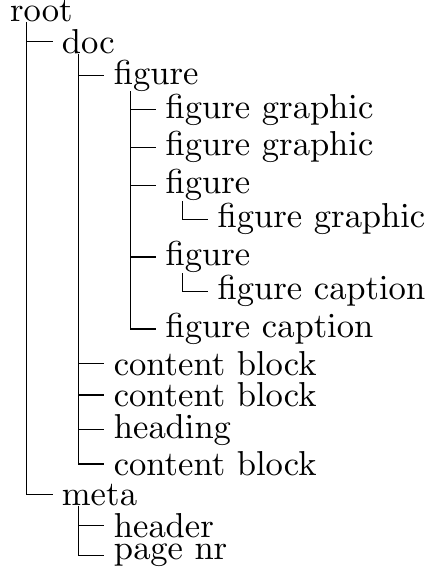}
    \caption{Raw pred.}
\end{subfigure}
\begin{subfigure}[b]{0.3\linewidth}
    \centering
    \includegraphics[width=\linewidth]{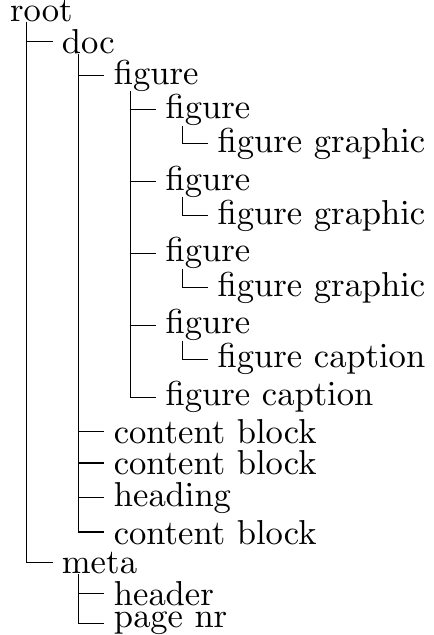}
    \caption{Predictions}
\end{subfigure}
    \caption{Raw predictions and structure-based refinement in \sys/.}
    \label{fig:structure_refinement1}
\end{figure}

\begin{figure}[t!]
\centering
\begin{subfigure}[b]{0.3\linewidth}
    \centering
    \adjincludegraphics[width=0.95\linewidth,trim={.12\width} {.05\height} {0.12\width} {.05\height}, clip]{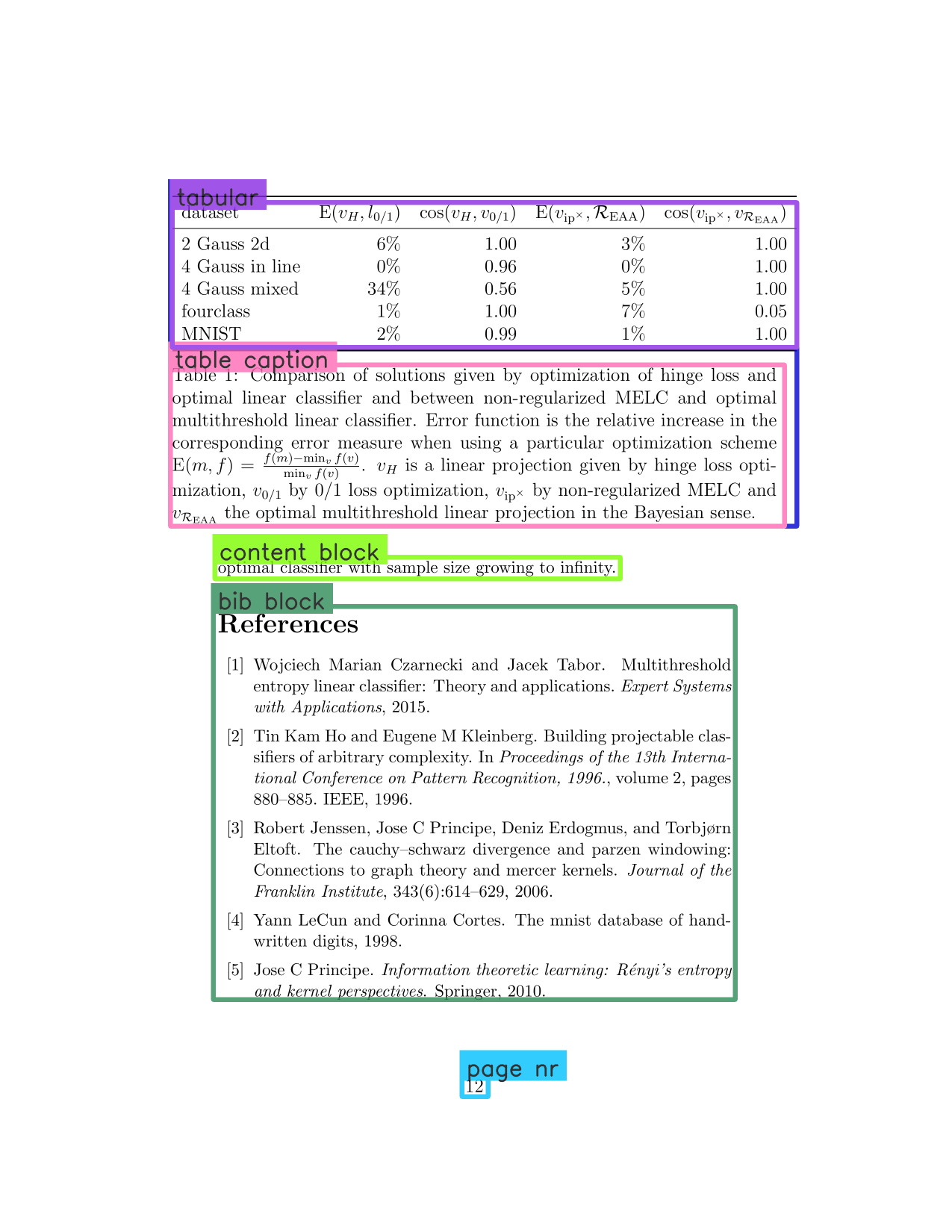}
\end{subfigure}
\begin{subfigure}[b]{0.3\linewidth}
    \centering
    \adjincludegraphics[width=0.95\linewidth,trim={.12\width} {.05\height} {0.12\width} {.05\height},clip]{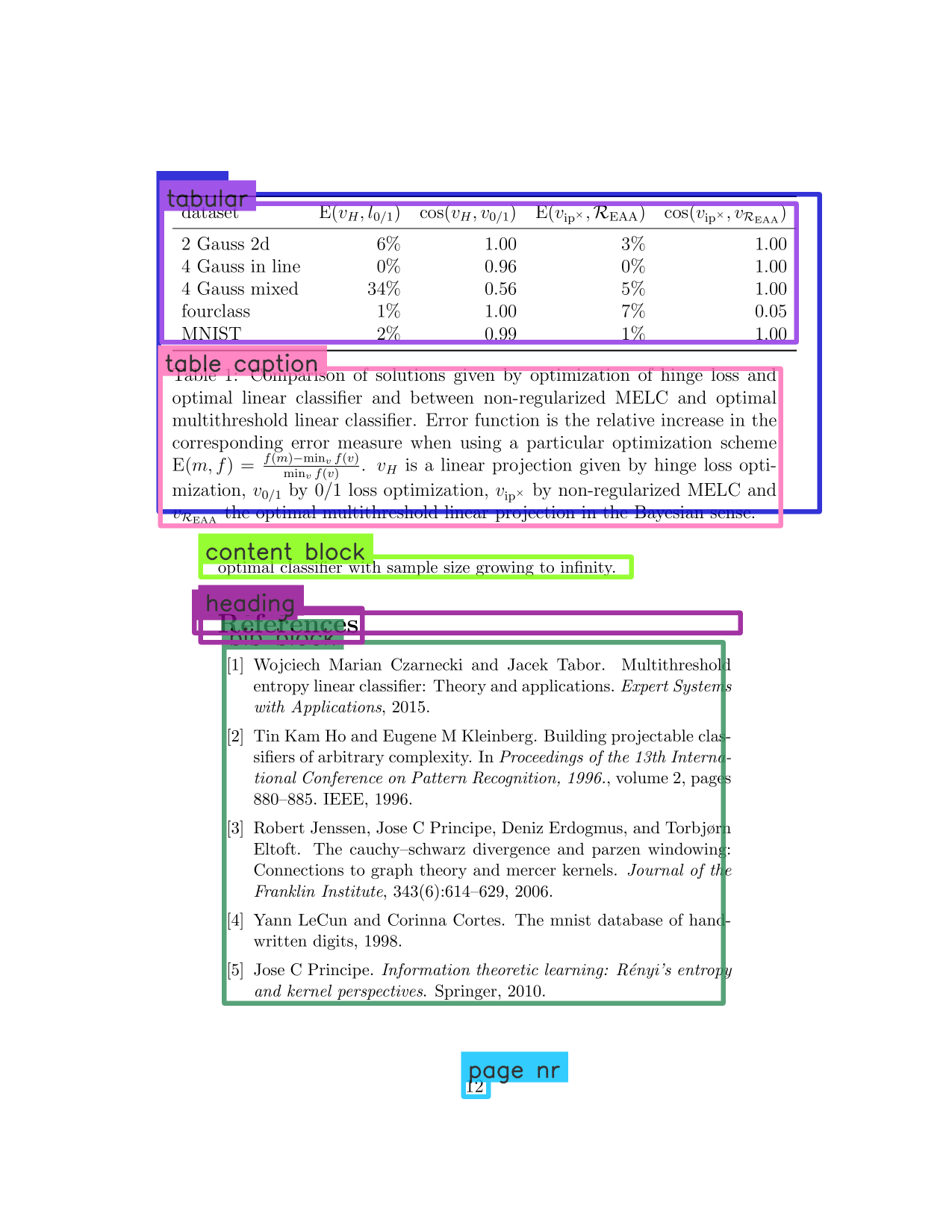}
\end{subfigure}
\begin{subfigure}[b]{0.3\linewidth}
    \centering
    \adjincludegraphics[width=0.95\linewidth,trim={.12\width} {.05\height} {0.12\width} {.05\height},clip]{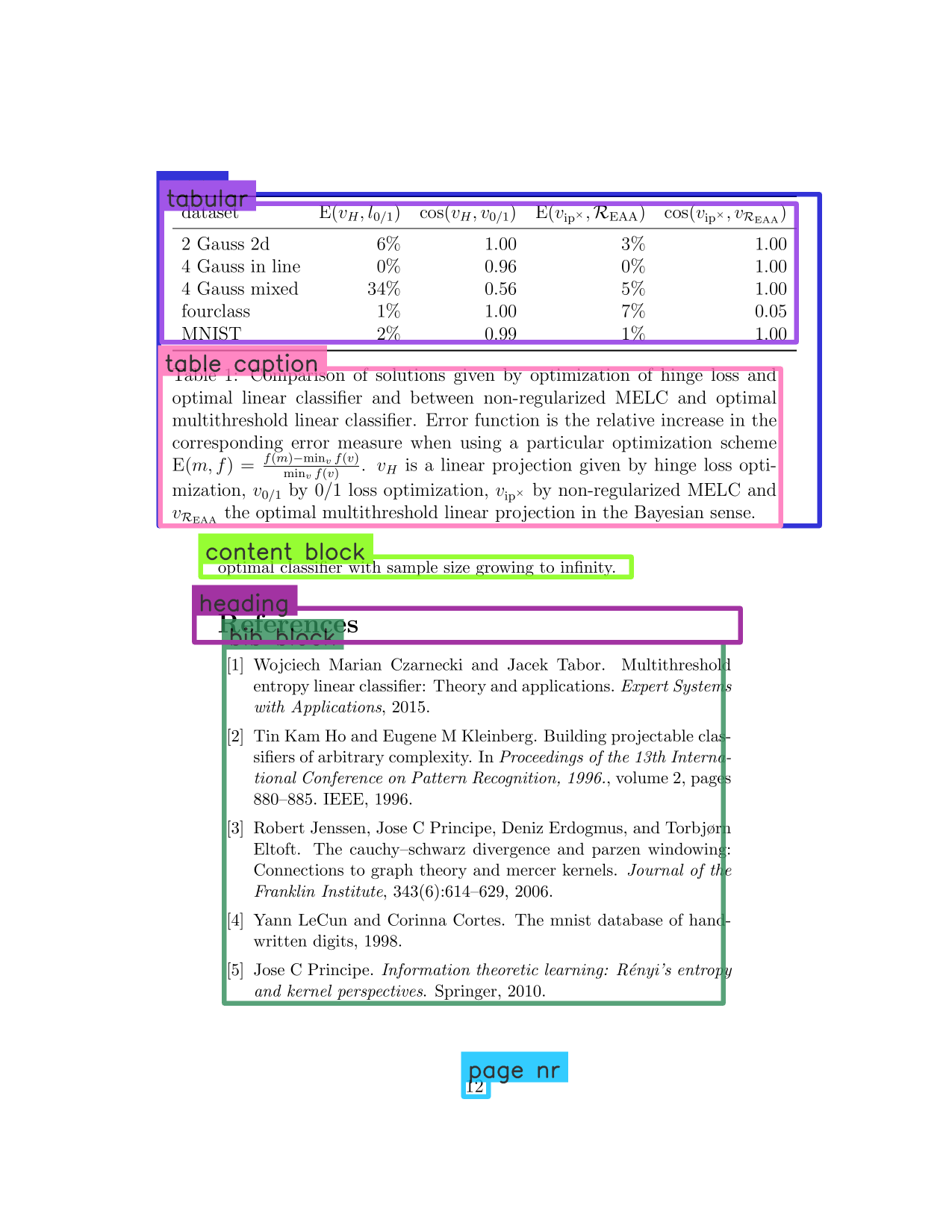}
\end{subfigure}
\begin{subfigure}[b]{0.3\linewidth}
    \centering
    \includegraphics[width=\linewidth]{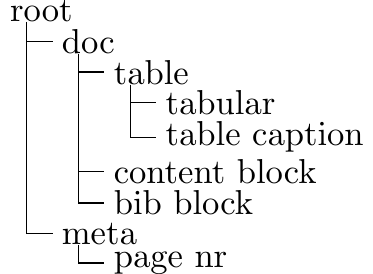}
    \caption{Ground truth}
\end{subfigure}
\begin{subfigure}[b]{0.3\linewidth}
    \centering
    \includegraphics[width=\linewidth]{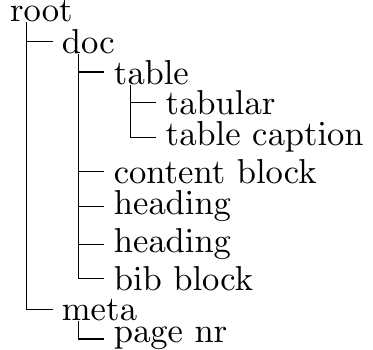}
    \caption{Raw pred.}
\end{subfigure}
\begin{subfigure}[b]{0.3\linewidth}
    \centering
    \includegraphics[width=\linewidth]{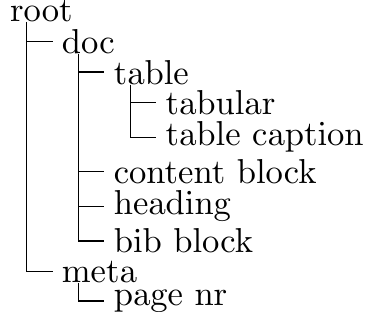}
    \caption{Predictions}
\end{subfigure}
    \caption{Raw predictions and structure-based refinement in \sys/.}
    \label{fig:structure_refinement2}
\end{figure}

We illustrate the effects of our structure-based refinement in Figure \ref{fig:structure_refinement1} and Figure \ref{fig:structure_refinement2}. We observe that bounding boxes of parent \entities/ from the raw predictions are refined such that they fully enclose all of their classified child \entities/. We particularly achieve improvement of the resulting predicted structure. For instance, for multi-figures, our refinement encloses figure graphics into individual figure structures to match the defined document grammar (see Figure \ref{fig:structure_refinement1}). Figure \ref{fig:structure_refinement2} shows how two nested \entities/ of the \textsc{heading} category are merged into a single \entity/ during refinement.

\begin{figure}[t!]
\centering
\begin{subfigure}[b]{0.32\linewidth}
    \centering
    \adjincludegraphics[width=\linewidth,trim={.0\width} {.05\height} {0.0\width} {.05\height}, clip]{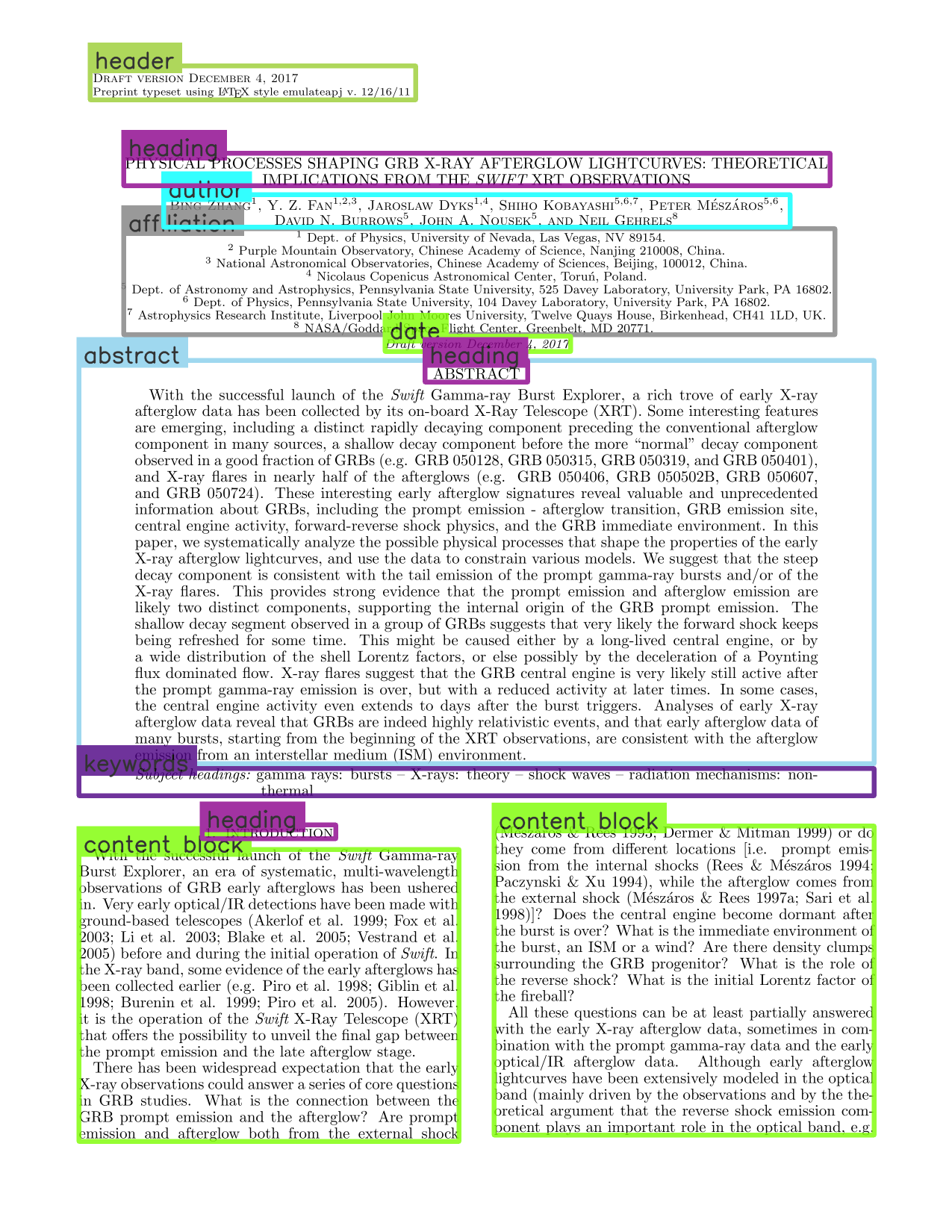}
\end{subfigure}
\begin{subfigure}[b]{0.32\linewidth}
    \centering
    \adjincludegraphics[width=\linewidth,trim={.0\width} {.05\height} {0.0\width} {.05\height},clip]{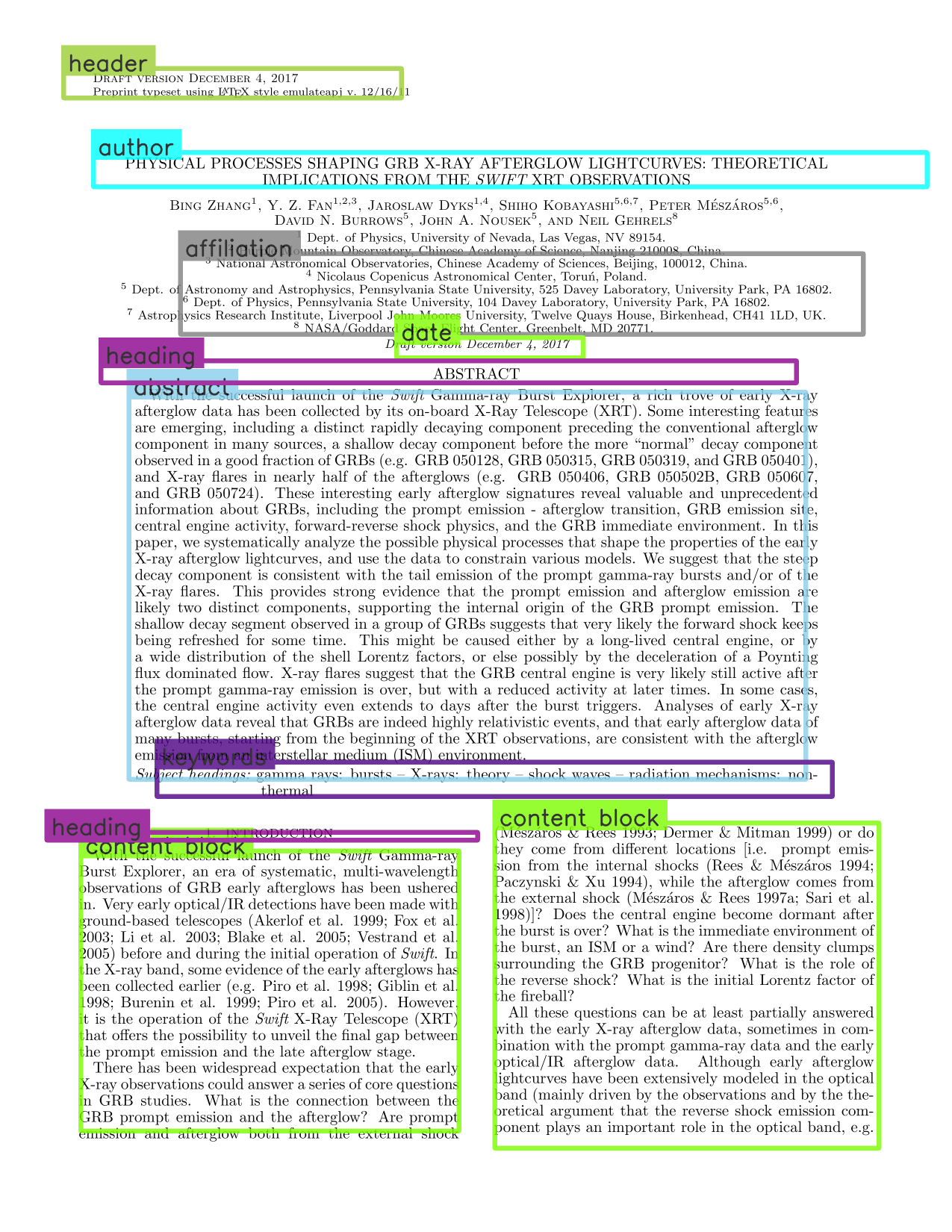}
\end{subfigure}
\begin{subfigure}[b]{0.32\linewidth}
    \centering
    \includegraphics[width=0.9\linewidth]{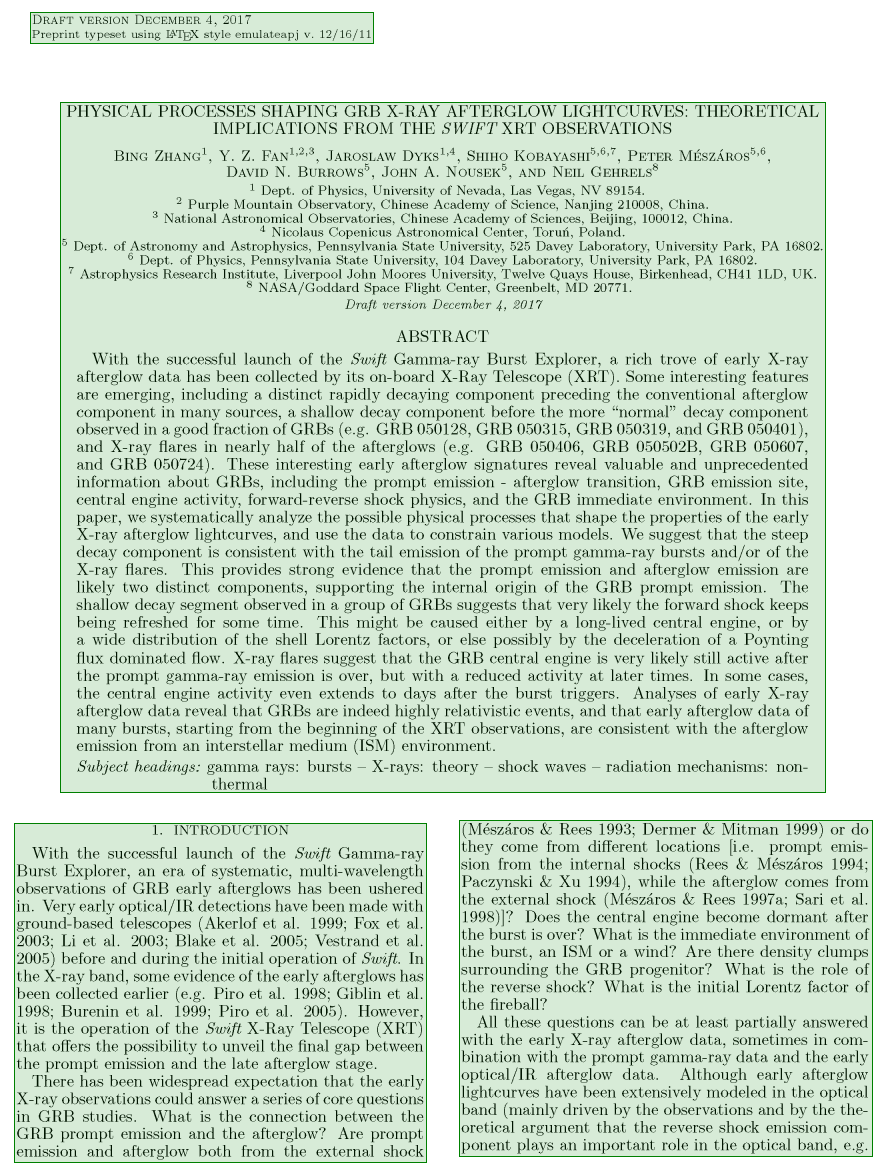}
\end{subfigure}\\
\begin{subfigure}[b]{0.3\linewidth}
    \centering
    \includegraphics[width=\linewidth]{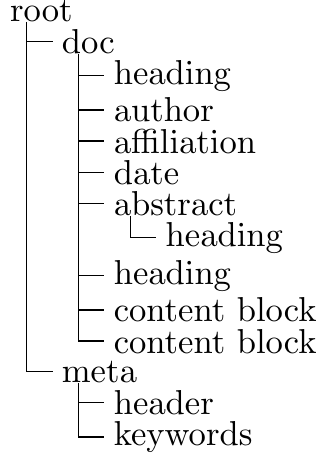}
    \caption{Ground truth}
\end{subfigure}
\begin{subfigure}[b]{0.3\linewidth}
    \centering
    \includegraphics[width=\linewidth]{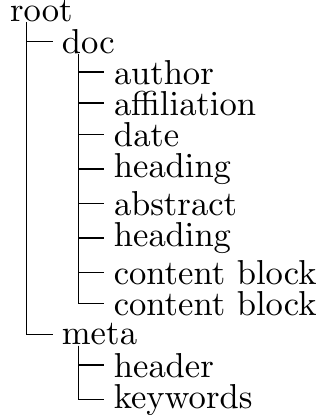}
    \caption{Predictions}
\end{subfigure}
\begin{subfigure}[b]{0.3\linewidth}
    \centering
    \includegraphics[width=\linewidth]{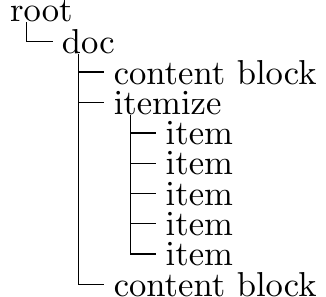}
    \caption{SOTA OCR}
\end{subfigure}
    \caption{Output with low F1 score (0.267), compared to state-of-the-art OCR software. Green regions in the OCR page recognition correspond to ``text areas''. Using the OCR tool, we convert the page to HTML and use our tree-graph to represent the resulting structure. The affiliation section is converted into a list by the OCR software during conversion to HTML.}
    \label{fig:comparison_ocr1}
\end{figure}

\begin{figure}[t!]
\centering
\begin{subfigure}[b]{0.32\linewidth}
    \centering
    \adjincludegraphics[width=\linewidth,trim={0.0\width} {.05\height} {0.0\width} {.05\height}, clip]{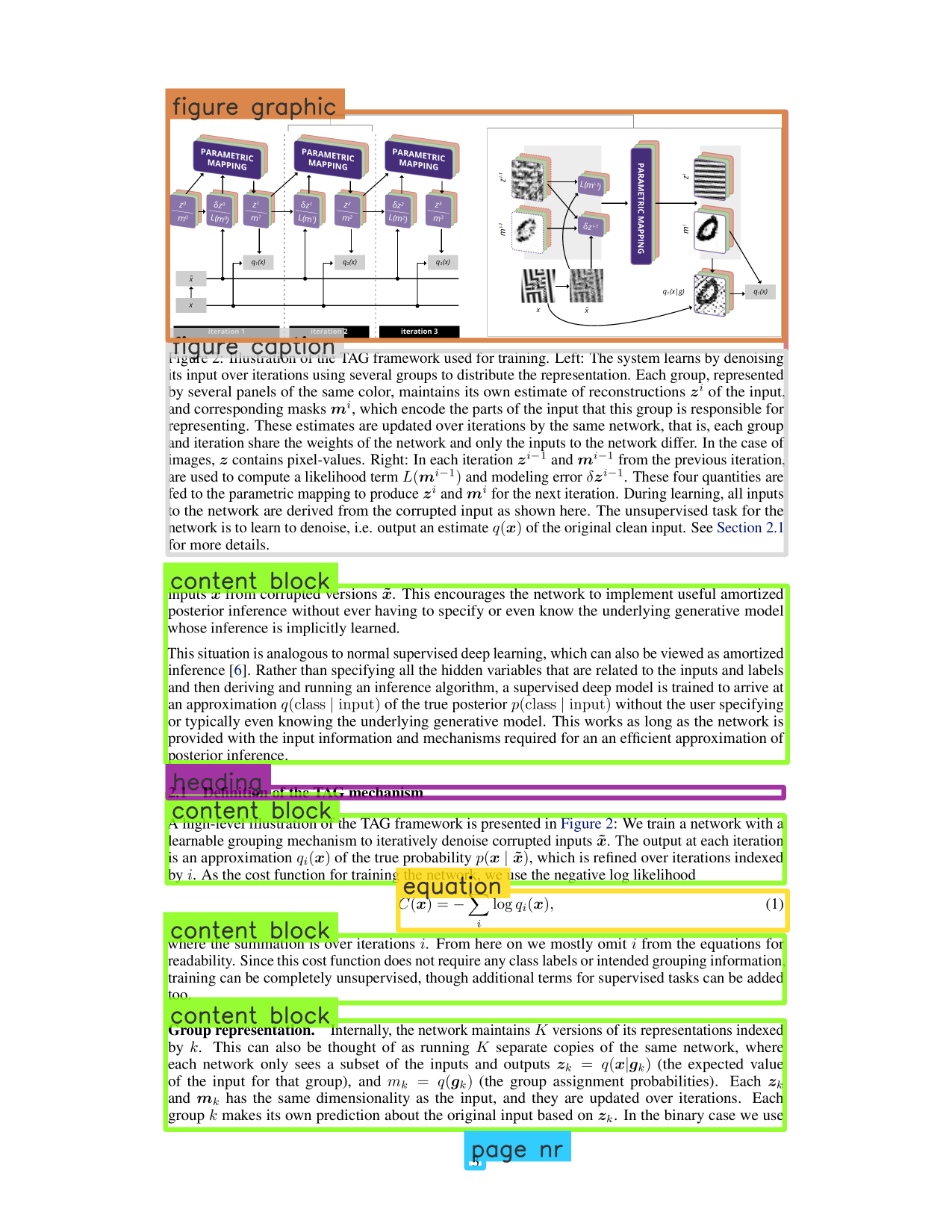}
\end{subfigure}
\begin{subfigure}[b]{0.32\linewidth}
    \centering
    \adjincludegraphics[width=\linewidth,trim={.0\width} {.05\height} {0.0\width} {.05\height},clip]{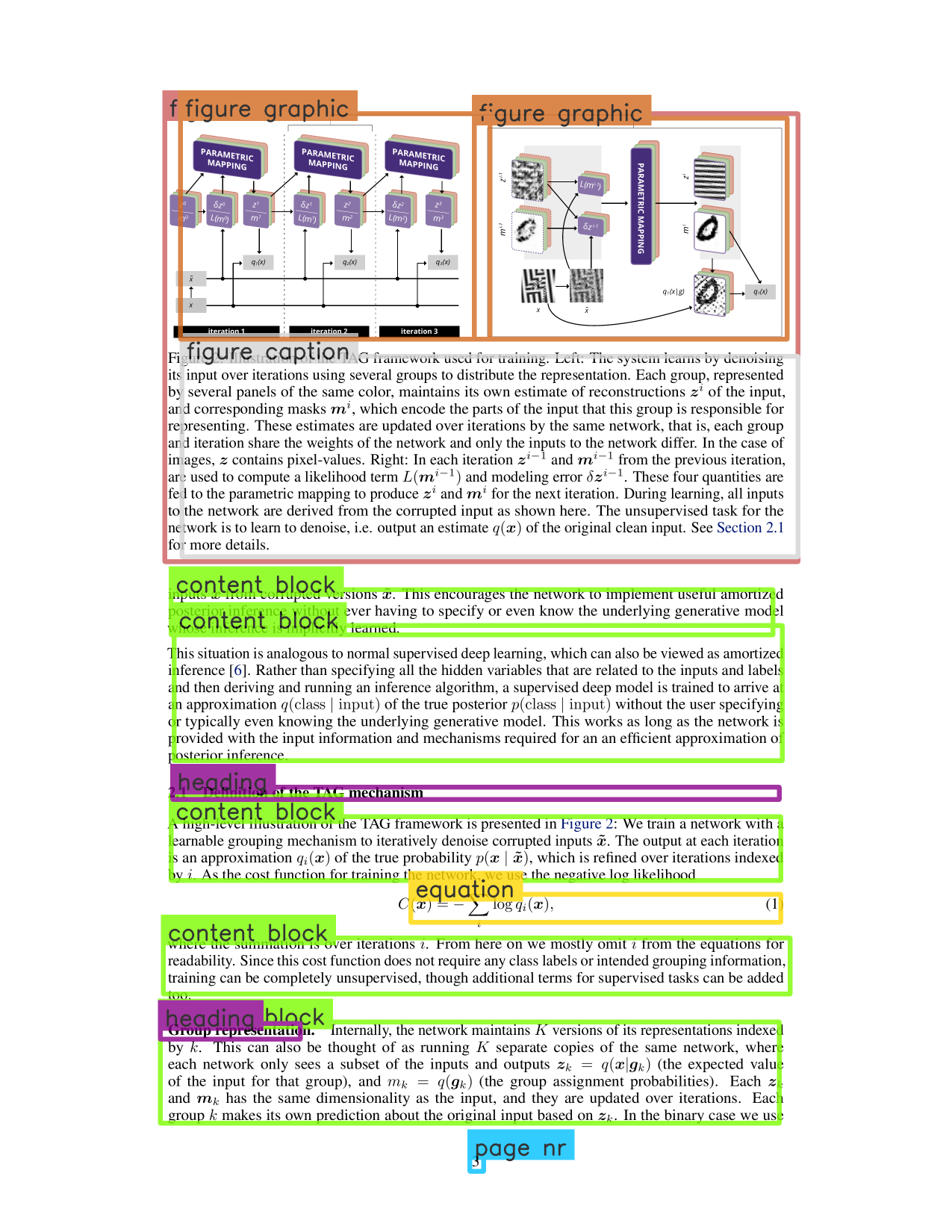}
\end{subfigure}
\begin{subfigure}[b]{0.32\linewidth}
    \centering
    \includegraphics[width=0.67\linewidth]{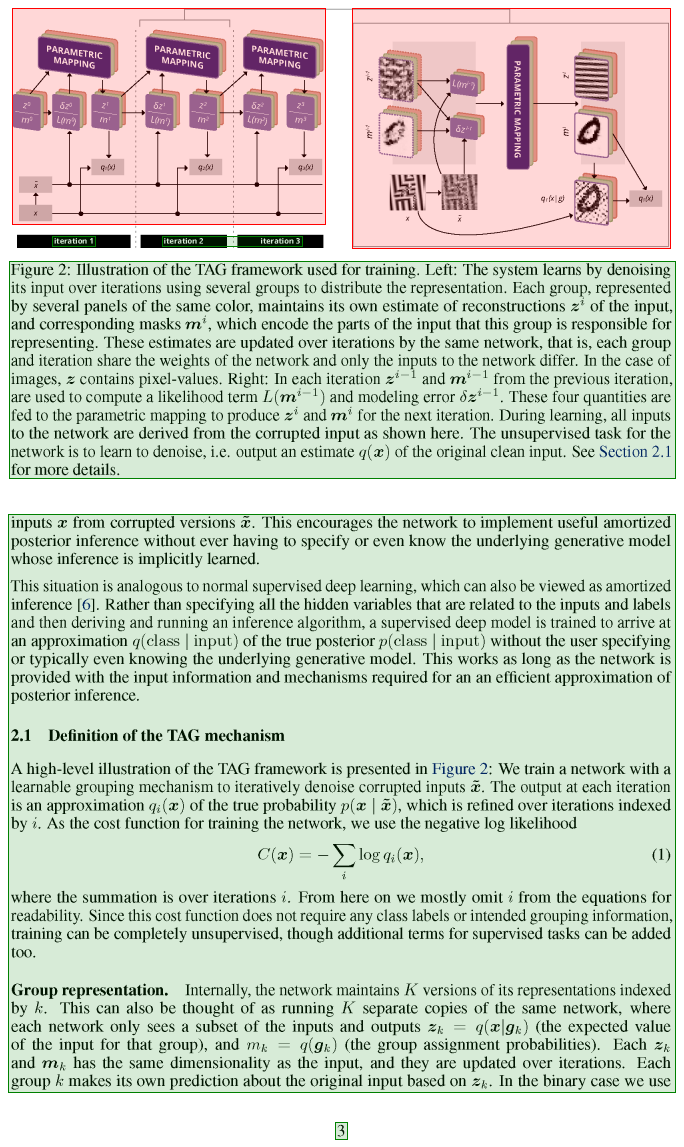}
\end{subfigure}\\
\begin{subfigure}[b]{0.3\linewidth}
    \centering
    \includegraphics[width=\linewidth]{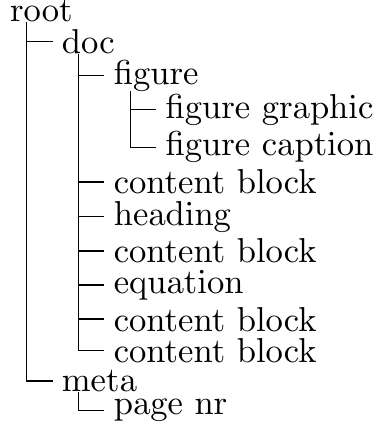}
    \caption{Ground truth}
\end{subfigure}
\begin{subfigure}[b]{0.3\linewidth}
    \centering
    \includegraphics[width=\linewidth]{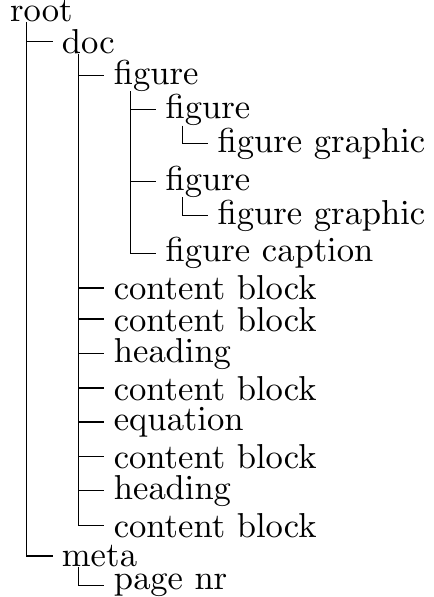}
    \caption{Predictions}
\end{subfigure}
\begin{subfigure}[b]{0.3\linewidth}
    \centering
    \includegraphics[width=\linewidth]{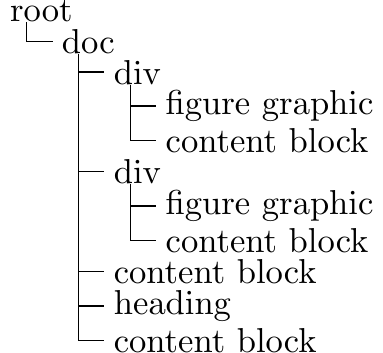}
    \caption{SOTA OCR}
\end{subfigure}
    \caption{Output with low F1 score (0.417), compared to state-of-the-art OCR software. Green and red regions in the OCR page recognition correspond to ``text areas'' and ``picture areas'', respectively. Using the OCR tool, we convert the page to HTML and use our tree-graph to represent the resulting structure. The content block in the second ``div'' section corresponds to the full figure caption text.}
    \label{fig:comparison_ocr2}
\end{figure}

We furthermore investigate the how the F1 measure for relation classification relates to overall parsing quality. Figure \ref{fig:comparison_ocr1} depicts the detected \entities/ and \relations/ for a document with an F1 score of 0.267. We note that the overall quality of the parsed page is still high. Our relation classification requires \entities/ in the page graph to be exactly matched with the corresponding \entities/ in the ground truth by surpassing the IoU threshold. For instance, the detected \textsc{header} \entities/ are not matched with the ground truth, due to the shape mismatch. This causes a penalty to the F1 score, as several relation triples in the prediction that involve the headings are considered mismatches. Figure \ref{fig:comparison_ocr2} shows another prediction with a low F1 score of 0.417. Here, mismatches can be accounted to the interpretation of entities that could be considered ambiguous. For instance, \sys/ detects an inline heading in the last content block, while this text segment is interpreted as standard text in the ground truth. We additionally compare our results qualitatively to a state-of-the-art OCR software.\footnote{We compare to outputs of ABBYY Finereader 15.} We observe that the page region detection fails to differentiate between many of the considered semantic categories, \eg/ \textsc{heading} \textsc{header} and \textsc{keywords} in Figure \ref{fig:comparison_ocr1} or \textsc{equation} in Figure \ref{fig:comparison_ocr2}. We note that the OCR software has access to the original PDF files of full resolution and all meta information, while \sys/ only operates on document renderings.

\subsection{Reproducibility}
For reproducibility purposes, we report results of \sys/ on the validation set. Table \ref{tab:val_entity_results} and Table \ref{tab:val_parsing_results} show the performance of the variants of \sys/ for \entity/ detection and prediction of hierarchical \relations/, respectively. Additionally, we include the complete results (including for IoU=0.8) on the test set in Table \ref{tab:test_entity_results} and Table \ref{tab:test_parsing_results}.

We report average scores over three runs with differing random seeds for all fine-tuning experiments on \datasetmanual/ to account for the small number of training samples.

\begin{table*}[t]
\small
    \centering
\sisetup
  { 
    round-precision    = 1           ,
    round-integer-to-decimal,
    round-mode = places,
    table-align-text-post=false,
    table-number-alignment=center,
    table-format=2.1,
  }
\renewrobustcmd{\bfseries}{\fontseries{b}\selectfont}
\begingroup
\setlength{\tabcolsep}{2.25pt} 
    \begin{tabular}{@{}l  SSS  SSS  SSS@{}}
    \toprule
\multicolumn{1}{c}{} &    \multicolumn{3}{c}{IoU=0.5} & \multicolumn{3}{c}{IoU=0.65} & \multicolumn{3}{c}{IoU=0.8} \\
\cmidrule(lr){2-4} \cmidrule(lr){5-7} \cmidrule(lr){8-10}
\multicolumn{1}{@{}l}{AP}   & \multicolumn{1}{c}{Baseline} & \multicolumn{1}{c}{WS} & \multicolumn{1}{c}{WS+FT}& \multicolumn{1}{c}{Baseline} & \multicolumn{1}{c}{WS} & \multicolumn{1}{c}{WS+FT}& \multicolumn{1}{c}{Baseline} & \multicolumn{1}{c}{WS} & \multicolumn{1}{c@{}}{WS+FT}\\
\midrule
          mAP &           50.05 &           41.17 &  \bfseries 70.95 &           37.17 &           37.89 &  \bfseries 59.05 &         15.58 &           24.86 &  \bfseries 37.67 \\
\midrule
     abstract &           78.13 &           99.63 &  \bfseries 100.0 &           78.13 &           81.93 &  \bfseries 100.0 &         50.98 &           52.08 &  \bfseries 66.27 \\
  affiliation &           50.73 &               0 &  \bfseries 57.64 &           39.06 &               0 &  \bfseries 39.24 &          1.54 &               0 &   \bfseries 1.94 \\
       author &           26.23 &               0 &  \bfseries 31.08 &  \bfseries 23.96 &               0 &           18.82 &             0 &               0 &   \bfseries 5.54 \\
   bib. block &           27.62 &            75.1 &  \bfseries 78.84 &           14.29 &            75.1 &  \bfseries 81.75 &         14.29 &           57.14 &  \bfseries 60.32 \\
  cont. block &           82.94 &           69.86 &  \bfseries 90.66 &           76.05 &           65.17 &  \bfseries 86.85 &         57.81 &           53.98 &  \bfseries 73.59 \\
         date &               0 &               0 &  \bfseries 34.26 &    \bfseries 0.0 &               0 &               0 &  \bfseries 0.0 &               0 &               0 \\
     equation &           59.77 &           54.23 &  \bfseries 83.43 &           35.87 &           46.67 &  \bfseries 73.86 &         11.64 &            30.8 &  \bfseries 39.81 \\
 fig. caption &           50.69 &           39.76 &  \bfseries 69.95 &           50.55 &           14.46 &   \bfseries 62.3 &         22.36 &           18.07 &   \bfseries 44.6 \\
 fig. graphic &           40.32 &            5.42 &   \bfseries 77.0 &           31.73 &             7.5 &   \bfseries 74.5 &         11.99 &             4.5 &   \bfseries 52.8 \\
       figure &  \bfseries 71.05 &           39.26 &           53.94 &  \bfseries 60.04 &           40.82 &           51.56 &         18.45 &           17.13 &  \bfseries 41.69 \\
       footer &           62.12 &               0 &  \bfseries 74.48 &           35.24 &               0 &  \bfseries 56.64 &         26.22 &               0 &  \bfseries 28.81 \\
       header &           64.98 &               0 &  \bfseries 68.12 &  \bfseries 45.92 &               0 &           44.45 &          9.42 &               0 &   \bfseries 12.2 \\
      heading &           56.77 &           58.89 &  \bfseries 72.62 &           38.33 &            54.8 &  \bfseries 65.78 &          7.83 &           21.06 &  \bfseries 21.39 \\
         item &               0 &           68.64 &  \bfseries 69.55 &               0 &  \bfseries 68.72 &           63.73 &             0 &  \bfseries 36.37 &           33.93 \\
      itemize &               0 &  \bfseries 69.14 &           63.49 &               0 &  \bfseries 63.49 &           53.18 &             0 &           33.33 &  \bfseries 44.71 \\
     keywords &           40.74 &               0 &  \bfseries 48.47 &  \bfseries 18.52 &               0 &           15.11 &             0 &               0 &  \bfseries 12.35 \\
     page nr. &           57.53 &               0 &  \bfseries 60.29 &           18.66 &               0 &  \bfseries 23.19 &          0.13 &               0 &   \bfseries 2.44 \\
 tab. caption &            56.4 &           81.17 &  \bfseries 91.47 &           38.08 &           88.65 &  \bfseries 89.13 &          9.55 &   \bfseries 50.8 &           44.17 \\
        table &           92.03 &           94.59 &  \bfseries 97.37 &           62.85 &  \bfseries 88.26 &           87.61 &         18.69 &           69.28 &  \bfseries 77.52 \\
      tabular &           83.03 &           67.67 &  \bfseries 96.25 &           76.19 &           62.13 &   \bfseries 93.2 &         50.76 &           52.61 &  \bfseries 89.29 \\
\bottomrule
    \end{tabular}
    \endgroup
\caption{Validation set: Comparison of \entity/ detection (average precision) without structure-based refinement.}
    \label{tab:val_entity_results}
\end{table*}

\begin{table*}[t]
\small %
    \centering
\sisetup
  { 
    round-precision    = 1           ,
    round-integer-to-decimal,
    round-mode = places,
    table-align-text-post=false,
    table-number-alignment=center,
    table-format=2.1,
  }
\renewrobustcmd{\bfseries}{\fontseries{b}\selectfont}
\begingroup
\setlength{\tabcolsep}{2.25pt} 
    \begin{tabular}{@{}l  SSS  SSS SSS@{}}
    \toprule
\multicolumn{1}{c}{} &    \multicolumn{3}{c}{IoU=0.5} & \multicolumn{3}{c}{IoU=0.65} & \multicolumn{3}{c}{IoU=0.8} \\
\cmidrule(lr){2-4} \cmidrule(lr){5-7} \cmidrule(lr){8-10}
\multicolumn{1}{@{}l}{AP}   & \multicolumn{1}{c}{Baseline} & \multicolumn{1}{c}{WS} & \multicolumn{1}{c}{WS+FT}& \multicolumn{1}{c}{Baseline} & \multicolumn{1}{c}{WS} & \multicolumn{1}{c}{WS+FT}& \multicolumn{1}{c}{Baseline} & \multicolumn{1}{c}{WS} & \multicolumn{1}{c@{}}{WS+FT}\\
\midrule
          mean AP &           49.93 &           34.59 &  \bfseries 69.35 &           38.53 &           32.41 &  \bfseries 56.54 &           14.73 &            23.5 &  \bfseries 35.59 \\
\midrule
     abstract &  \bfseries 95.24 &           90.48 &           95.16 &           90.48 &           81.01 &  \bfseries 95.24 &           48.86 &           28.34 &   \bfseries 75.1 \\
  affiliation &  \bfseries 51.62 &               0 &           46.02 &            5.92 &               0 &   \bfseries 16.2 &   \bfseries 0.96 &               0 &            0.81 \\
       author &           17.95 &               0 &  \bfseries 23.61 &  \bfseries 20.37 &               0 &            16.7 &            4.94 &               0 &   \bfseries 8.01 \\
   bib. block &           42.42 &           79.09 &   \bfseries 94.7 &           43.18 &  \bfseries 93.94 &           80.26 &           12.73 &  \bfseries 96.21 &           65.62 \\
  cont. block &  \bfseries 89.31 &           69.75 &           88.41 &           83.17 &           67.03 &  \bfseries 84.38 &           64.44 &           55.58 &  \bfseries 74.22 \\
         date &               0 &               0 &  \bfseries 24.07 &               0 &               0 &   \bfseries 9.26 &    \bfseries 0.0 &               0 &               0 \\
     equation &           65.84 &           54.53 &  \bfseries 82.05 &            40.6 &           52.14 &   \bfseries 72.8 &            8.89 &           36.43 &  \bfseries 38.37 \\
 fig. caption &           47.77 &            30.5 &  \bfseries 69.23 &           43.95 &           17.73 &  \bfseries 59.54 &            16.7 &           19.55 &  \bfseries 39.84 \\
 fig. graphic &           22.28 &            5.19 &  \bfseries 60.21 &           15.93 &            4.36 &   \bfseries 54.5 &            6.15 &            1.62 &  \bfseries 36.61 \\
       figure &           47.82 &           35.28 &  \bfseries 63.52 &           43.96 &           33.94 &  \bfseries 59.39 &            22.5 &           20.99 &  \bfseries 51.31 \\
       footer &            55.7 &               0 &  \bfseries 69.26 &           48.86 &               0 &  \bfseries 59.68 &            5.02 &               0 &    \bfseries 7.9 \\
       header &           79.69 &               0 &  \bfseries 88.28 &  \bfseries 64.84 &               0 &           56.56 &  \bfseries 12.08 &               0 &            6.52 \\
      heading &           53.74 &            52.1 &  \bfseries 66.35 &           33.11 &  \bfseries 46.04 &           45.39 &            6.66 &   \bfseries 26.3 &           16.67 \\
         item &               0 &           33.57 &  \bfseries 50.49 &               0 &  \bfseries 35.26 &           33.47 &               0 &   \bfseries 53.0 &           10.43 \\
      itemize &               0 &              25 &  \bfseries 58.33 &               0 &              25 &   \bfseries 50.0 &               0 &               0 &  \bfseries 58.33 \\
     keywords &           36.36 &               0 &  \bfseries 58.98 &           36.36 &               0 &  \bfseries 42.95 &           20.45 &               0 &  \bfseries 22.29 \\
     page nr. &           74.72 &               0 &  \bfseries 77.31 &           28.53 &               0 &  \bfseries 42.04 &            0.75 &               0 &   \bfseries 1.99 \\
 tab. caption &           55.18 &           69.11 &  \bfseries 76.64 &           40.16 &           61.56 &  \bfseries 63.42 &           16.46 &  \bfseries 41.48 &            28.6 \\
        table &           84.47 &  \bfseries 96.33 &           94.31 &           62.67 &           87.85 &  \bfseries 89.62 &           14.64 &            68.2 &  \bfseries 80.19 \\
      tabular &           78.41 &           50.82 &  \bfseries 99.98 &           68.44 &            42.4 &  \bfseries 99.45 &           32.41 &           22.26 &   \bfseries 89.0 \\
\bottomrule
    \end{tabular}
    \endgroup
\caption{Test set: Average precision (AP) of \entity/ detection.}
    \label{tab:test_entity_results}
\end{table*}

\begin{table*}[t]
\small
\sisetup
  { 
    round-precision    = 3           ,
    round-integer-to-decimal,
    round-mode = places,
    table-number-alignment=center,
    table-format=1.3,
  }
\renewrobustcmd{\bfseries}{\fontseries{b}\selectfont}
    \centering
\begingroup
\setlength{\tabcolsep}{1.25pt} 
\begin{tabular}{@{}l  SSS  SSS  SSS@{}}
    \toprule
 &    \multicolumn{3}{c}{IoU=0.5} & \multicolumn{3}{c}{IoU=0.65} & \multicolumn{3}{c}{IoU=0.8} \\
\cmidrule(lr){2-4} \cmidrule(lr){5-7} \cmidrule(lr){8-10}
  &{Baseline}&{WS} & {WS+FT}&{Baseline}&{WS}&{WS+FT}&{Baseline}&{WS}&{WS+FT}\\
\midrule
All                     &  0.4064 &  0.3693 &  \bfseries 0.5499 &    0.3370 &  0.3498 &   \bfseries 0.478 &    0.1309 &  0.2206 &    \bfseries 0.32 \\ 
\textit{followed\_by}   & 0.3599 &  0.3683 &  \bfseries 0.5313 &    0.3029 &  0.3561 &   \bfseries 0.454 &    0.1186 &  0.2156 &  \bfseries 0.2937 \\ 
\textit{parent\_of}     & 0.5192 &  0.3721 &  \bfseries 0.5918 &    0.4176 &  0.3326 &  \bfseries 0.5321 &    0.1604 &  0.2345 &  \bfseries 0.3792 \\ 
\midrule
\textbf{Refined}: & & & & & & & & & \\
All &   0.4470 &  0.4012 &  \bfseries 0.6578 &    0.3586 &  0.3848 &  \bfseries 0.5761 &    0.1636 &  0.2660 &  \bfseries 0.4095 \\ 
   \textit{followed\_by}  &     0.4019 &  0.3903 &  \bfseries 0.6019 &    0.3096 &  0.3839 &  \bfseries 0.5128 &    0.1402 &  0.2538 &  \bfseries 0.3375 \\ 
     \textit{parent\_of} &     0.5510 &  0.4277 &  \bfseries 0.7706 &    0.4695 &  0.3871 &   \bfseries 0.704 &    0.2174 &  0.2976 &  \bfseries 0.5548 \\ 

\bottomrule
    \end{tabular}
    \endgroup
\caption{Validation set: Performance in predicting hierarchical \relations/ (as measured by F1).}
    \label{tab:val_parsing_results}
\end{table*}

\begin{table*}[t]
\small
\sisetup
  { 
    round-precision    = 3           ,
    round-integer-to-decimal,
    round-mode = places,
    table-number-alignment=center,
    table-format=1.3,
  }
\renewrobustcmd{\bfseries}{\fontseries{b}\selectfont}
    \centering
\begingroup
\setlength{\tabcolsep}{1.25pt} 
\begin{tabular}{@{}l  SSS  SSS SSS@{}}
    \toprule
 &    \multicolumn{3}{c}{IoU=0.5} & \multicolumn{3}{c}{IoU=0.65} & \multicolumn{3}{c}{IoU=0.8} \\
\cmidrule(lr){2-4} \cmidrule(lr){5-7} \cmidrule(lr){8-10}
  &{Baseline}&{WS} & {WS+FT}&{Baseline}&{WS}&{WS+FT}&{Baseline}&{WS}&{WS+FT}\\
\midrule
All &    0.4155 &  0.3425 &  \bfseries 0.5037 &    0.3215 &  0.3181 &  \bfseries 0.4454 &    0.1136 &  0.2136 &  \bfseries 0.3134 \\
\textit{followed\_by} &  0.4130 &  0.3865 &  \bfseries 0.5056 &    0.3135 &  0.3659 &  \bfseries 0.4466 &    0.1184 &  0.2438 &  \bfseries 0.3078 \\
\textit{parent\_of} &   0.4208 &  0.2348 &  \bfseries 0.4999 &    0.3388 &  0.1981 &  \bfseries 0.4428 &    0.1033 &  0.1384 &  \bfseries 0.3246 \\ 
\midrule
\textbf{Refined}: & & & & & & & & & \\
All  &    0.4533 &  0.3815 &  \bfseries 0.6153 &    0.3633 &  0.3544 &  \bfseries 0.5577 &    0.1574 &  0.2542 &  \bfseries 0.3954 \\ 
   \textit{followed\_by}  &   0.4545 &  0.4101 &  \bfseries 0.5807 &    0.3510 &  0.3936 &  \bfseries 0.5235  &  0.1500 &  0.2769 &   \bfseries 0.352 \\ 
\textit{parent\_of}  &  0.4508 &  0.3169 &  \bfseries 0.6785 &    0.3887 &  0.2626 &  \bfseries 0.6203  &    0.1724 &  0.2016 &  \bfseries 0.4744 \\
\bottomrule
    \end{tabular}
    \endgroup
\caption{Test set: Performance in predicting hierarchical \relations/ (as measured by F1).}
    \label{tab:test_parsing_results}
\end{table*}

\section{Document Grammar}

Hierarchical \relations/ between \entity/ pairs follow a predefined grammar (see Table~\ref{tab:document_grammar}). All \entities/ with meta-information have no ordering, \ie/, their \relation/ type is $\Psi = \text{null}$. Some \entities/ (such as, \eg/, figures) have only a certain set of allowed child \entities/. For instance, a figure can contain a figure caption, a graphic, or a subfigure (\ie/, another nested figure), but not other \entities/ such as a table or an abstract. Finally, the hierarchical structures $T_i$ must form a tree. That is, an \entity/ is allowed to have multiple ordered siblings (\ie/, multiple \entities/ with the same nesting level). However, each \entity/ must only have one parent, \ie/, for an \entity/ $E$ there is exactly one \relation/ $(E', E, \mathit{parent\_of})$ with an \entity/ $E \neq E$.

\begin{table}[t]
    \centering
\small
\begin{tabular}{l l l l}
\toprule
\Entity/ ($\mathcal{C}$) & \Relation/ types $\Psi$ & Valid \entities/ & Notes\\
\midrule
Abstract & $\mathit{parent\_of}$ & Heading & \\
Figure & $\mathit{parent\_of}$ & Figure & Float\\
 & $\mathit{parent\_of}$ & Fig. graphic & \\
 & $\mathit{parent\_of}$ & Fig. caption & \\
Fig. graphic&  $\mathit{followed\_by}$ & Fig. caption & if nested\\
Item & $\mathit{parent\_of}$  & Equation &\\ 
Itemize & $\mathit{parent\_of}$ & Item &\\
Table & $\mathit{parent\_of}$ & Tabular & Float \\
 & $\mathit{parent\_of}$ & Tab. caption& \\
Tabular & $\mathit{parent\_of}$ & Tab. cell&\\
 & $\mathit{parent\_of}$ & Tab. row&\\
 & $\mathit{parent\_of}$ & Tab. col.&\\
Date&   {null} & ---  & Meta \\
Footer& {null} & ---  & Meta \\
Header & {null} & ---  & Meta \\
Keywords&{null} & ---  & Meta \\
PageNr &{null} & ---  & Meta \\
\midrule
All others & $\mathit{parent\_of}$, & --- \\ 
Any entity & $\mathit{followed\_by}$ & \emph{any sibling} & \\
\bottomrule
\end{tabular}
\caption{Document grammar for different \entity/ categories that is utilized in our heuristics. Every category can by default exist on the highest hierarchical level, \ie/, without being nested. Hierarchical nesting for the child \entities/ of floats, \eg/ captions, is, however, encouraged in the automatic refinement process. Further details are included in the supplements.}
\label{tab:document_grammar}
\end{table}

\section{Datasets with Document Structure: \datasetmanual/}
\begin{figure}[t!]
    \centering
\begin{adjustbox}{width=\linewidth}
\begin{tikzpicture}
    \begin{axis}[
        width=\textwidth,
        height=.35\textwidth,
        axis on top,
        /tikz/ybar interval,
        tick align=outside,
        ymin=0,
        axis y line*=left,
        axis x line*=bottom,
        x tick label style={rotate=90,anchor=east, font=\small},
        ymajorgrids = true,
       enlarge x limits=0.05,
        xlabel={Leaf nodes in document},
        ylabel={Frequency},
        width=\textwidth,
        xtick=data,
        xticklabels={0,,,,,50,,,,,100,,,,,150,,,,,200,,,,,250,,,,,300},
        label style={font=\large},
        y tick label style={
            /pgf/number format/.cd,
            fixed,
            fixed zerofill,
            precision=1,
            /tikz/.cd
        },
            ]
    \addplot [fill=ETH3] coordinates{(0, 22) (10, 40) (20, 41) (30, 30) (40, 25) (50, 22) (60, 28) (70, 31) (80, 20) (90, 26) (100, 14) (110, 11) (120, 10) (130, 4) (140, 4) (150, 4) (160, 2) (170, 2) (180, 0) (190, 0) (200, 5) (210    , 2) (220, 4) (230, 0) (240, 1) (250, 3) (260, 1) (270, 0) (280, 0) (290, 0) (300, 3) (310, 0)}  ;;
        \end{axis}
    \clipright
\end{tikzpicture}
\end{adjustbox}
\caption{Number of leaf nodes in documents.}
\label{fig:leaf_histogram_manual}
\end{figure}
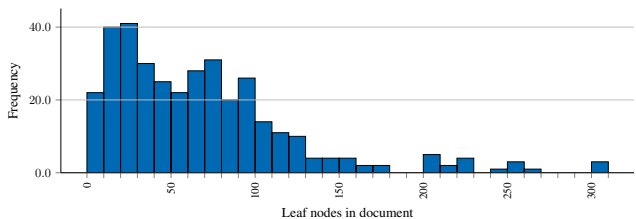

Figure \ref{fig:leaf_histogram_manual} shows the number of leaf nodes in the document graph. Furthermore, Table \ref{tab:dataset_cat_statistics_manual} reports the frequency and average depth of the different \entities/ in the dataset.

\begin{table}[t]
\small
\sisetup
  { 
    round-mode         = places      ,
    round-precision    = 2           ,
    round-integer-to-decimal,
    table-align-text-post   = false,
  }
\renewrobustcmd{\bfseries}{\fontseries{b}\selectfont}
    \centering
    \begin{tabular}{l S[zero-decimal-to-integer = true, table-format=5.0] S[table-format=2.2, table-omit-exponent,  fixed-exponent = -2] S[table-format=1.2]}
    \toprule
    Category& {Frequency} & {\%} & {Avg. depth } \\
    \midrule
 abstract& 63.0& 0.00201606451406445000&  1.0 \\
affiliation& 82.0& 0.00262408397068706212&  1.0 \\
author& 89.0& 0.00284809113891644526&  1.0 \\
bibliogr. block& 32.0& 0.00102403276904860954&  2.0 \\
bibliography& 24.0& 0.00076802457678645721&  1.0833333333333333 \\
code& 3.0& 0.00009600307209830715&  2.0 \\
content block& 1009.0& 0.03228903324906397254&  2.045589692765114 \\
content line& 10729.0& 0.34333898684757913067&  3.0579737160965608 \\
content lines& 627.0& 0.02006464206854619362&  3.041467304625199 \\
date& 16.0& 0.00051201638452430477&  1.0 \\
equation& 353.0& 0.01129636148356747381&  1.971671388101983 \\
equation formula& 364.0& 0.01164837274792793372&  3.008241758241758 \\
equation label& 275.0& 0.00880028160901148888&  2.949090909090909 \\
figure& 607.0& 0.01942462158789081197&  2.441515650741351 \\
figure caption& 404.0& 0.01292841370923869526&  3.1633663366336635 \\
figure graphic& 454.0& 0.01452846491087714763&  3.6255506607929515 \\
footer& 81.0& 0.00259208294665429304&  1.0 \\
header& 106.0& 0.00339210854747351923&  1.009433962264151 \\
heading& 398.0& 0.01273640756504208077&  2.0979899497487438 \\
item& 63.0& 0.00201606451406445000&  3.0793650793650795 \\
itemize& 24.0& 0.00076802457678645721&  1.9583333333333333 \\
page nr& 261.0& 0.00835226727255272174&  1.0 \\
section& 527.0& 0.01686453966526928888&  1.2903225806451613 \\
keywords& 36.0& 0.00115203686517968565&  1.0 \\
table& 185.0& 0.00592018944606227410&  1.8054054054054054 \\
table caption& 183.0& 0.00585618739799673594&  2.80327868852459 \\
table cell& 11146.0& 0.35668341386924379277&  3.5415395657635025 \\
table col& 1109.0& 0.03548913565234087381&  3.692515779981966 \\
table row& 1812.0& 0.05798585554737751419&  3.6788079470198674 \\
tabular& 187.0& 0.00598419149412781227&  2.802139037433155 \\
    \bottomrule
    \end{tabular}
    \caption{Statistics by \entity/ of \datasetmanual/.}
    \label{tab:dataset_cat_statistics_manual}
\end{table}

Annotators are given a set of instructions for annotating \entities/: All bounding boxes should fully enclose the contained contents and at most extend to full column width. Figure graphics or captions should always be enclosed by a \textsc{figure} \entity/. If a figure contains multiple subfigures, each subfigure should consist of an individual nested \textsc{figure} that contains a \textsc{figure graphic}. Furthermore \textsc{content block} or \textsc{bibliography block} entities are text or bibliography regions that should extend at most a single column/page and contains no other categories. 
To give annotators the freedom to handle the large variety of document appearances, we do not enforce a strict document grammar during manual annotation.

\section{Datasets with Document Structure: \datasetauto/}

 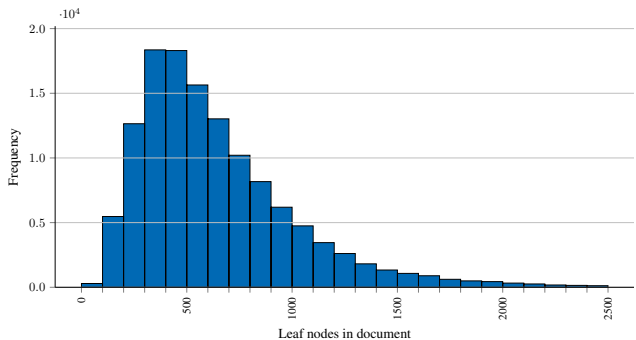
\begin{figure}[t!]
     \centering
 \begin{adjustbox}{width=\linewidth}
 \begin{tikzpicture}
     \begin{axis}[
         width=\textwidth,
         height=.5\textwidth,
         axis on top,
         /tikz/ybar interval,
         tick align=outside,
         ymin=0,
         axis y line*=left,
         axis x line*=bottom,
         x tick label style={rotate=90,anchor=east, font=\small},
         ymajorgrids = true,
       enlarge x limits=0.05,
         xlabel={Leaf nodes in document},
         ylabel={Frequency},
         width=\textwidth,
         xtick=data,
         xticklabels={0,,,,,500,,,,,1000,,,,,1500,,,,,2000,,,,,2500},
         label style={font=\large},
         y tick label style={
             /pgf/number format/.cd,
             fixed,
             fixed zerofill,
             precision=1,
             /tikz/.cd
         },
             ]
     \addplot [fill=ETH3] coordinates{(0, 293) (100, 5471) (200, 12645) (300, 18354) (400, 18303) (500, 15642) (600, 13022) (700, 10208) (800, 8170) (900, 6191) (1000, 4748) (1100, 3453) (1200, 2610) (1300, 1812) (1400, 1335) (1500, 1074) (1600, 891) (1700, 617) (1800, 494) (1900, 438) (2000, 324) (2100, 257) (2200, 177) (2300, 147) (2400, 121) (2500, 0)} ;;
         \end{axis}
     \clipright
 \end{tikzpicture}
 \end{adjustbox}
 \caption{Number of leaf nodes in documents for weak supervision.}
 \label{fig:leaf_histogram}
 \end{figure}

Figure~\ref{fig:leaf_histogram}
 and Table~\ref{tab:dataset_cat_statistics} show the 
 descriptive statistics of the dataset. Evidently, the most common category in the dataset is content line. Content lines typically represent leaf nodes in the graph and are children of larger \entities/, such as abstract, captions, or content blocks. 

 \begin{table}[t]
 \small
 \sisetup
   { 
     round-mode         = places      ,
     round-precision    = 2           ,
     round-integer-to-decimal,
     table-align-text-post   = false,
   }
 \renewrobustcmd{\bfseries}{\fontseries{b}\selectfont}
     \centering
     \begin{tabular}{l S[zero-decimal-to-integer = true, table-format=8.0] S[table-format=2.2, table-omit-exponent,  fixed-exponent = -2] S[table-format=1.2]}
     \toprule
     Category& {Frequency} & {\%} & {Avg. Depth } \\
     \midrule
 abstract&89291.0&0.00094397496778532528& 1.0019486846378693\\
 author&48.0&0.00000050745090158802& 3.0\\
 bibliogr. block&242412.0&0.00256275391574488191& 2.9434158415841583\\
 bibliography&80864.0&0.00085488561887527904& 1.9282967066942036\\
 caption&26.0&0.00000027486923836018& 4.384615384615385\\
 content block&5033714.0&0.05321588974242130732& 2.4060347886272444\\
 content line&63339623.0&0.66961976661656430476& 3.471712154577852\\
 date&5.0&0.00000005285946891542& 2.0\\
 equation&1489078.0&0.01574237445072668509& 2.440061947449436\\
 equation formula&1743705.0&0.01843426404903193705& 3.431982991589853\\
 equation label&1503778.0&0.01589778128933801507& 3.4415208975780263\\
 figure&478086.0&0.00505427441117934674& 2.50303100029285\\
 figure caption&263495.0&0.00278564115237363557& 3.483839257926658\\
 figure graphic&408088.0&0.00431426299015105503& 3.518840991685311\\
 heading&975414.0&0.01031197320253278947& 2.535006674089156\\
 item&436222.0&0.00461169264984433083& 3.5763225013302997\\
 itemize&140415.0&0.00148445246555169554& 2.6393204298623187\\
 meta&127477.0&0.00134767330378615874& 3.0\\
 section&1296707.0&0.01370864867178109466& 1.5627400792931634\\
 table&292110.0&0.00308815589297657504& 2.431403227076493\\
 table caption&206215.0&0.00218008307647860186& 3.415961047904482\\
 table cell&12343327.0&0.13049234197386899714& 4.4024482694671425\\
 table col&1285945.0&0.01359487395088754789& 4.423232071074114\\
 table row&2533799.0&0.02678705389568365736& 4.406505829892338\\
 tabular&280572.0&0.00296617738250735561& 3.4281299566929837\\
 title&16.0&0.00000016915030052934& 3.0\\
 \bottomrule
 \end{tabular}
 \caption{Summary statistics by \entity/ of \datasetauto/ dataset.}
 \label{tab:dataset_cat_statistics}
 \end{table}

\textbf{Component 4: Structure-Based Refinement}

\textbf{(4)} 
In our experiments, we use $r=30$ for structure-based refinement. During development, we observed only minor differences for values of $r=10$ and higher. To confirm this, we analyze the performance of \sys/~WS+FT for $r={2,5,10,20,30}$ on the validation set (see Table \ref{tab:iterationlimit}).\footnote{Note that results are given for a single model and can differ from the detailed \relation/ classification evaluation, where we average over three models.} Here we observe that the accuracy of our system remains unchanged for values of $r\geq10$.

\begin{table}[t]
\small
\sisetup
  { 
    round-precision    = 3           ,
    round-integer-to-decimal,
    round-mode = places,
    table-number-alignment=center,
    table-format=1.3,
  }
\renewrobustcmd{\bfseries}{\fontseries{b}\selectfont}
    \centering
\begingroup
\setlength{\tabcolsep}{1.25pt} 
\begin{tabular}{@{}l  SSS@{}}
    \toprule
$r$ &    \multicolumn{1}{c}{IoU=0.5} & \multicolumn{1}{c}{IoU=0.65} & \multicolumn{1}{c}{IoU=0.8} \\
\midrule
2 & 0.6191446028513238 & 0.5631364562118126 & 0.38391038696537677 \\ 
5 & 0.6800401203610832 & 0.6188565697091274 & 0.42728184553660986 \\ 
10 & 0.6800401203610832 & 0.6188565697091274 & 0.42728184553660986 \\ 
20 & 0.6800401203610832 & 0.6188565697091274 & 0.42728184553660986 \\ 
30 & 0.6800401203610832 & 0.6188565697091274 & 0.42728184553660986 \\ 
\bottomrule
    \end{tabular}
    \endgroup
\caption{Impact of $r$ on the \relation/ classification performance on the development set for a \sys/~WS+FT model.}
    \label{tab:iterationlimit}
\end{table}

We additionally provide pseudo-code for our refinement procedure in Algorithm \ref{alg:refinement}.
\begin{algorithm}
\KwData{Detected Entities}
\KwResult{Refined Entities; Hierarchical Relations}
counter=0\;
\While{$\mathrm{counter} \leq \theta$}{
counter++\;
Classify all hierarchical relations\;
\If{(1) Parent entity bounding boxes don't fully enclose children}{
Expand parent bounding boxes s.t. they enclose children\;
Go to start of loop\;
}
\If{(2) Directly nested entities of same category exist}{
Merge directly nested entities into a single entity\;
Go to start of loop\;
}
\If{(3) Siblings found that are not allowed to co-exist in hierarchy}{
Enclose groups of siblings with new, valid parent entities\;
Go to start of loop\;
}
\If{(4) Possible parent found in neighborhood of a parent-less entity}{
Expand matched parents bounding boxes to enclose child entities\;
Go to start of loop\;
}
Exit loop\;
}
Classify all hierarchical relations\;
\caption{Structure-based refinement.}
\label{alg:refinement}
\end{algorithm}

\textbf{Component 5: Scalable Weak Supervision}

To analyze the degree of noise in \datasetauto/, we evaluate the average precision for the weak annotations against the manually generated ground truth in \datasetmanual/. Table \ref{tab:weaklabel_accuracy} shows the accuracies of \datasetauto/ for different IoU values, as measured on the training split of \datasetmanual/. We observe various AP values of $0$, indicating the absence of the respective categories in \datasetauto/. Furthermore, for the majority of categories, the measured is relatively low ($\text{AP}\leq0.5$ for $\text{IoU}>=0.5$). This emphasizes the systematic noise in \datasetauto/ and confirms the positioning of our experimental setting in the domain of weak supervision.

\begin{table}[t]
\small %
    \centering
\sisetup
  { 
    round-precision    = 1           ,
    round-integer-to-decimal,
    round-mode = places,
    table-align-text-post=false,
    table-number-alignment=center,
    table-format=2.1,
  }
\renewrobustcmd{\bfseries}{\fontseries{b}\selectfont}
\begingroup
\setlength{\tabcolsep}{2.25pt} 
    \begin{tabular}{@{}l  SSS@{}}
    \toprule
\multicolumn{1}{c}{AP} &    \multicolumn{1}{c}{IoU=0.5} & \multicolumn{1}{c}{IoU=0.65} & \multicolumn{1}{c}{IoU=0.8} \\
\midrule
mAP & 29.977554779734234 & 25.475914865315847 & 19.366606625962824 \\ 
\midrule
abstract & 61.06577246283129 & 43.51343101343102 & 9.826007326007327 \\ 
affiliation & 0 & 0 & 0 \\ 
author & 0 & 0 & 0 \\ 
bib. block & 47.582417582417584 & 47.582417582417584 & 31.605894105894105 \\ 
cont. block & 38.93668092676062 & 32.74791185984912 & 23.168206579008373 \\ 
date & 0 & 0 & 0 \\ 
equation & 26.18786176370446 & 24.505425626467517 & 22.83341845420497 \\ 
fig. caption & 24.930927647166012 & 23.822782223547655 & 23.822782223547655 \\ 
fig. graphic & 18.48664462477492 & 14.713396665899808 & 14.713396665899808 \\ 
figure & 26.516771788717854 & 19.497255020713723 & 11.230335168582743 \\ 
footer & 0 & 0 & 0 \\ 
header & 0 & 0 & 0 \\ 
heading & 33.58515427761111 & 33.58515427761111 & 33.58515427761111 \\ 
item & 25.09578544061302 & 9.616858237547893 & 4.789272030651341 \\ 
itemize & 54.54545454545454 & 42.42424242424242 & 42.42424242424242 \\ 
keywords & 0 & 0 & 0 \\ 
page nr. & 0 & 0 & 0 \\ 
tab. caption & 76.25673156586579 & 76.25673156586579 & 72.65225421154378 \\ 
table & 74.66081277630549 & 60.483797846566546 & 44.60536573813014 \\ 
tabular & 91.70008019246201 & 80.76889296215676 & 52.07580331393269 \\ 
\bottomrule
\end{tabular}
    \endgroup
\caption{Average precision (AP) of \entities/ in \datasetauto/, compared to the training split of \datasetmanual/}
\label{tab:weaklabel_accuracy}
\end{table}

\section{Computational Setup}
\textbf{Mask R-CNN} 
\begin{figure*}[t!]
    \centering
    \includegraphics[width=0.8\linewidth]{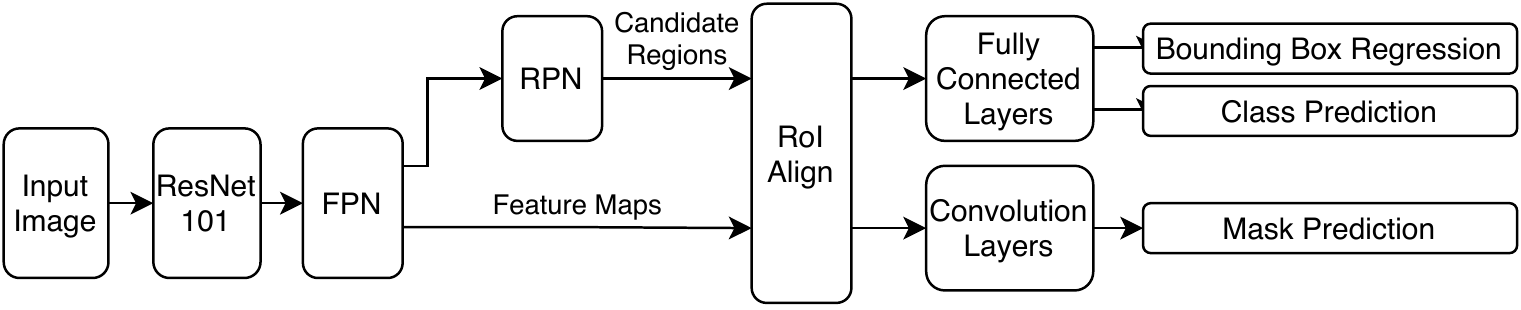}
    \caption{Mask R-CNN overview.}
    \label{fig:maskrcnn}
\end{figure*}
Our used Mask R-CNN model is illustrated in Figure \ref{fig:maskrcnn}.

\section{Related Work}
\subsection{Weak Supervision for Document Layout:} \citep{zhong2019publaynet} (PN) use weak supervision for detection of page layout entities. The dataset features 5 coarse categories, compared to 23 fine-grained categories in \dataset/. Furthermore the system does not contain a relation classification component. Following, we examine differences and correspondences between the five classes in \cite{zhong2019publaynet} and our \dataset/:
\begin{itemize}
    \item \textsc{Text:} Corresponds to \textsc{content block} in \dataset/. In contrast to our dataset, the \textsc{text} category corresponds to individual paragraphs (instead of uninterrupted text on a single column) and is used for captions.
    \item \textsc{Title:} corresponds to our \textsc{header} category. 
    \item \textsc{List:} corresponds to our \textsc{itemize} category. A difference here is that \textsc{list} entities in PN are separated by columns.
    \item \textsc{Table:} corresponds to our \textsc{tabular}. In contrast to \dataset/, they do not feature nesting relations that contain, for instance, \textsc{table caption} entities. Fine-grained children, such as cells, rows and columns are also not featured.
    \item \textsc{Figure:} corresponds roughly to the concept of \textsc{figure graphic} in \dataset/. However, no nesting relations (\ie/ sub-figures) or captions are featured.
\end{itemize}
We evaluate the feasibility of using the dataset presented in PN for pre-training. We use the same pre-training procedure as in our experiments that utilize \dataset/. To account for the difference of pre-training and target domains, we use an extended fine-tuning procedure of PN that matches the pre-training scheme of up to 80,000 iterations. Table \ref{tab:publaynet} shows results for entity detection. Here we observe that pre-training improves the performance of the system, when compared to \sys/~Baseline that does not use weak supervision. We also observe that pre-training with the PN dataset results in significantly lower mAP values, \eg/ $60.0$ after fine-tuning compared to $69.4$ in \sys/~WS+FT at $\text{IoU} = 0.5$. For some entity categories, we observe higher individual AP values for PN, e.g. \textsc{affiliation} at $\text{IoU} = 0.5$. We attribute this to the higher occurrence of more compact \textsc{text} entities in PN. Additionally, this could also be caused by our experimental protocol in which early stopping is applied function of the mAP value, instead of individual AP values. As such, there is a performance trade-off between individual entity categories.

\begin{table*}[t]
\small %
    \centering
\sisetup
  { 
    round-precision    = 1           ,
    round-integer-to-decimal,
    round-mode = places,
    table-align-text-post=false,
    table-number-alignment=center,
    table-format=2.1,
  }
\renewrobustcmd{\bfseries}{\fontseries{b}\selectfont}
\begingroup
\setlength{\tabcolsep}{2.25pt} 
    \begin{tabular}{@{}l  SS  SS SS@{}}
    \toprule
\multicolumn{1}{c}{} &    \multicolumn{2}{c}{IoU=0.5} & \multicolumn{2}{c}{IoU=0.65} & \multicolumn{2}{c}{IoU=0.8} \\
\cmidrule(lr){2-3} \cmidrule(lr){4-5} \cmidrule(lr){6-7}
\multicolumn{1}{@{}l}{AP}   &  \multicolumn{1}{c}{WS(PN)} & \multicolumn{1}{c}{WS+FT(PN)}&  \multicolumn{1}{c}{WS(PN)} & \multicolumn{1}{c}{WS+FT(PN)}&  \multicolumn{1}{c}{WS(PN)} & \multicolumn{1}{c@{}}{WS+FT(PN)}\\
\midrule
          mean AP &            8.42 &   \bfseries 60.0 &           3.79 &  \bfseries 48.32 &           2.46 &  \bfseries 25.86 \\
\midrule
     abstract &               0 &  \bfseries 89.15 &              0 &  \bfseries 80.45 &              0 &  \bfseries 50.15 \\
  affiliation &               0 &  \bfseries 63.84 &              0 &  \bfseries 35.83 &              0 &   \bfseries 4.81 \\
       author &               0 &  \bfseries 34.26 &              0 &  \bfseries 22.04 &              0 &   \bfseries 1.06 \\
   bib. block &               0 &  \bfseries 60.33 &              0 &  \bfseries 60.32 &              0 &  \bfseries 45.45 \\
  cont. block &           34.89 &  \bfseries 90.26 &          20.06 &  \bfseries 86.83 &          14.45 &  \bfseries 76.09 \\
         date &               0 &  \bfseries 16.67 &   \bfseries 0.0 &               0 &   \bfseries 0.0 &               0 \\
     equation &               0 &  \bfseries 77.33 &              0 &  \bfseries 57.05 &              0 &  \bfseries 17.97 \\
 fig. caption &               0 &  \bfseries 64.81 &              0 &  \bfseries 62.36 &              0 &  \bfseries 27.88 \\
 fig. graphic &               0 &  \bfseries 38.35 &              0 &  \bfseries 33.23 &              0 &  \bfseries 19.78 \\
       figure &           23.03 &   \bfseries 49.9 &           5.99 &  \bfseries 46.32 &           1.47 &  \bfseries 38.02 \\
       footer &               0 &  \bfseries 69.08 &              0 &  \bfseries 56.39 &              0 &   \bfseries 3.92 \\
       header &               0 &  \bfseries 76.88 &              0 &  \bfseries 63.68 &              0 &   \bfseries 5.73 \\
      heading &           27.43 &  \bfseries 63.82 &          12.29 &  \bfseries 46.44 &           2.88 &  \bfseries 22.69 \\
         item &               0 &   \bfseries 1.67 &              0 &   \bfseries 1.67 &   \bfseries 0.0 &               0 \\
      itemize &  \bfseries 36.54 &              25 &  \bfseries 12.5 &               0 &  \bfseries 25.0 &               0 \\
     keywords &               0 &   \bfseries 50.0 &              0 &  \bfseries 48.92 &              0 &  \bfseries 42.42 \\
     page nr. &               0 &  \bfseries 80.34 &              0 &  \bfseries 37.98 &              0 &   \bfseries 1.04 \\
 tab. caption &               0 &  \bfseries 62.67 &              0 &  \bfseries 50.03 &              0 &  \bfseries 21.08 \\
        table &           46.51 &  \bfseries 90.82 &          24.95 &  \bfseries 82.09 &           5.42 &  \bfseries 59.76 \\
      tabular &               0 &  \bfseries 94.78 &              0 &   \bfseries 94.7 &              0 &  \bfseries 79.38 \\
\bottomrule
\end{tabular}
    \endgroup
\caption{Average precision (AP) of \entity/ detection on the test set, using \cite{zhong2019publaynet} (PN) and structure-based refinement.}
\label{tab:publaynet}
\end{table*}

\section{Robustness Check: Table Structure Parsing}

We perform robustness checks of \sys/ on the table structure parsing task. \sys/ is evaluated for \entity/ detection on \datasetmanual/ and structure parsing on the ICDAR 2013 table structure dataset.  

We received the outputs for the ICDAR ``competition'' dataset from the authors of \cite{Nurminen2013}. We used the evaluation script provided by the competition organizers to calculate the ICDAR~50\,\% performance.

We match our table cell predictions with the text element locations provided by \citep{Nurminen2013} in order to generate XML files that are compared to the ground truth by the scripts provided on the competition website. Matches are determined by the fraction of overlap  between cell and text bounding boxes $\gamma = \frac{\mathit{area}(B_{\text{cell}} \cap B_{\text{text}})}{\mathit{area}(B_{\text{text}})}$, using $\gamma \geq 0.5$.

\subsection{Table Structure Heuristics}
For the ordering of table structure \entities/, we draw upon a set of special heuristics. The reason for this is that nesting relationships are often too complex to model with the previously described parent-child relationships, \eg/ for cells belonging to multiple rows and/or columns. Due to these complex \relations/, bottom-up creation of table row and table column \entity/ bounding boxes from associated children is also challenging. We, therefore, generate rows, columns, and cells on the same hierarchical levels and store structure information in an additional attribute in each \entity/. 

The following heuristics are applied: 
\begin{enumerate}
    \item Rows are sorted, based on the $y$-coordinate of their centroids. Columns are analogously sorted, based on their centroid $x$-coordinates. 
    \item Row \entities/ that are located such that their bounding box is fully contained inside the bounding box of other row \entities/ are determined. All such direct nestings are resolved as follows: (1) If a row \entity/ contains exactly one other row \entity/, remove the contained \entity/. (2) Remove row \entities/ that contain more than one other row \entity/. Analogously, we proceed to discard column \entities/ with direct nesting.
    \item The bounding box (\ie/, ``union'') of all row and column \entities/ is computed. However, the size of this bounding box might differ from the bounding boxes of the row and column \entities/. Hence, the bounding boxes of all rows are adjusted so that all adjacent rows have the width as the ``union''. Analogously, the height for all bounding boxes belonging to columns are adjusted. 
    \item The location of rows might not be located at the center of adjacent rows. This is achieved by setting the $y$-coordinate of each row to the average of its adjacent rows. An analogous adjustment is performed for the $x$-coordinates of columns. 
    \item Row and column numbers are assigned to separately detected cells as follows: for all cell \entities/ from \sys/,  we calculate the overlap between the vertical cell border and all vertical row borders. We then calculate the rows for which the length of the overlap is equal or larger than 50\,\% of the height of a row. The number of the corresponding row is then assigned to the row range of the cell. Analogously, we match cells to columns based on their horizontal overlap. If a cell is matched with more than one row or column, its bounding box is adjusted such that its borders lie on the grid of row and column borders. All other cells without assignment are dismissed.
    \item A grid of rectangular cells is generated from the intersection of all rows and columns for all positions in the table where no multi-row or multi-column cell exists. 
\end{enumerate}  

\subsection{Implementation Details}
\textbf{\Entity/ Detection}
We use the hierarchical document annotations in \datasetauto/ to identify \num{222195} table structure \entities/ that are used for weak supervision. The corresponding cropped tabular regions and their child \entities/, \ie/, rows, columns, and cells, are used as training input for the specialized system. The sampling process is stratified to bolster prediction performance: we use all row and column annotations, but only a subset of all table cell annotations. The reason is that regular cells can be reconstructed from robust detections of rows and columns. Row and column detection performance can, however, be adversely affected by category imbalance during sampling. The comparably large number of individual table cells per input creates such imbalance. Therefore, we only sample table cells that appear in the first table row and column, as well as cells spanning multiple rows or columns. Altogether, this aids the detection of multi-row and -column cells. Again, these cells can not be robustly reconstructed from regular rows and columns otherwise. The parameters for \entity/ samples per image, ground truth samples per image and maximum number of predictions per image are set to $200$, $200$ and $400$, respectively. 

The train, validation and test splits of \datasetmanual/ contain $87$, $39$, and $61$ tabular \entities/, respectively. Crops of the \entities/ are used for training and evaluation of the system specialized for table structure. 

\textbf{ICDAR 2013 Table Structure Dataset:} The ICDAR 2013 table structure dataset \cite{Gobel2013} is designed to evaluate table structure parsing. This dataset is later leveraged as part of our robustness check so that we can evaluate our weak supervision against state-of-the-art approaches for structure parsing. The dataset consists of $123$ images, for which structure annotations, including cells, rows, and columns were created. The dataset comes without predefined train/test split; hence, we follow \citet{Schreiber2018} and split the so-called ``competition'' part of the dataset with a $50\%/50\%$-ratio. One of the splits is used for evaluation. The other split is used in addition to the so-called ``practice'' part of the dataset for training and validation. We follow the official competition rules from ICDAR 2013 as follows: we operate directly on table sub-regions and thus create individual cropped images of these regions for training, validation, and evaluation. We generate rectangular row and column bounding boxes from the provided cell bounding boxes and their respective row- and column ranges. The resulting rows and columns are then further modified as follows: A tabular bounding box is determined as union bounding box of all cells. Bounding boxes of rows that share a border with the outer tabular are extended such that their borders fully align with the tabular. Afterwards, we move the borders of all pairs of neighboring rows to their respective midpoint. Analogously, we adjust all column bounding boxes. Cell bounding boxes are newly created from row and column intersections in a final step.

\subsubsection{\Entity/ Detection on \datasetmanual/ }

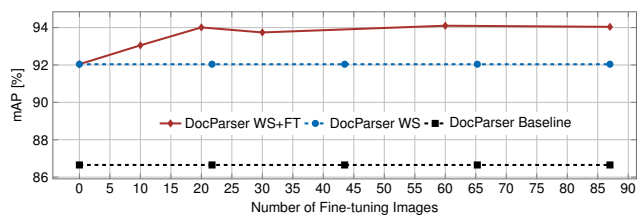
\begin{figure}[t!]
    \centering
\begin{adjustbox}{width=\linewidth}
\begin{tikzpicture}
  \begin{axis}[
    width=\textwidth,
    height=.35\textwidth,
      xlabel={Number of Fine-tuning Images},
      ylabel={mAP [\%]},
        enlarge x limits={0.05},
        legend style={draw=none},
        ymajorgrids = true,
        xmajorgrids = true,
        axis line style = {ultra thin, ETH6},
        legend style={at={(0.9,0.4)},
        legend columns=-1},
      tick label style = {font=\large\sansmath\sffamily},
      every axis label = {font=\large\sansmath\sffamily},
      legend style = {font=\large\sansmath\sffamily},
      label style = {font=\large\sansmath\sffamily},
      every axis plot/.append style={ultra thick},
  ]
        \addplot[mark=diamond*, ETH7, mark options={solid}] table[x=model,y=test]{\lowleveldetectionFTtransp};
     \addlegendentry{\sys/~WS+FT};
     \addplot[mark=*, mark options={solid}, dashed, ETH3, domain=0:87, samples=5] {92.04};
    \addlegendentry{\sys/~WS};
     \addplot[mark=square*, mark options={solid}, dashed, domain=0:87, samples=5] {86.65};
    \addlegendentry{\sys/~Baseline};
  \end{axis}
\end{tikzpicture}
\end{adjustbox}
 \caption{Comparison of test mAP (IoU=0.5) for three variants of \sys/ for detection of table structure annotations. We follow the same procedure as described for fine-tuning with the default document entities. The weakly supervised system \sys/~WS outperforms the baseline system without fine-tuning (FT). 
 Fine-tuning with 10 or more images yields additional performance gains.}
\label{fig:subsampling_finetuning_lowlevel}
\end{figure}

Analogously to our evaluation on full documents, we measure mAP for table rows and table columns on a subset of table regions in \datasetmanual/. Average precision for joint detection ofgg table rows and columns and the impact of fine-tuning are shown in Figure~\ref{fig:subsampling_finetuning_lowlevel}. Compared to full document pages, we measure higher mAPs for all systems. We observe that the weakly supervised model outperforms \sys/~Baseline without having been trained on the target domain. We observe additional significant performance improvements in \sys/~WS systems that were fine-tuned with $10$ to $87$ images. Because of the intricacies evaluating hierarchical structure parsing for tables, we perform a separate evaluation of \sys/ for this task.

\bibliographyappendix{references}

\end{appendices}

\end{document}